\documentclass{article}
\usepackage{graphicx}
\usepackage{amsmath}
\usepackage{hyperref}
\usepackage{fontawesome5}
\usepackage{geometry}
\usepackage{csquotes}
\usepackage{comment}
\usepackage[table,xcdraw]{xcolor}
\usepackage{listings}
\usepackage{authblk}
\usepackage{lipsum}
\usepackage{minitoc}
\usepackage{booktabs} 
\usepackage{longtable}
\usepackage{array}
\usepackage[justification=justified]{caption}
\usepackage{cite}
\usepackage{lmodern}
\usepackage{orcidlink}

\hypersetup{
    colorlinks=true,
    linkcolor=[RGB]{0,51,153},  
    urlcolor=[RGB]{0,51,153}    
    citecolor=[RGB]{0,51,153},    
    filecolor=[RGB]{0,51,153},
    allcolors=[RGB]{0,51,153} 
}

\noptcrule 
\mtcsetfont{parttoc}{section}{\fontsize{7.5pt}{7.5pt}\sffamily}

\newcommand{\github}[1]{%
   \href{#1}{\faGithubSquare}%
}

\title{Submissions and Reflections from the 2024 Large Language Model (LLM) Hackathon for Applications in Materials Science and Chemistry}

\author[38]{Yoel Zimmermann$^\dagger$\orcidlink{0009-0003-1720-4368}}
\author[54]{Adib Bazgir$^\dagger$\orcidlink{0000-0001-6475-8505}}
\author[1]{Zartashia Afzal}
\author[2]{Fariha Agbere}
\author[3]{Qianxiang Ai\orcidlink{0000-0002-5487-2539}}
\author[4]{Nawaf Alampara\orcidlink{0009-0001-7697-7315}}
\author[5]{Alexander Al-Feghali\orcidlink{0009-0004-8377-7049}}
\author[6]{Mehrad Ansari\orcidlink{0000-0001-5696-9193}}
\author[7]{Dmytro Antypov\orcidlink{0000-0003-1893-7785}}
\author[8]{Amro Aswad}
% ...

\affil[1]{University of the Punjab}
\affil[2]{University of Maryland, Baltimore County}
\affil[3]{Massachusetts Institute of Technology}
\affil[4]{Friedrich-Schiller-Universität Jena}
\affil[5]{McGill University}
\affil[6]{Acceleration Consortium}
\affil[7]{University of Liverpool}
\affil[8]{University of Toronto}
% ...
\affil[38]{ETH Zurich}
% ...
\affil[54]{University of Missouri-Columbia}
% ...
\date{}

\begin{document}
\doparttoc 
\faketableofcontents 
\maketitle
\thanks{$^*$Corresponding author: blaiszik@uchicago.edu}
\thanks{$^\dagger$These authors also contributed substantially to compiling team results and other paper writing tasks}

\begin{abstract}
    Here, we present the outcomes from the second Large Language Model (LLM) Hackathon for Applications in Materials Science and Chemistry, which engaged participants across global hybrid locations, resulting in 34 team submissions. The submissions spanned seven key application areas and demonstrated the diverse utility of LLMs for applications in (1) molecular and material property prediction; (2) molecular and material design; (3) automation and novel interfaces; (4) scientific communication and education; (5) research data management and automation; (6) hypothesis generation and evaluation; and (7) knowledge extraction and reasoning from scientific literature. Each team submission is presented in a summary table with links to the code and as brief papers in the appendix. Beyond team results, we discuss the hackathon event and its hybrid format, which included physical hubs in Toronto, Montreal, San Francisco, Berlin, Lausanne, and Tokyo, alongside a global online hub to enable local and virtual collaboration. Overall, the event highlighted significant improvements in LLM capabilities since the previous year's hackathon, suggesting continued expansion of LLMs for applications in materials science and chemistry research. These outcomes demonstrate the dual utility of LLMs as both multipurpose models for diverse machine learning tasks and platforms for rapid prototyping custom applications in scientific research. 
\end{abstract}

\section*{Introduction}

Science hackathons have emerged as a powerful tool for fostering collaboration building, innovation, and rapid problem-solving in the scientific community~\cite{nolte2020support, pe2019understanding, heller2023hack}. By leveraging social media, virtual platforms, and hybrid event structures, such hackathons can be organized in a cost-effective manner while maximizing their impact and reach. In this article, we first introduce the project submissions to the second Large Language Model Hackathon for Applications in Materials Science and Chemistry, detailing the broad classes of problems addressed by teams, while analyzing trends and patterns in the approaches taken. We then present each team submission in turn, plus a summary table with the names of team members and links to code repositories where available. Finally, we include the detailed project documents submitted by each team, showcasing the depth and breadth of innovation demonstrated during the hackathon.

\section*{Overview of Submissions}

The hackathon resulted in 34 team submissions (with 32 submissions with a written description included here), categorized as shown in \autoref{tab:submissions}. From these submissions, we identified seven key application areas:
\begin{enumerate}
    \item \textbf{Molecular and Material Property Prediction}: Forecasting chemical and physical properties of molecules and materials using LLMs, particularly excelling in low-data environments and combining structured/unstructured data.

   \item \textbf{Molecular and Material Design}: Generation and optimization of novel molecules and materials using LLMs, including peptides, metal-organic frameworks, and sustainable construction materials.

    \item \textbf{Automation and Novel Interfaces}: Development of natural language interfaces and automated workflows to simplify complex scientific tasks, making advanced tools and techniques more accessible to researchers.

    \item \textbf{Scientific Communication and Education}: Enhancement of academic communication, automation of educational content creation, and facilitation of learning in materials science and chemistry.

    \item \textbf{Research Data Management and Automation}: Streamlining the handling, organization, and processing of scientific data through LLM-powered tools and multimodal agents.

    \item \textbf{Hypothesis Generation and Evaluation}: Generation, assessment, and validation of scientific hypotheses using LLMs, often combining multiple AI agents and statistical approaches.

    \item \textbf{Knowledge Extraction and Reasoning}: Extraction of structured information from scientific literature and sophisticated reasoning about chemical and materials science concepts through knowledge graphs and multimodal approaches.
    
\end{enumerate}
We next discuss each application area in more detail and highlight exemplar projects in each.

\subsection*{1. Molecular and Material Property Prediction}
LLMs have rapidly advanced in 
this area, 
employing both textual and numerical data to forecast a wide range of properties. Recent studies~\cite{Jablonka_Schwaller_Ortega-Guerrero_Smit_2024, qian2023largelanguagemodelsempower, jacobs2024regressionlargelanguagemodels} show LLMs performing comparably to, or even surpassing, conventional machine learning methods in this domain, \textit{particularly in low-data environments}. Their flexibility in processing both structured and unstructured data, as well as their general applicability to regression tasks~\cite{vacareanu2024from}, makes them a powerful tool for diverse predictive tasks in molecular and materials science.

\vspace{0.7ex}
\noindent 
\underline{\textbf{Exemplar projects}}:
The \textbf{Learning LOBSTERs} team (Ueltzen et al.) integrated bonding analysis data into LLMs to predict the phonon density of states (DOS) for crystal structures, showing that combining covalent bonding information with LLM capabilities can improve prediction accuracy. The \textbf{Liverpool Materials} team (Ottomano et al.) demonstrated how adding context from materials literature, via models like MatSciBert~\cite{Gupta2022main}, could improve the prediction of Li-ion conductivity, particularly when dealing with limited training data. The \textbf{MolFoundation} team (Harb et al.) benchmarked ChemBERTa~\cite{chithrananda2020chembertamain} and T5-Chem~\cite{Lu2022} for molecular property predictions, finding that pre-trained models performed comparably to fine-tuned models, suggesting that expensive fine-tuning might not always be required. Another contribution came from the \textbf{Geometric Geniuses} team (Weinreich et al.), who focused on incorporating 3D molecular geometries into LLMs, experimenting with encoding geometric features to improve prediction outcomes. 

\subsection*{2. Molecular and Material Design}

LLMs have also advanced in molecular and material design,
proving capable in both settings~\cite{Bhattacharya2024Large, Liu2024Multimodal, Jia2024LLMatDesign, Jang2024Can, Lu2024Generative}, especially if pre-trained or fine-tuned with domain-specific data~\cite{pmlr-v235-kristiadi24a}. However, despite these advancements, LLMs still face limitations in practical applications~\cite{Miret2024Are}.

\vspace{0.7ex}
\noindent 
\underline{\textbf{Exemplar projects}}:
During the hackathon, teams tackled these challenges through different approaches. The team behind \textbf{MC-Peptide} (Bran et al.) developed a workflow harnessing LLMs for the design of macrocyclic peptides (MCPs). By employing semantic search and constraint-based generation, they automated the extraction of literature data to propose MCPs with improved permeability, crucial for drug development~\cite{Ji2024main, Merz2024main}. Meanwhile, the \textbf{MOF Innovators} team (Ansari et al.) focused on metal-organic frameworks (MOFs). Their AI agent utilized retrieval-augmented generation (RAG)~\cite{Lewis2020Retrieval} to incorporate design rules extracted from the literature, and an ensemble of fine-tuned surrogate models to optimize MOF's band gap property. In pursuit of sustainable materials, the \textbf{Green Construct} team (Canalicchio et al.) investigated small-scale LLMs such as Llama 3~\cite{Dubey2024Llama} and Phi-3~\cite{Abdin2024Phi} to streamline the design of alkali-activated concrete. Their models provided insights into reducing emissions of traditional materials like Portland cement through zero-shot learning.

\subsection*{3. Automation and Novel Interfaces}
LLMs are increasingly used 
to simplify and streamline access to complex tools~\cite{Bran2023ChemCrow, Song2023RestGPT, Zhang2024HoneyComb}, autonomously plan and execute tasks~\cite{Boiko2023Autonomous}, and interface with robotic systems in lab settings~\cite{Darvish2024ORGANA, Tom2024Self}. These capabilities enhance the efficiency of scientific workflows, allowing researchers to focus on higher-level problem-solving rather than routine tasks. As LLMs continue to evolve, they are expected to further transform laboratory practices and democratize access to advanced experimental and computational techniques.

\vspace{0.7ex}
\noindent 
\underline{\textbf{Exemplar projects}}:
The team behind \textbf{LangSim} (Chiang et al.) addressed the complexity of atomistic simulation software by creating a natural language interface to automate the calculation of material properties such as bulk modulus. By integrating LangChain~\cite{Bagatur2024langchain} agents with LLMs, LangSim enables flexible construction and execution of simulation workflows, reducing user barriers and expanding functionality to multi-component alloys. Similarly, \textbf{LLMicroscopilot} (Schloz \& Gonzalez) seeks to simplify the operation of sophisticated microscopes by using LLM-powered agents to automate tasks like experimental parameter estimation. This could reduce the need for highly trained operators, making advanced microscopy tools more accessible. In the realm of Density Functional Theory (DFT) calculations, the team behind \textbf{T2Dllama} (Parida \& Petersen) developed a tool that uses RAG to extract optimized simulation parameters from scientific literature, aiming to assist experimentalists with complex DFT setups. The tool simplifies the process of obtaining reliable parameters and reduces dependency on computational chemistry expertise. Complementing this, \textbf{Materials Agent} (Datar et al.) provides a tool-calling capability for cheminformatics, combining molecular property calculations, simulations, and document interaction through a natural language interface. This agent was developed to lower the barrier for researchers in cheminformatics and to accelerate the pace of research by integrating tools like RDKit~\cite{rdkitmain} and custom simulations into an intuitive system.

\subsection*{4. Scientific Communication and Education}

LLMs are transforming how scientific and educational content is created and shared, enhancing accessibility and personalized learning~\cite{Yan2023Practical, Wang2024Large, Kasneci2023ChatGPT, Schäfer2023Notorious}. By automating tasks like question generation, feedback, and grading, LLMs streamline educational processes, freeing educators to focus on individual learning needs. Additionally, LLMs assist in translating complex scientific findings into accessible formats, broadening public engagement~\cite{Schäfer2023Notorious}. However, technological readiness, transparency, and ethical concerns around data privacy and bias remain critical challenges to address~\cite{Kasneci2023ChatGPT, Yan2023Practical}.

\vspace{0.7ex}
\noindent 
\underline{\textbf{Exemplar projects}}:
One submission in this category is \textbf{MaSTeA} (Circi et al.), an automated system focused on evaluating the effectiveness of LLMs as teaching assistants by testing their performance on materials science questions from the MaScQA~\cite{Zaki2023MaScQA} dataset. By analyzing different types of questions—such as multiple-choice, match-the-following, and numerical problems—across 14 topical categories, the team identified the strengths and weaknesses of various models. An interactive platform was developed to facilitate easy model testing and, as models get better, to eventually help students practice answering questions and understand solution steps. Meanwhile, the \textbf{LLMy Way} team (Zhu et al.) focused on simplifying the process of creating academic presentations by using LLMs to automatically generate structured slide decks from research articles. The tool formats summaries of papers into typical sections—such as background, methods, and results—and transforms them into presentation slides. It offers customization based on audience expertise and time constraints, aiming to make the communication of complex topics more efficient and effective. Lastly, the \textbf{WaterLLM} team (Baibakova et al.) sought to address water pollution challenges, particularly in decentralized communities lacking centralized water treatment infrastructure. They developed a chatbot that uses LLMs enhanced with RAG to suggest optimal water purification methods for microplastic contamination. Grounded in up-to-date scientific literature, the chatbot provides tailored purification protocols based on contaminant characteristics, resource availability, and cost considerations, promoting effective water treatment solutions for underserved areas.

\subsection*{5. Research Data Management and Automation}
Various submission were received in this area that attempt to enhance the management, accessibility, and automation of scientific data workflows using LLMs. These efforts, often leveraging multimodal agents, aim to simplify complex data handling, improve reproducibility, and accelerate insights across diverse scientific disciplines. 

\vspace{0.7ex}
\noindent 
\underline{\textbf{Exemplar projects}}:
\textbf{yeLLowhaMmer} (Evans et al.), a multimodal LLM-based data management agent automates data handling within electronic lab notebooks (ELNs) and laboratory information management systems (LIMS). The tool processes text and image instructions to generate Python code targeting the API of the \emph{datalab} ELN/LIMS system \cite{Evans2024a}, performing tasks such as summarizing experiments or digitizing handwritten notes. Powered by the Claude 3 Haiku model~\cite{Claude3} and LangChain~\cite{Bagatur2024langchain}, yeLLowhaMmer was developed to simplify the interaction with scientific data and has future potential for integrating audiovisual data processing. In parallel, the team behind \textbf{LLMads} (Kapoor et al.) explored how LLMs can be used to automate data parsing by converting raw scientific data, like X-ray diffraction (XRD) measurements, into structured metadata schemas. LLMads aims to replace the need for custom parsers with models like Mixtral-8x7b~\cite{Jiang2024Mixtral}, which extract data from raw files and populate schemas automatically. The process involves chunking raw data and prompt engineering, although challenges such as hallucinations and the extraction of multi-dimensional data still require further refinement. The \textbf{NOMAD Query Reporter} team (Daelman et al.) developed an LLM-based agent that uses RAG to generate context-aware summaries from large materials science repositories like NOMAD~\cite{Draxl2019NOMAD}. The tool produces detailed reports on experimental methodologies and results, which could help researchers in drafting publication sections. Using a self-hosted Llama3 70B model, the system engages in multi-turn conversations to ensure context retention. Further, in the \textbf{Speech-schema-filling} project, Näsström et al. integrated speech recognition and LLMs to automate the conversion of spoken language into structured data within lab environments. This method, utilizing OpenAI's Whisper~\cite{Radford2022Robust} for transcription and LLMs for schema selection and population, allows researchers to input experimental data verbally, which is then structured into JSON or Pydantic models for entry into ELNs, which could be especially helpful in situations where manual entry is difficult.

\subsection*{6. Hypothesis Generation and Evaluation}
LLMs can be leveraged to streamline scientific inquiry, hypothesis generation, and verification. Recent work across psychology, astronomy, and biomedical research demonstrates their capacity to generate novel, validated hypotheses by integrating domain-specific data structures like causal graphs~\cite{Zhou2024Hypothesis, Abdel2024Scientific, Tong2024Automating, Ciuca2023Harnessing}. Although still largely untapped in chemistry and materials science, this approach holds substantial promise for accelerating discovery and innovation in these fields~\cite{Liu2024Beyond, Shir2024Towards}.

\vspace{0.7ex}
\noindent 
\underline{\textbf{Exemplar projects}}:
In an individual contribution, \textbf{Marcus Schwarting} employed LLMs and temporal Bayesian statistics to assess scientific claims, exemplified by evaluating the hypothesis that LK-99 is a room-temperature superconductor. By using LLM-based natural language inference (NLI) to classify relevant studies and applying Bayesian updating, his system tracks how the scientific consensus evolves over time, with the goal of allowing even non-experts to gauge quickly the validity of a claim. In a related project, the \textbf{Thoughtful Beavers} team (Ozhan \& Mahjoubi) prototyped a Multi-Agent System of specialized LLMs that aims to accelerate scientific hypothesis generation by coordinating agents that extract background information, generate inspirations, and propose hypotheses, which are then evaluated for feasibility, utility, and novelty. By applying the ``Tree of Thoughts" framework~\cite{NEURIPS2023_271db992}, this system streamlines the creative process and improves the quality of scientific hypotheses, as demonstrated through a case study on sustainable concrete design. Another project, \textbf{ActiveScience} (Chiu), integrated LLMs, knowledge graphs, and RAG to extract high-level insights from scientific articles. Focused on materials science research, particularly alloys, this framework organizes extracted information into a Neo4j knowledge graph, allowing for complex querying and discovery. Additionally, a first-pass peer review system was developed using fine-tuned LLMs, \textbf{G-Peer-T} (Al-Feghali \& Zhang), to assess materials science papers. By analyzing the log probabilities of abstracts, the system flags works that deviate from typical scientific language patterns, helping to identify both highly innovative and potentially nonsensical research. Preliminary findings suggest that highly cited papers tend to use less typical language, highlighting the potential for LLMs to support the peer review process and detect outlier research.

\subsection*{7. Knowledge Extraction and Reasoning}
Here the interest is in LLMs enabling extraction of structured scientific knowledge from unstructured text, assisting researchers in navigating complex academic content~\cite{Shamsabadi2024Large, Dagdelen2024Structured, xu2024largelanguagemodelsgenerative, Li2024Generative}. These systems streamline tasks like named entity recognition and relation extraction, offering flexible solutions tailored to materials science and chemistry~\cite{Dagdelen2024Structured}. Tool-augmented frameworks help LLMs address complex reasoning by leveraging scientific tools and resources, expanding their utility as assistants in scientific research~\cite{ma2024sciagenttoolaugmentedlanguagemodels}.

\vspace{0.7ex}
\noindent 
\underline{\textbf{Exemplar projects}}:
The team behind \textbf{ChemQA} (Khalighinejad et al.) created a multimodal question-and-answering dataset for chemical understanding. It highlights the importance of combining textual and visual inputs (e.g., SMILES and molecule images) to improve model accuracy in tasks like counting atoms, molecular weight calculations, and retrosynthesis planning. This multimodal approach shows how foundational models in chemistry benefit from rich, diverse data representations. In the realm of lithium metal batteries, \textbf{LithiumMind} (Ni et al.), integrated LLMs to extract Coulombic Efficiency (CE) and electrolyte information from scientific literature. The team also developed a chatbot powered by RAG to answer queries on lithium battery research. By constructing a knowledge graph, the system visualizes relationships between materials and properties, enhancing research accessibility and fostering domain-specific insights. The \textbf{KnowMat} project (Sayeed et al.) tackled the transformation of unstructured materials science literature into structured formats, using LLMs to extract essential information and convert it into machine-readable data (JSON). KnowMat's customizable prompts and integration capabilities streamline data extraction and analysis, making it a potentially valuable tool for researchers looking to accelerate their workflow in materials science. A similar project, \textbf{Ontosynthesis} (Ai et al.), explored extracting structured data from organic synthesis texts using LLMs. By employing Resource Description Framework (RDF) graphs, the project improved how chemical reactions are represented, with the extracted data aiding in retrosynthesis and condition recommendation tasks. This approach could help bridge the gap between unstructured chemical descriptions and standardized data formats. For high entropy alloys (HEAs) in hydrogen storage, the \textbf{Insight Engineers} team (Pradhan \& Biswajeet) explored synthetic data generation as a means to accelerate predictive modeling. The team proposed using LLMs like GPT-4~\cite{OpenAI2023GPT}, alongside custom prompts and a RAG framework, to generate synthetic data for machine learning interatomic potentials (MLIPs) to overcome the computational challenges associated with HEA properties prediction. Lastly, \textbf{GlossaGen} (Lederbauer et al.) tackled the challenge of generating glossaries for academic papers and grant proposals. It uses LLMs to extract scientific terms and definitions from PDF or TeX files and presents them in a knowledge graph. This visualization of relationships between concepts enhances understanding, and there are possible expansions of the tool to offer LaTeX integration and more refined ontology creation to aid researchers in navigating complex scientific terminology.



\section*{Hackathon Event Overview}
On May 9th, 2024, we hosted the second annual Large Language Model (LLM) Hackathon for Applications in Materials Science and Chemistry. The event brought together students, postdocs, researchers from industry, citizen scientists, and more, with participants joining both virtually and at in-person sites across multiple continents (Figure 1). The event was a follow-up to the previous hackathon described in detail here.~\cite{hackathon-2023} 
The event began with a kickoff panel featuring leading researchers from academia and industry, including Elsa Olivetti (MIT), Jon Reifsneider (Duke), Michael Craig (Valence Laboratories), and Marwin Segler (Microsoft). The charge of the hackathon was intentionally open-ended; i.e., to explore the vast potential application space, and create tangible demonstrations of the most innovative, impactful, and scalable solutions in a constrained time using open-source and best-in-class multimodal models applied to problems in materials science and chemistry.

\begin{figure}
    \centering
    \includegraphics[width=0.75\linewidth]{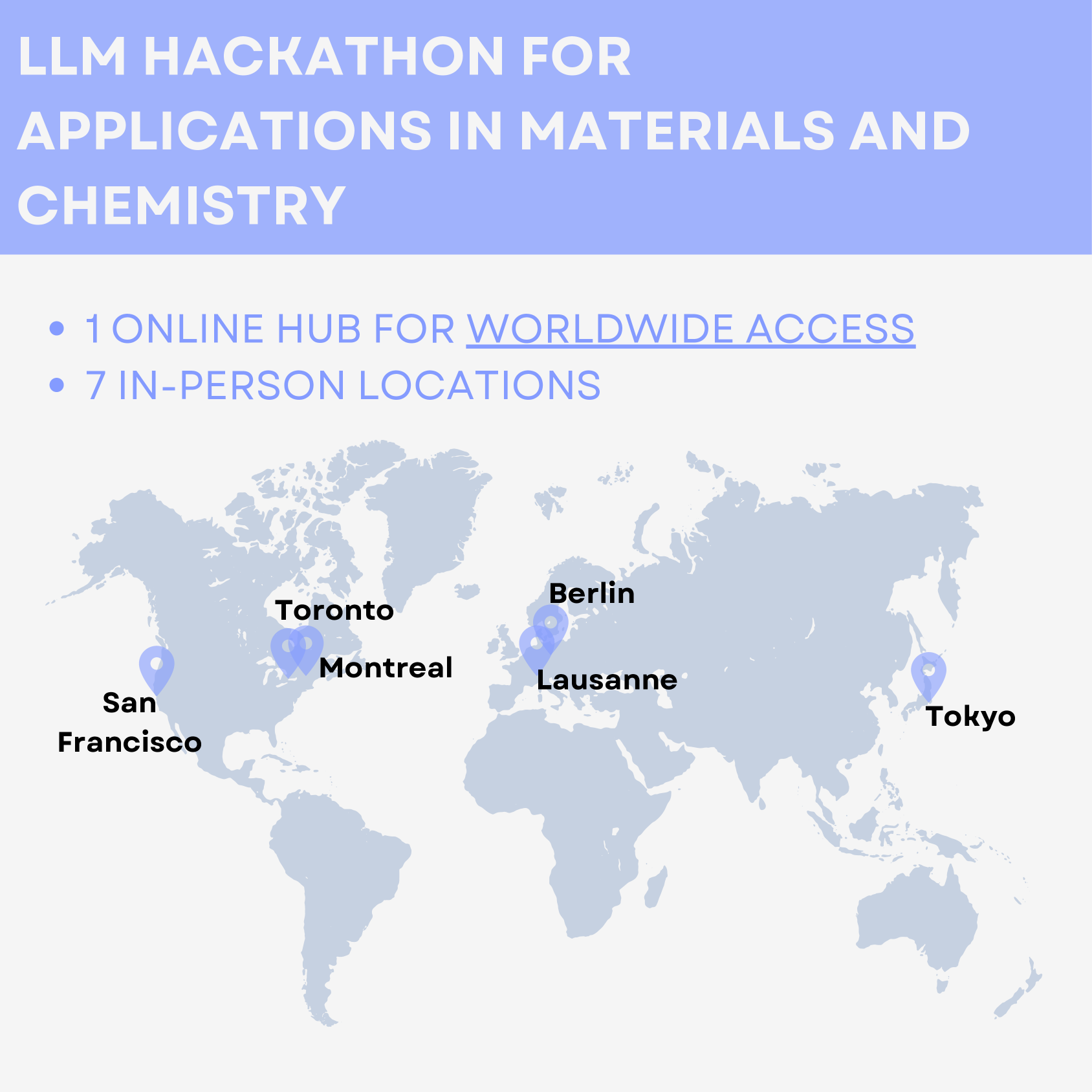}
    \caption{LLM Hackathon for Applications in Materials and Chemistry hybrid hackathon. Researchers were able to participate from both remote and in-person locations (purple pins).}
    \label{fig:map}
\end{figure}

Event registration included 556 participants and over 120 researchers comprising 34 teams that submitted completed projects. The hybrid format proved particularly successful, with physical hub locations in Toronto, Montreal, San Francisco, Berlin, Lausanne, and Tokyo facilitating local collaboration while maintaining global connectivity across the physical hubs and remote participants through virtual platforms (\autoref{fig:map}). This distributed approach enabled researchers to participate from either a local site or remotely from anywhere on Earth. This format blended the strengths of in-person events with the flexibility of remote participation leading to an inclusive event that led to team formation that crossed international and institutional boundaries, the submitted projects in this paper, and the growth of a new persistent online community of 483 researchers via Slack.

\section*{Conclusion}

The LLM Hackathon for Applications in Materials Science and Chemistry has demonstrated the dual utility and immense promise of large language models serving as 1) multipurpose models for diverse machine learning tasks and 2) platforms for rapid prototyping. Participants effectively utilized LLMs to tackle specific challenges while rapidly evaluating their ideas over a short 24-hour period, showcasing their ability to enhance the efficiency and creativity of research processes in highly diverse ways. It's important to note that many projects benefited from significant advancements in LLM performance since last year’s hackathon. That is, the performance across the diverse application space was improved simply via the release of new versions of Gemini, ChatGPT, Claude, Llama, and other models. If this trend continues, we expect to see even broader applications in subsequent hackathons, and in materials science and chemistry more generally.

Additionally, the hackathon hybrid format has proven effective towards creating new scientific collaborations and communities. By uniting individuals from various backgrounds and areas of expertise, these events facilitate knowledge exchange and promote interdisciplinary approaches, which are essential for advancing research in this rapidly evolving field.

As the integration of LLMs continues to expand, collaborative initiatives like hackathons will play a critical role in driving innovation and addressing complex challenges in chemistry, materials science, and beyond. The outcomes from this event highlight the significance of leveraging LLMs for their adaptability and their potential to accelerate the development of new concepts and applications.

\renewcommand{\arraystretch}{1.5} 

\begin{longtable}{>{\raggedright\arraybackslash}p{0.50\linewidth} >{\raggedright\arraybackslash}p{0.35\linewidth} >{\raggedright\arraybackslash}p{0.08\linewidth}}
\caption{Overview of the tools developed by the various tools, and links to source code repositories. Full descriptions of the projects can be found in the appendix.\label{tab:submissions}}\\
\toprule
\textbf{Project} & \textbf{Authors} & \textbf{Links} \\
\midrule
\endfirsthead

\toprule
\textbf{Project} & \textbf{Authors} & \textbf{Links} \\
\midrule
\endhead

\bottomrule
\endfoot

\textbf{Molecular and Material Property Prediction}\\

\hyperref[sec:learning-lobsters]{Leveraging Orbital-Based Bonding Analysis Information in LLMs} & Katharina Ueltzen, Aakash Naik, Janine George & \href{https://github.com/kaueltzen/LLM_Hackathon_2024}{GitHub} \\

\hyperref[sec:liverpool-materials]{Context-Enhanced Material Property Prediction (CEMPP)} & Federico Ottomano, Elena Patyukova, Judith Clymo, Dmytro Antypov, Chi Zhang, Aritra Roy, Piyush Ranjan Maharana, Weijie Zhang, Xuefeng Liu, Erik Bitzek & \href{https://github.com/fedeotto/lpoolmat_llms}{GitHub} \\

\hyperref[sec:molfoundation]{MolFoundation: Benchmarking Chemistry LLMs on Predictive Tasks} & Hassan Harb, Xuefeng Liu, Anastasiia Tsymbal, Oleksandr Narykov, Dana O'Connor, Shagun Maheshwari, Stanley Lo, Archit Vasan, Zartashia Afzal, Kevin Shen & \href{https://github.com/shagunm1210/MolFoundation}{GitHub} \\

\hyperref[sec:geometric-geniuses]{3D Molecular Feature Vectors for Large Language Models} & Jan Weinreich, Ankur K. Gupta, Amirhossein D. Naghdi, Alishba Imran&  \href{https://github.com/janweinreich/geometric-geniuses}{GitHub}\\

\hyperref[sec:LLMSpectrometry]{LLMSpectrometry} & Tyler Josephson, Fariha Agbere, Kevin Ishimwe, Colin Jones, Charishma Puli, Samiha Sharlin, Hao Liu & \href{https://github.com/ATOMSLab/LLMSpectroscopy}{GitHub} \\

\midrule

\textbf{Molecular and Material Design}\\

\hyperref[sec:mc-peptide]{MC-Peptide: An Agentic Workflow for Data-Driven Design of Macrocyclic Peptides} & Andres M. Bran, Anna Borisova, Marcel M. Calderon, Mark Tropin, Rob Mills, Philippe Schwaller & \href{https://github.com/doncamilom/mc-peptide}{GitHub} \\

\hyperref[sec:porevoyant]{Leveraging AI Agents for Designing Low Band Gap Metal-Organic Frameworks} & Sartaaj Khan, Mahyar Rajabi, Amro Aswad, Seyed Mohamad Moosavi, Mehrad Ansari&  \href{https://github.com/mehradans92/PoreVoyant}{GitHub}\\

\hyperref[sec:llama3-small-organics]{How Low Can You Go? Leveraging Small LLMs for Material Design} & Alessandro Canalicchio, Alexander Moßhammer, Tehseen Rug, Christoph Völker&  \href{https://github.com/sandrocan/LLMs-for-design-of-alkali-activated-concrete-formulations}{GitHub}\\
\midrule

\textbf{Automation and Novel Interfaces}\\

\hyperref[sec:langsim]{LangSim} & Yuan Chiang, Giuseppe Fisicaro, Greg Juhasz, Sarom Leang, Bernadette Mohr, Utkarsh Pratiush, Francesco Ricci, Leopold Talirz, Pablo A. Unzueta, Trung Vo, Gabriel Vogel, Sebastian Pagel, Jan Janssen & \href{https://github.com/jan-janssen/LangSim}{GitHub} \\

\hyperref[sec:llmicroscopilot]{LLMicroscopilot: assisting microscope operations through LLMs} & Marcel Schloz, Jose C. Gonzalez & \href{https://gitlab.com/Schlozma/llm_autotem}{GitHub} \\

\hyperref[sec:t2dllama]{T2Dllama: Harnessing Language Model for Density Functional Theory (DFT) Parameter Suggestion} & Chiku Parida, Martin H. Petersen & \href{https://github.com/chiku-parida/T2Dllama}{GitHub} \\

\hyperref[sec:materials-agent]{Materials Agent: An LLM-Based Agent with Tool-Calling Capabilities for Cheminformatics} & Archit Datar, Kedar Dabhadkar & \href{https://github.com/dkedar7/materials-agent}{GitHub} \\

\hyperref[sec:materials-agent]{LLM with Molecular Augmented Token} & Luis Pinto, Xuan Vu Nguyen, Tirtha Vinchurkar, Pradip Si, Suneel Kuman & \href{https://github.com/luispintoc/LLM-mol-encoder}{GitHub} \\
\midrule

\textbf{Scientific Communication and Education}\\

\hyperref[sec:mastea]{MaSTeA: Materials Science Teaching Assistant} & Defne Circi, Abhijeet S. Gangan, Mohd Zaki & \href{https://github.com/abhijeetgangan/MaSTeA}{GitHub} \\

\hyperref[sec:llmy-way]{LLMy-Way} & Ruijie Zhu, Faradawn Yang, Andrew Qin, Suraj Sudhakar, Jaehee Park, Victor Chen & \href{https://github.com/Ray16/LLM_My_way}{GitHub} \\

\hyperref[sec:waterllm]{WaterLLM: Creating a Custom ChatGPT for Water Purification Using PromptEngineering Techniques} & Viktoriia Baibakova, Maryam G. Fard, Teslim Olayiwola, Olga Taran & \href{https://github.com/ViktoriiaBaib/WaterLLM}{GitHub} \\
\midrule

\textbf{Research Data Management and Automation}\\

\hyperref[sec:yellowhammer]{yeLLowhaMMer: A Multi-modal Tool-calling Agent for Accelerated Research Data Management} & Matthew L. Evans, Benjamin Charmes, Vraj Patel, Joshua D. Bocarsly& \href{https://github.com/datalab-org/yellowhammer}{GitHub} \\

\hyperref[sec:llmads]{LLMads} & Sarthak Kapoor, José M. Pizarro, Ahmed Ilyas, Alvin N. Ladines, Vikrant Chaudhary & \href{https://github.com/ka-sarthak/llmads/tree/llm-hackathon-submission}{GitHub} \\

\hyperref[sec:nomad-query-reporter]{NOMAD Query Reporter: Automating Research Data Narratives} & Nathan Daelman, Fabian Schöppach, Carla Terboven, Sascha Klawohn, Bernadette Mohr & \href{https://github.com/ndaelman-hu/nomad_query_reporter}{GitHub} \\

\hyperref[sec:speech-schema-filling]{Speech-schema-filling: Creating Structured Data Directly from Speech} & Hampus Näsström, Julia Schumann, Michael Götte, José A. Márquez & \href{https://github.com/hampusnasstrom/speech-schema-filling}{GitHub} \\
\midrule

\textbf{Hypothesis Generation and Evaluation}\\

\hyperref[sec:lk99-hypothesis]{Leveraging LLMs for Bayesian Temporal Evaluation of Scientific Hypotheses} & Marcus Schwarting & \href{https://github.com/meschw04/lk99_inference}{GitHub} \\

\hyperref[sec:hyp]{Multi-Agent Hypothesis Generation and Verification through Tree of Thoughts and Retrieval Augmented Generation} & Aleyna Beste Ozhan, Soroush Mahjoubi & \href{https://github.com/soroushmahj/ThoughtfulBeavers}{GitHub} \\

\hyperref[sec:activescience]{ActiveScience} & Min-Hsueh Chiu & \href{https://github.com/minhsueh/ActiveScience}{GitHub} \\

\hyperref[sec:g-peer-t]{G-Peer-T: LLM Probabilities For Assessing Scientific Novelty and Nonsense} & Alexander Al-Feghali, Sylvester Zhang & \href{https://github.com/alxfgh/G-Peer-T/}{GitHub} \\
\midrule

\textbf{Knowledge Extraction and Reasoning}\\

\hyperref[sec:chemqa]{ChemQA} &  Ghazal Khalighinejad, Shang Zhu, Xuefeng Liu & \href{https://github.com/ghazalkhalighinejad/multimodalQA}{GitHub} \\

\hyperref[sec:lithium-heart]{LithiumMind - Leveraging Language Models for Understanding Battery Performance} & Xinyi Ni, Zizhang Chen, Rongda Kang, Sheng-Lun Liao, Pengyu Hong, Sandeep Madireddy & \href{https://github.com/KKbeckang/LiGPT-Beta}{GitHub} \\

\hyperref[sec:knowmat]{KnowMat: Transforming Unstructured Material Science Literature into Structured Knowledge} & Hasan M. Sayeed, Ramsey Issa, Trupti Mohanty, Taylor Sparks & \href{https://github.com/sparks-sayeed/LLMs_for_Materials_and_Chemistry_Hackathon}{GitHub} \\

\hyperref[sec:ontosynthesis]{Ontosynthesis} & Qianxiang Ai, Jiaru Bai, Kevin Shen, Jennifer D'Souza, Elliot Risch & \href{https://github.com/qai222/ontosynthesis}{GitHub} \\

\hyperref[sec:kg-rag-polymers]{Knowledge Graph RAG for Polymer Simulation} & Jiale Shi, Weijie Zhang, Dandan Tang, Chi Zhang & \href{https://github.com/shijiale0609/KG-RAG-LLM-Polymers}{GitHub} \\

\hyperref[sec:hea-synthetic-data]{Synthetic Data Generation and Insightful Machine Learning for High Entropy Alloy Hydrides} & Tapashree Pradhan, Devi Dutta Biswajeet & \href{https://github.com/tapashreepradhan/LLM-materialsChem-hack24}{GitHub} \\

\hyperref[sec:chemsense]{Chemsense: Are large language models aligned with human chemical preference?} &  Martiño Ríos-García, Nawaf Alampara, Mara Schilling-Wilhelmi, Abdelrahman Ibrahim, Kevin Maik Jablonka & \href{https://github.com/lamalab-org/chemsense}{GitHub}\\

\hyperref[sec:glossagen]{GlossaGen} & Magdalena Lederbauer, Dieter Plessers, Philippe Schwaller & \href{https://github.com/mlederbauer/glossagen}{GitHub} \\
\end{longtable}

\section*{Acknowledgments}
Planning for this event was supported by NSF Awards \#2226419 and \#2209892. We would like to thank event sponsors who provided platform credits and prizes for teams, including RadicalAI, Iteratec, Reincarnate, Acceleration Consortium, and Neo4j.


\newpage

\addcontentsline{toc}{section}{Appendix} 
\part{Appendix: Individual Project Reports} 
\mtcsetdepth{parttoc}{1} 
\parttoc 
\newpage
\section{Leveraging Orbital-Based Bonding Analysis Information in LLMs}\label{sec:learning-lobsters}

\textbf{\textit{Authors: Katharina Ueltzen, Aakash Naik, Janine George
}}

LLMs were recently demonstrated to perform well for materials property prediction, especially in the low-data limit \cite{Jablonka2024a, Choudhary2024}. The Learning LOBSTERs team fine-tuned multiple Llama 3 models on textual descriptions of 1264 crystal structures to predict the highest-frequency peak in their phonon density of states (DOS) \cite{Petretto2018, Dunn2020}. This target is relevant to the thermal properties of materials, and this target dataset is part of the MatBench benchmark project \cite{Petretto2018, Dunn2020}.

The text descriptions were generated using two packages: the Robocrystallographer package \cite{Ganose2019} generates descriptions of structural features like bond lengths, coordination polyhedra, or structure type. It has recently emerged as a popular tool for materials property prediction models that require text input \cite{Sayeed2023, Rubungo2023, Moro2024}. Further, text descriptions of orbital-based bonding analyses containing information on covalent bond strengths or antibonding states were generated using LobsterPy \cite{Naik2024}. The data used here is available on Zenodo \cite{Naik2023} and was generated as part of our previous study, in which the importance of such bonding information for the same target via an RF model was demonstrated \cite{Naik2023}.

\begin{figure}[h!]
    \centering
    \includegraphics[width=0.8\textwidth]{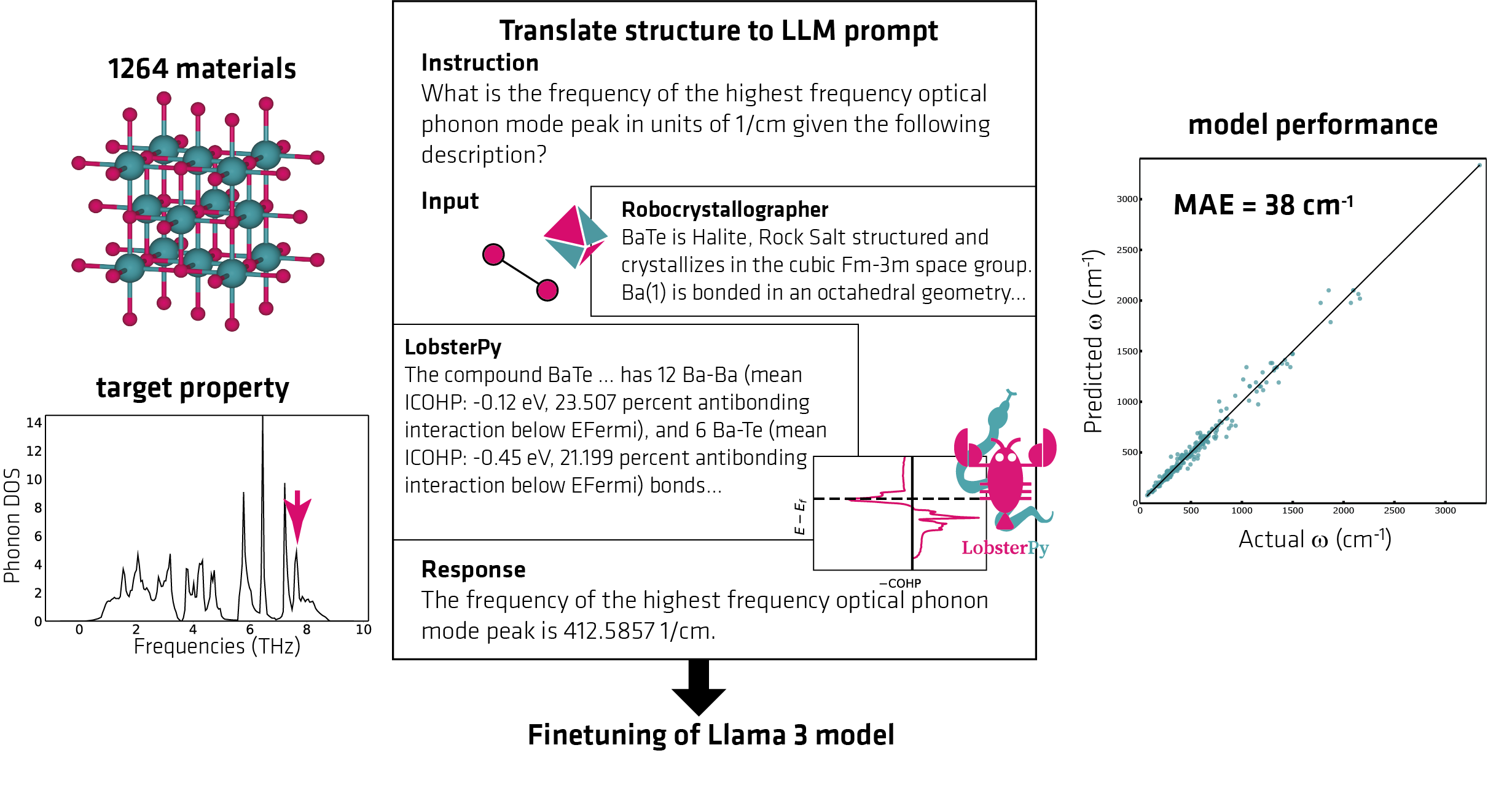} 
    \caption{Schematic depicting the prompt for fine-tuning the LLM with Alpaca prompt format.}
    \label{fig:schematic}
\end{figure}

In the hackathon, one Llama model was fine-tuned with the Alpaca prompt format using both Robocrystallographer and LobsterPy text descriptions, and another one using solely Robocrystallographer input. \autoref{fig:schematic} depicts the prompt used to fine-tune an LLM to predict the last phonon DOS peak. The train/test/validation split was 0.64/0.2/0.16. The models were trained for 10 epochs with a validation step after each epoch. The textual output was converted back into numerical frequency values for the computation of MAEs and RMSEs. Our results show that including bonding-based information improved the model's prediction. The results also corroborate our previous finding that quantum-chemical bond strengths are relevant for this particular target property \cite{Naik2023}. Both model performances (Robocrystallographer: 44 cm\textsuperscript{-1}, Robocrystallographer+LobsterPy: 38 cm\textsuperscript{-1}) are comparable to other models of the MatBench test suite, with MAEs ranging from 29 cm\textsuperscript{-1} to 68 cm\textsuperscript{-1} at time of writing \cite{Matbench2024}. However, due to the time constraints of the hackathon, no five-fold cross-validation was implemented for our model.

Although the preliminary results seem very promising, the models have not yet been exhaustively analyzed or improved. As the prediction of a numerical value and not its text embedding is of interest to our task, further model adaptation might be beneficial. For example, Rubungo et al. \cite{Rubungo2023} modified T5 \cite{Raffel2020}, an encoder-decoder model, for regression tasks by removing its decoder and adding a linear layer on top of its encoder. Halving the number of model parameters allowed them to fine-tune on longer input sequences, improving model performance \cite{Rubungo2023}.

Easy-to-use packages like Unsloth \cite{Unsloth2024} allowed us to integrate our materials data into fine-tuning an LLM for property prediction with very limited resources and time.

\newpage
\section{Context-Enhanced Material Property Prediction (CEMPP)}\label{sec:liverpool-materials}

\textbf{\textit{Authors: Federico Ottomano, Elena Patyukova, Judith Clymo, Dmytro Antypov, Chi Zhang, Aritra Roy, Piyush Ranjan Maharana, Weijie Zhang, Xuefeng Liu, Erik Bitzek
}}

\subsection{Introduction}

The Liverpool Materials team sought to improve composition-based property prediction models for novel materials by providing natural language input alongside the chemical composition of interest. In doing so, we leverage the ability of large language models (LLMs) to capture the broader context relevant to the composition. We demonstrate that enriching materials representation can be beneficial where training data is limited. Code to reproduce the experiments described below is available at \url{https://github.com/fedeotto/lpoolmat_llms}.

\begin{figure}[h!]
    \centering
    \includegraphics[width=0.8\textwidth]{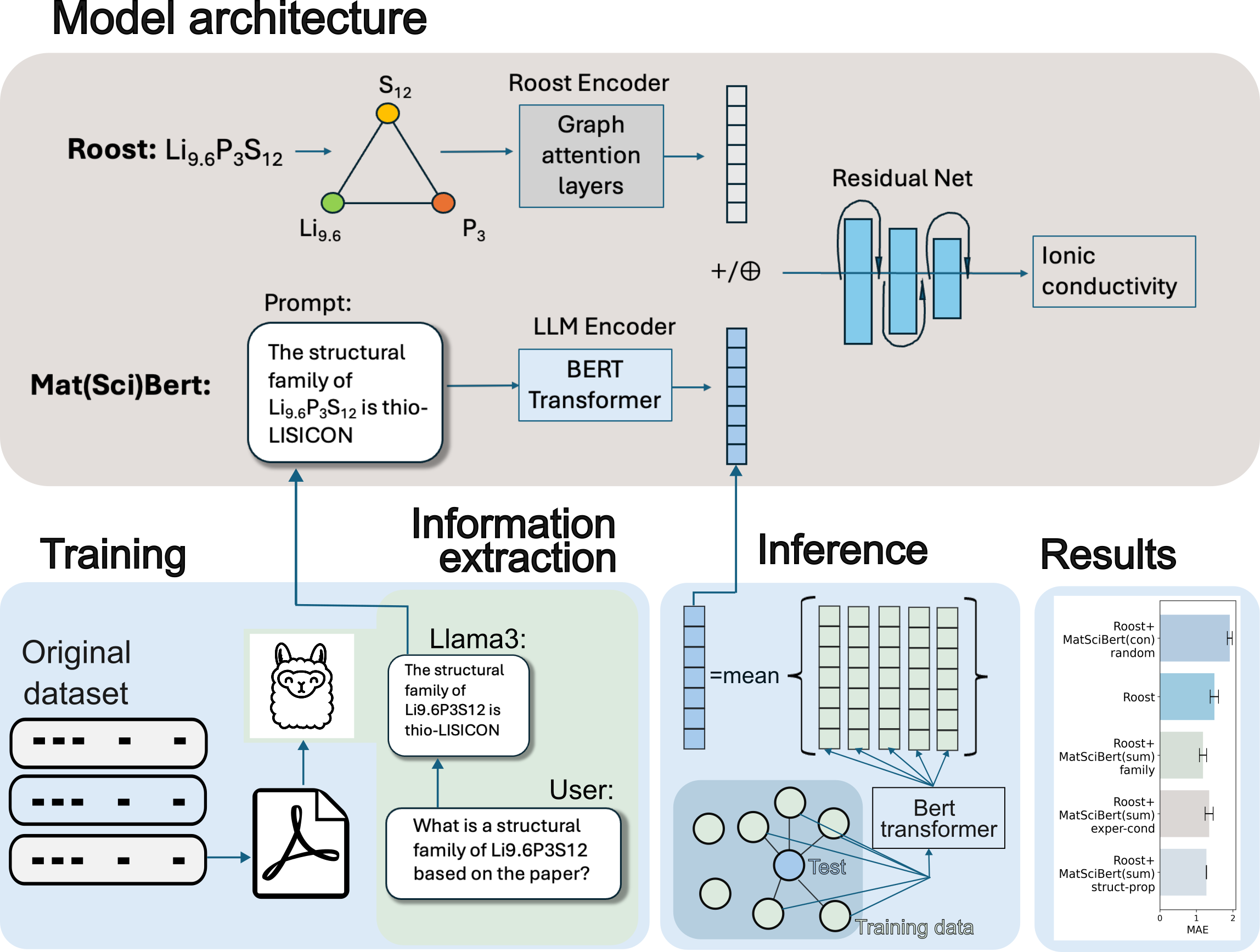} 
    \caption{Model architecture and the schema of the second experiment. Material composition is encoded with Roost encoder, additional information extracted from cited paper with Llama3, and encoded with Mat(Sci)Bert. Composition and LLM embeddings are aggregated and passed through the Residual Net projection head to predict the property. At the inference stage, the average LLM embedding from 5 nearest neighbors in composition space is taken. Results show MAE for adding different types of context (top to bottom): adding random context; not adding context; adding consistently structured context for chemical and structural family (data extracted by humans); adding automatically extracted context for experimental conditions and structure-property relationship.}
    \label{fig:context_enhanced}
\end{figure}

Our experiments are based on Roost \cite{Goodall2020}, a deep neural network for predicting inorganic material properties from their composition. Roost consists of a graph attention network that creates an embedding of the input composition (which we will enrich with context embedding), and a residual network that acts on this embedding to predict the target property value (Figure \ref{fig:context_enhanced}, top).

\subsection{Experiment 1: Using latent knowledge}

We prompt two LLMs trained on materials science literature, MatBert \cite{Trewartha2022} and MatSciBert \cite{Gupta2022a}, to directly attempt the prediction task ("What is the {property} of {material}?"). The embedding of the response and Roost’s composition embedding are aggregated (via summation or concatenation) and passed to the residual network.

We consider two datasets: matbench-perovskites \cite{matbench_perovskites} has a target property of formation energy and Li-ion conductors \cite{LiIonDatabase} dataset has a target property of ionic conductivity. The former has low stoichiometric diversity in the inputs (all entries have a general formula of ABX3) and the latter is limited in size (only 403 distinct compositions), making the prediction tasks especially challenging. We observe a 26-32\% decrease in mean absolute error (MAE) for all four settings tested (two LLMs and two aggregation schemes) in the Li-ion conductors task, and a 2-4\% decrease in MAE in the matbench-perovskites task.

\subsection{Experiment 2: Using knowledge from literature}

We obtain information from the papers reporting materials in the Li-ion conductors dataset by using PyMuPDF \cite{PyMuPDFa} parser and scanning for keywords to find relevant sections. We prompt Llama3-8b-instruct \cite{llama3modelcard} to extract and summarize specific information from the text, creating an embedding from the response (Figure \ref{fig:context_enhanced}).

Given our goal of predicting properties for compositions not yet synthesized, and therefore not referenced in any academic text, for testing and inference we average the summary embeddings of the five closest materials in the training set to the composition of interest. We define the distance between compositions using the Element Movers Distance \cite{hargreaves2020}.

We find that providing context embeddings generated from automatically extracted experimental conditions and structure-property relationships reduce MAE by 3--10\% and 10--15\% respectively. Using context embeddings based on information extracted by human experts \cite{hargreaves2022} from the same academic papers reduces MAE by 17--21\%. In contrast, using random embeddings as context increases MAE by 7--28\%. We conclude that automatically extracted context can aid property prediction for new compositions. The performance boost is less than that achieved using human-extracted and consistently structured context, which highlights both the further potential of our method and the harmful effect of noise in the automatically extracted text.

\clearpage
\newpage
\section{MolFoundation: Benchmarking Chemistry LLMs on Predictive Tasks}\label{sec:molfoundation}

\textbf{\textit{Authors: Hassan Harb, Xuefeng Liu, Anastasiia Tsymbal, Oleksandr Narykov, Dana O'Connor, Shagun Maheshwari, Stanley Lo, Archit Vasan, Zartashia Afzal, Kevin Shen
}}

\subsection{Summary}
The MolFoundation team is focused on enhancing the prediction capabilities of pre-trained large language models (LLMs) such as ChemBERTa and T5-Chem for specific molecular properties, utilizing the QM9 database. Targeting properties like dipole moment and zero-point vibrational energy (ZPVE), our approach involves retraining the last layer of these models with a selected dataset and fine-tuning the embeddings to refine accuracy. After making predictions and evaluating performance, our results indicate that T5-Chem outperforms ChemBERTa. Additionally, we found little difference between finetuned and pre-trained results, suggesting that the computationally expensive task of finetuning may be avoided.

\begin{figure}[h!]
    \centering
    \includegraphics[width=0.99\textwidth]{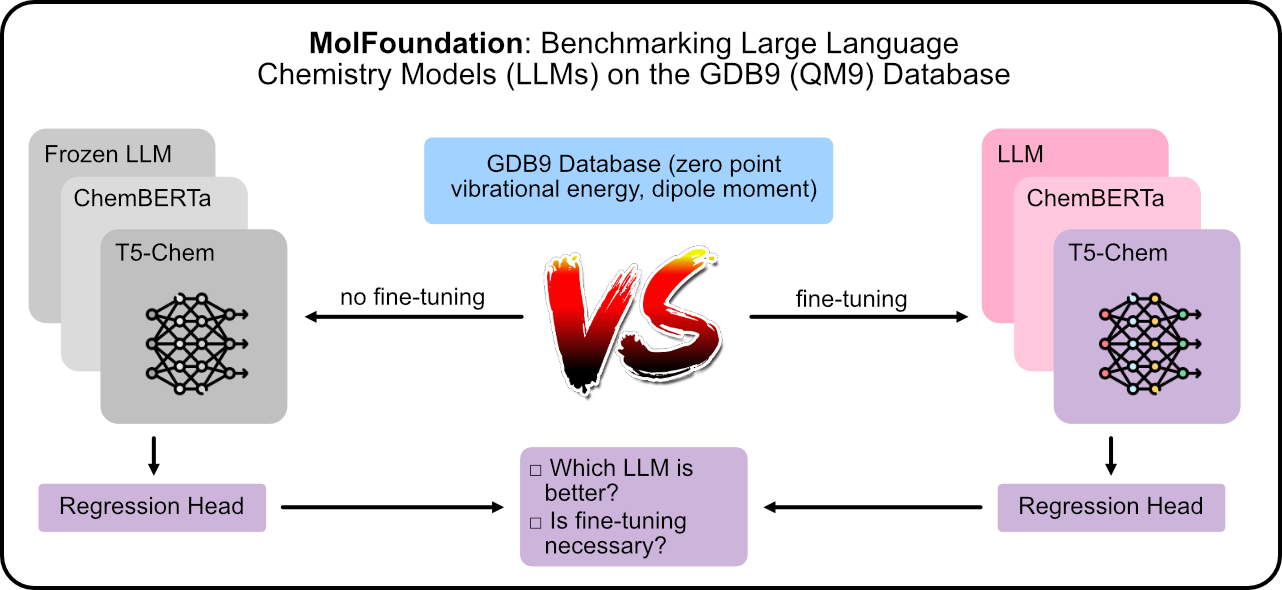} 
    \caption{
    \label{fig:molfoundation_1}}
\end{figure}

\subsection{Methods}
The models are downloaded from HuggingFace (ChemBERT and T5-Chem). Using the provided code we can tokenize our datasets (QM9). The datasets must contain SMILES and we checked that the tokenizer had all the necessary tokens for our datasets.
To make the LLMs compatible with the regression tasks in the QM9 dataset, we froze the LLM embeddings and fine-tuned on the regression layer. Training on the full LLM end-to-end is infeasible given our resources, so training on a single linear layer was much more efficient.

\begin{figure}[h!]
    \centering
    \includegraphics[width=0.99\textwidth]{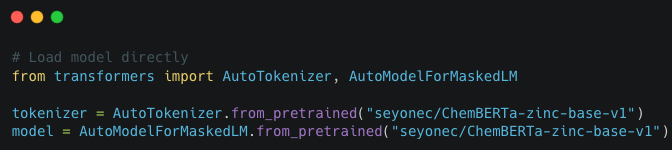} 
    \caption{
    \label{fig:molfoundation_2}}
\end{figure}

\subsection{Results}
\begin{figure}[h!]
    \centering
    \includegraphics[width=0.99\textwidth]{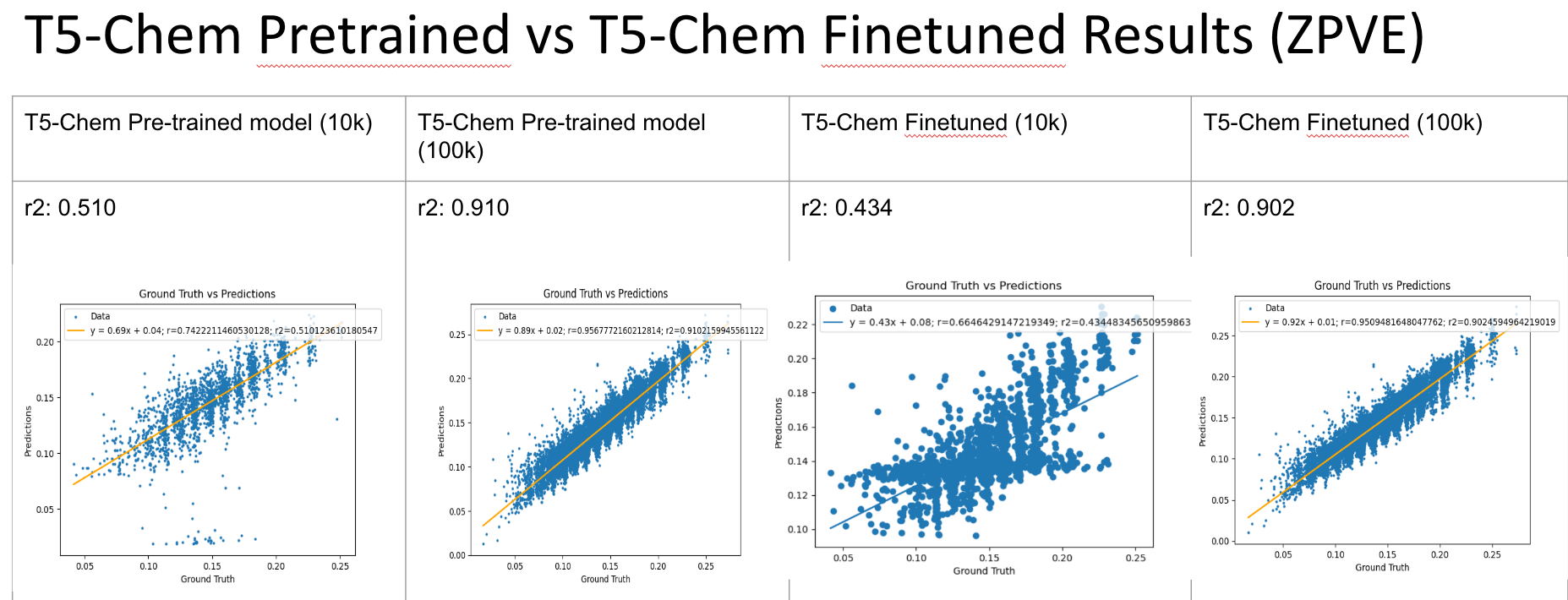} 
    \caption{ Model comparison of the pre-trained and fine-tuned T5-Chem on zero-point vibrational energy (ZPVE).
    \label{fig:molfoundation_3}}
\end{figure}

\begin{figure}[h!]
    \centering
    \includegraphics[width=0.99\textwidth]{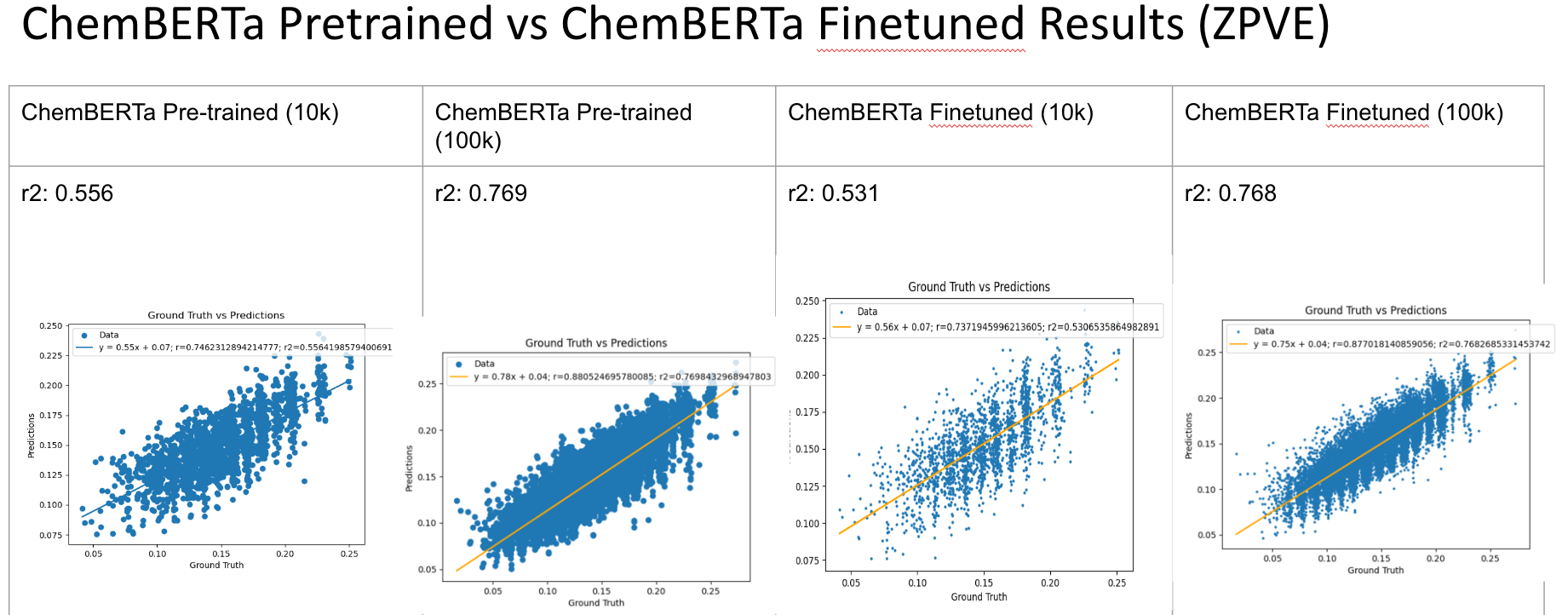} 
    \caption{Model comparison of the pre-trained and fine-tuned ChemBERTa on zero-point vibrational energy (ZPVE).
    \label{fig:molfoundation_4}}
\end{figure}

We compare the out-of-the-box (pre-trained) and fine-tuned ChemBERT and T5-Chem to predict all the molecules' zero-point vibrational energy (ZPVE) in the QM9 dataset. We hypothesized that the fine-tuned models would perform better. However, across all models, there is no significant improvement in the fine-tuned models as measured by $R^2$ (\autoref{fig:molfoundation_3}, \autoref{fig:molfoundation_4}). We noticed that the LLMs required approximately 100K datapoints to show improvements in the modeling performance, indicating a saturation regime for the models. Lastly, the T5-Chem model performs significantly better than ChemBERT.

\subsection{Transferability + Outlook}
From our experiments, it was unexpected to find that there was a lack of transfer learning from the pre-trained LLMs to the predictive tasks on the QM9 dataset. Nevertheless, we need more experimentation on different datasets, models, and fine-tuning strategies (i.e., end-to-end retraining, multi-task fine-tuning) to determine the efficacy of LLMs on chemistry prediction tasks.

MolFoundation Github: \url{https://github.com/shagunm1210/MolFoundation}

\newpage
\section{3D Molecular Feature Vectors for Large Language Models}\label{sec:geometric-geniuses}

\textbf{\textit{Authors: Jan Weinreich, Ankur K. Gupta, Amirhossein D. Naghdi, Wibe A. de Jong, Alishba Imran
}}

Link to code repo: \url{https://github.com/janweinreich/geometric-geniuses}

Direct link to tutorial: \url{https://github.com/janweinreich/geometric-geniuses/blob/main/tutorial.ipynb}

Accurate chemical property prediction is a central goal in computational chemistry and materials science. While quantum chemistry methods oﬀer high precision, they often suﬀer from computational bottlenecks. Large language models (LLMs) have shown promise as a computationally eﬃcient alternative \cite{Jablonka2023a}. However, common string-based molecular representations like SMILES and SELFIES, despite their success in LLM applications, inherently lack 3D geometric information. This limitation hinders their applicability in predicting properties for diﬀerent conformations of the same molecule - a capability essential for practical applications such as crystal structure prediction and molecular dynamics simulations. Moreover, recent studies have demonstrated that naive encoding of geometric information as numerical data can negatively impact LLM prediction accuracy \cite{Alampara2024}.

Molecular and materials property prediction typically leverages engineered feature vectors, such as those utilized in quantitative structure-activity relationship (QSAR) models. In contrast, physics-based representations, which center on molecular geometry due to their direct relevance to the Schrödinger equation, have demonstrated eﬃcacy in various deep learning architectures \cite{Batzner2022, Gupta2022b, Kaszuba2024}. This research investigates new strategies for encoding 3D molecular geometry for LLMs. We hypothesize that augmenting the simple SMILES representation with geometric features could enable the integration of complementary information from both modalities, ultimately enhancing the predictive power of LLMs in the context of molecular properties.

\begin{figure}[h!]
    \centering
    \includegraphics[width=0.99\textwidth]{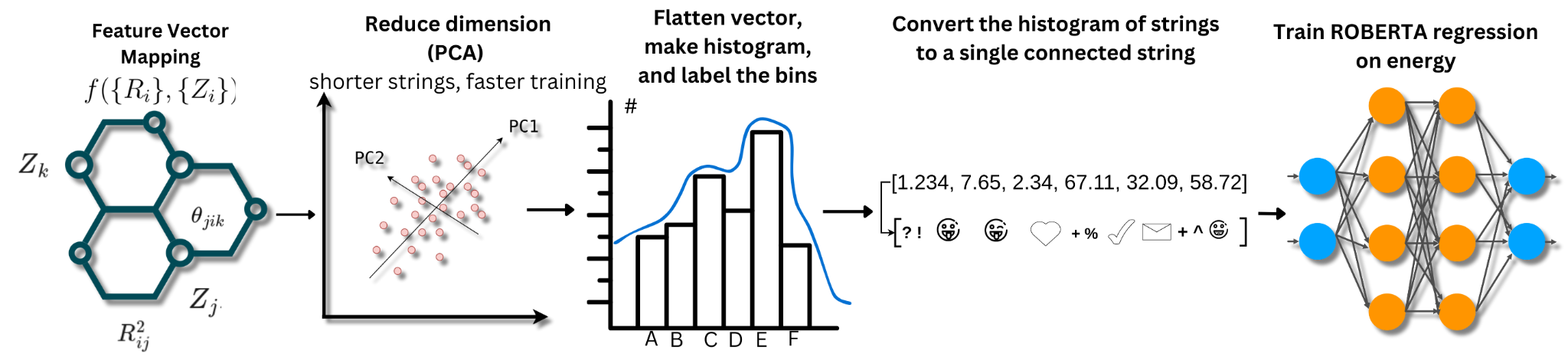} 
    \caption{Schematic representation of the training process for a regression model illustrating the novel string-based encoding of 3D molecular geometry for LLM-based energy prediction. The workﬂow involves (1) Computation of high-dimensional feature vectors representing the 3D molecular geometry of each conformer. (2) Dimensionality reduction to obtain a more compact representation. (3) Conversion of the reduced vectors into a histogram, where unique string characters are assigned to each bin. (4) Input of the resulting set of strings (one per conformer) into an LLM for energy prediction.
    \label{fig:GeomG1}}
\end{figure}

To achieve the integration of geometric information, we begin by computing a high-dimensional feature vector for each molecular geometry. While any representation capable of faithfully encoding the 3D structure of a molecule could, in principle, be utilized, this work speciﬁcally employs a previously established geometry-dependent representation \cite{Khan2023} based on many-body expansions of distances and angles. As a benchmark dataset, we utilize a diverse set of ethanol and benzene structures generated using molecular dynamics calculations utilizing density functional theory (DFT) \cite{Chmiela2023, Bowman2022}. The target property we aim to predict is the absolute energy of each conformer. The energy scale is shifted such that the lowest energy conformer within the dataset is assigned an energy value of zero. We transform our high-dimensional geometric feature vectors into string representations suitable for LLM training. First, we apply principal component analysis (PCA) to reduce the dimensionality of the vectors. This step is crucial for generating compact string representations that can be eﬃciently processed by LLMs. Next, we compute histograms of the reduced vectors, ensuring consistent binning intervals across all dimensions. Each bin is uniquely labeled with a distinct character, and negative values are preﬁxed with a special character, following the approach of \cite{Weinreich2023}. Finally, we count the occurrences of values within each bin, eﬀectively converting the numerical vectors into strings. These string representations serve as the input for training a RoBERTa regression model.

\begin{figure}[h!]
    \centering
    \includegraphics[width=0.49\textwidth]{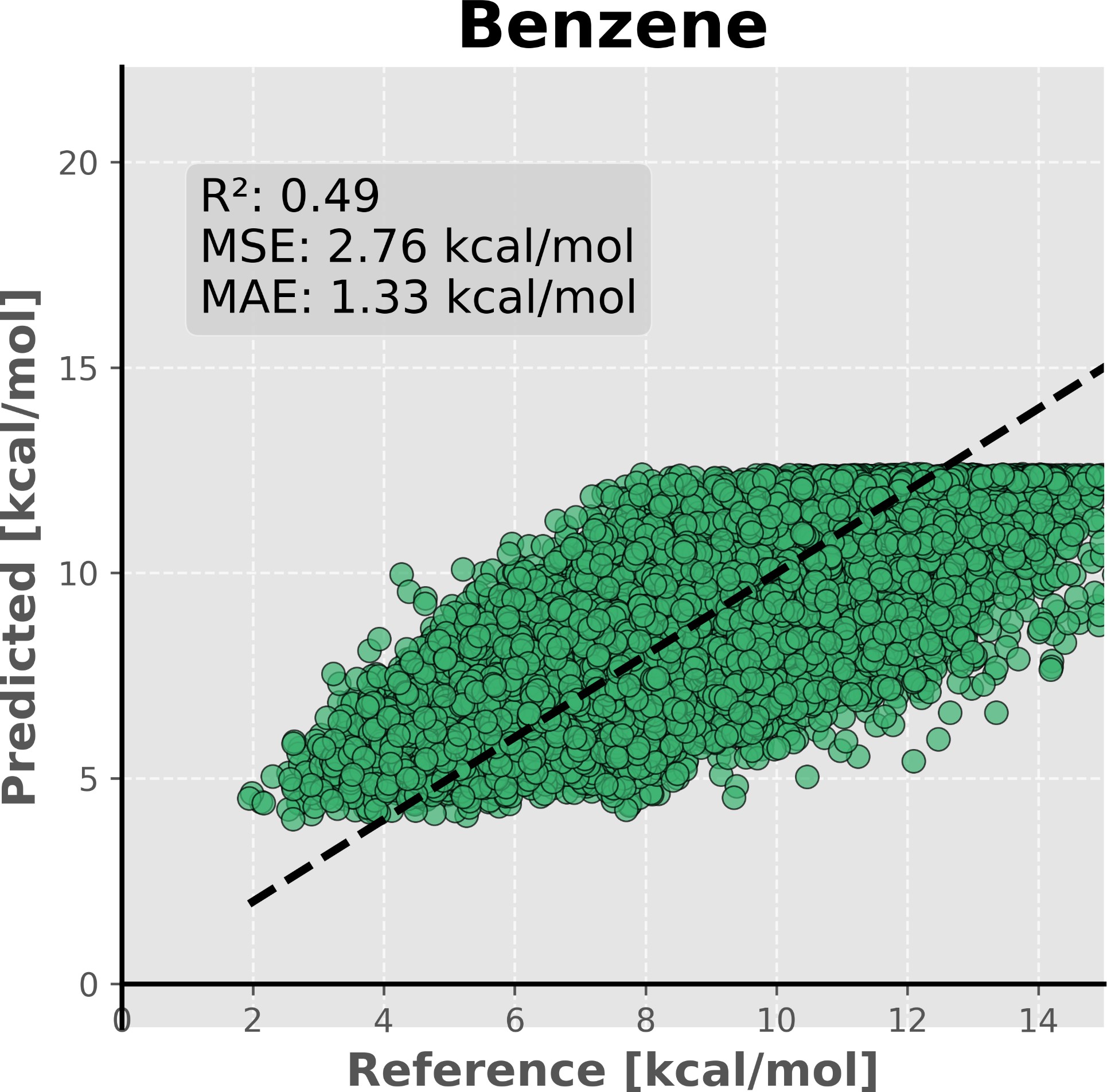} 
    \includegraphics[width=0.49\textwidth]{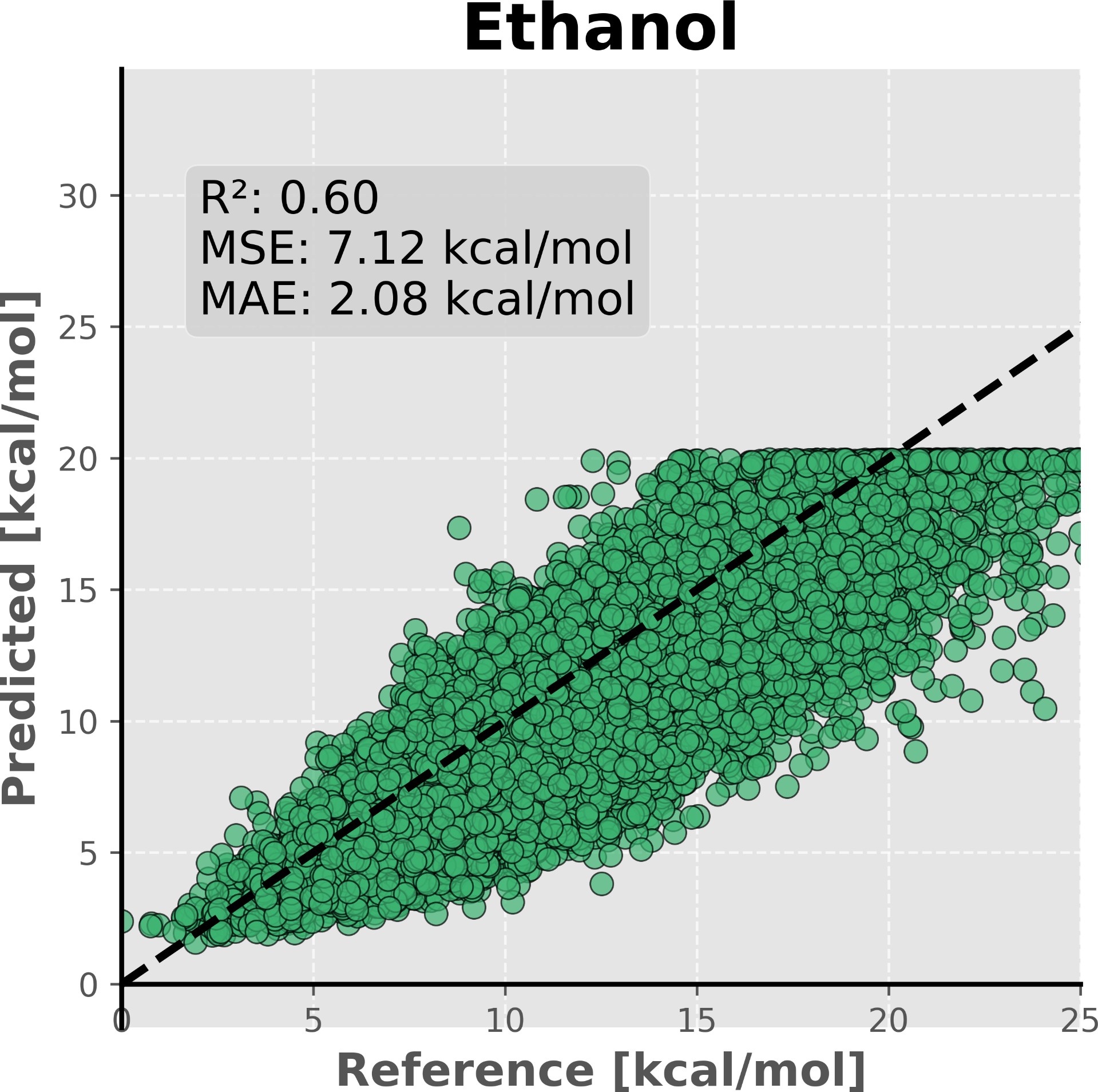} 
    \caption{Performance of an LLM in predicting total energies of benzene and ethanol structures, where the model was trained on a large dataset of MD-generated conﬁgurations. Each point represents a different molecular structure sampled from MD simulations \cite{Chmiela2023}.
    \label{fig:GeomG2}}
\end{figure}

The ﬁnal step involves randomly partitioning the dataset into 80\% training and 20\% testing sets. The string-based representations are then employed to train a RoBERTa model augmented with a regression head. In \autoref{fig:GeomG2}, we showcase scatter plots illustrating the predicted versus true total energies for benzene (trained for 4 epochs) and ethanol (trained for 20 epochs) using a dataset of 80,000 molecular conﬁgurations for each molecule. Our results do not yet attain the accuracy levels of state-of-the-art equivariant neural networks like MACE, which reports a mean absolute error (MAE) of 0.009 kcal/mol for benzene \cite{Batatia2022}. Nonetheless, it is important to underscore that this represents a novel capability for LLMs, which were previously unable to process and predict properties of 3D molecular structures only

diﬀering by bond rotations. This initial investigation paves the way for advancements through the exploration of alternative string encoding schemes of numerical vectors in combination with larger LLMs.

\subsection*{Acknowledgements}

A.K.G. and W.A.D. acknowledge funding for this project from the U.S. Department of Energy (DOE), Office of Science, Office of Basic Energy Sciences, through the Rare Earth Project in the Separations Program at Lawrence Berkeley National Laboratory under Contract DE-AC02-05CH11231.

J.W. thanks EPFL for computational resources and NCCR Catalysis (grant number 180544), a National Centre of Competence in Research funded by the Swiss National Science Foundation for funding as well as the Laboratory for Computational Molecular Design.

\newpage
\section{LLMSpectrometry}\label{sec:LLMSpectrometry}

\textbf{\textit{Authors: Tyler Josephson, Fariha Agbere, Kevin Ishimwe, Colin Jones, Charishma Puli, Samiha Sharlin, Hao Liu
}}

\subsection{Introduction}

Nuclear Magnetic Resonance (NMR) spectroscopy is a chemical characterization technique that uses oscillating magnetic fields to characterize the structure of molecules. Different atoms in molecules resonate at different frequencies based on their local chemical environments. The resulting NMR spectrum can be used to infer interactions between particular atoms in a molecule and determine the molecule’s entire structure. Solving NMR spectral tasks is critical, as multiple aspects, such as the number, intensity, and shape of signals, as well as chemical shifts, need to be considered.

Machine learning tools, including the Molecular Transformer \cite{MolecularTransformer}, have been used to learn a function to map spectrum to structure, but these require thousands to millions of labeled examples \cite{NMRTransformer}, far more than what humans typically encounter when learning to assign spectra. In contrast, we recognize NMR spectral tasks as being fundamentally about multi-step reasoning, and we aim to explore the reasoning capabilities of large language models (LLMs) for solving this task.

The project aims to investigate the capabilities of Large Language Models (LLMs), specifically GPT-4, for NMR structure determination. In particular, we were interested in evaluating whether GPT-4 could use chain-of-thought reasoning \cite{ChainOfThought} with a scratchpad to evaluate the components of the spectra and synthesize an answer in the form of a molecule. Interpreting NMR data is crucial for organic chemists; an AI-assisted tool could benefit the pharmaceutical or food industry, as well as forensic science, medicine, research, and teaching.

\subsection{Method}

\begin{figure}[h!]
    \centering
    \includegraphics[width=0.8\textwidth]{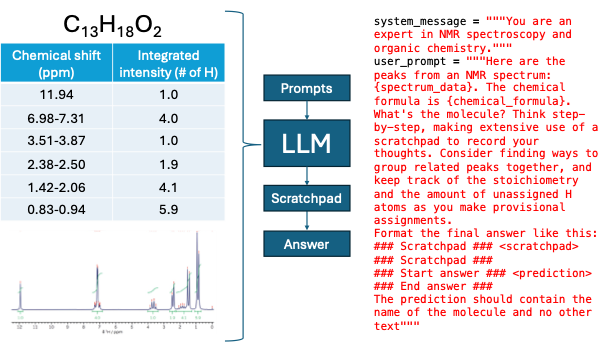} 
    \caption{Scheme of the system. Data is first converted into text format, with peak positions and intensities represented as (x,y) pairs. These are passed into an LLM prompt, which is tasked to use a scratchpad as it reasons about the data and the formula, before providing a final answer.}
    \label{fig:scheme}
\end{figure}

We manually gathered 19 experimental \textsuperscript{1}H NMR data from the Spectral Database for Organic Compounds (SDBS) website \cite{SDBS}. We then combined Python scripts, LangChain, and API calls to GPT-4 to automate structure elucidation. The components of the model are shown in \autoref{fig:scheme}. First, NMR peak data and the chemical formula were formatted as text and inserted into a prompt. This prompt instructs the LLM to reason about the data
step-by-step while using a scratchpad to record its “thoughts,” then report its answer according to an output template. We then used Python to parse the output and compare the answer to the true answer by matching with chemical names from the NIST Chemistry WebBook.

\subsection{Results}

Our 2024 LLM Hackathon for Applications in Materials and Chemistry (May 2024) results found GPT-4 successful in just 3 of the 19 NMR datasets. The scratchpad reveals how GPT-4 follows the prompt and systematically approaches the problem. It analyzes the molecular formula and carefully reads peak values and intensity to predict the molecule. The model correctly identified nonanal, ethanol, and acetic acid - 3 relatively simple molecules with few structural isomers.
Incorrect answers included more complex molecules, with significant branching, many functional groups, and aromatic rings, leading to more structural isomers consistent with the chemical formula.

\subsection{Conclusions and Outlook}

NMR spectral tasks challenge students to develop complex problem solving skills, and these also prove to be difficult in a zero-shot chain-of-thought prompting setting for GPT-4, with only 3/19 spectra solved correctly. We noticed interesting patterns of (apparent) reasoning, and we speculate significant performance improvements are possible. Prompting strategies can be explored more systematically (including few-shot and many-shot approaches), as well as embedding the LLM’s answers inside an iterative loop, so it can self-correct simple mistakes such as generating a molecule with an incorrect chemical formula.

Since the hackathon, we identified a better benchmark dataset: NMR-Challenge.com [3], an interactive website with exercises categorized as Easy, Medium, and Hard. They also have data on human performance, which can enable comparison of GPT-4 to humans. Further analysis of proximity of incorrect answers to correct answers would provide more granular information about the performance of the AI, for example, an aromatic molecule with correct substituents in incorrect locations is further from the right answer than a molecule with incorrect substituents.
We think this could form a useful addition to existing LLM benchmarks, for evaluating chemistry knowledge intertwined with complex multistep reasoning.

Datasets, code, and results are available at: \url{https://github.com/ATOMSLab/LLMSpectroscopy}

\newpage
\section{MC-Peptide: An Agentic Workflow for Data-Driven Design of Macrocyclic Peptides}\label{sec:mc-peptide}

\textbf{\textit{Authors: Andres M. Bran, Anna Borisova, Marcel M. Calderon, Mark Tropin, Rob Mills, Philippe Schwaller
}}

\subsection{Introduction}
Macrocyclic peptides (MCPs) are a class of compounds composed of cyclized chains of amino acids forming a macrocyclic structure. They are promising for having improved binding affinity, specificity, proteolytic stability, and enhanced membrane permeability compared to linear peptides \cite{Ji2024}. Their unique properties make them highly suitable for drug development, enabling the creation of therapeutics that can address currently unmet medical needs \cite{Ji2024}. Indeed, it has been shown that MCPs of up to 12 amino acids show great promise for permeating cellular membranes \cite{Merz2024}. Despite the more constrained chemical space offered by this class of molecules, their design remains a challenging issue due to the vast space of amino acid combinations.

One important parameter of MCPs is permeability, which determines how well the structure can permeate into cells, making it a relevant factor in assessing the MCP’s pharmacological activity. However, data for MCPs and their permeabilities is scattered across scientific papers that report data without consensus on reporting form, making it challenging for systems to compile and use this raw data.

Here, we introduce MC-Peptide, an LLM-based agentic workflow created for tackling this issue in an end-to-end fashion. As shown in Figure 1, MC-Peptide’s main goal is to produce suggestions of novel MCPs, following from a reasoning process involving: (i) understanding of the design objective, (ii) gathering of relevant scientific information, (iii) data extraction from papers, and (iv) inference based on the extracted information. MC-Peptide leverages advances in LLMs such as semantic search, grammar-constrained generation, and agentic architectures, ultimately yielding candidate MCP variants with enhanced predicted permeability in comparison to reference designs.

\subsection{Methods}
\begin{figure}[h!]
    \centering
    \includegraphics[width=0.8\textwidth]{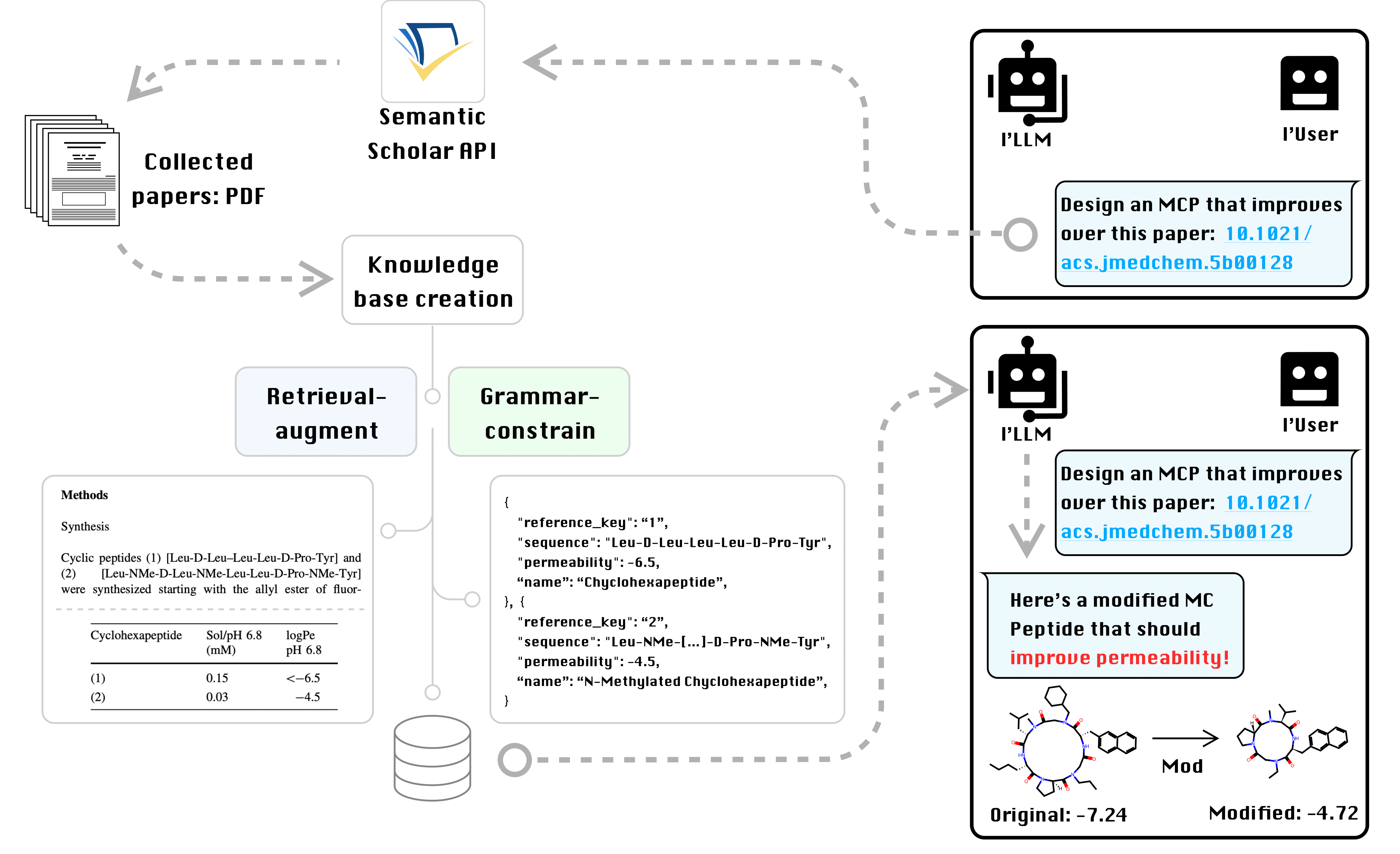} 
    \caption{MC-Peptide: Pipeline implemented in this work. The example illustrates a user request, followed by retrieval from the Semantic Scholar API, and the creation of a knowledge base. MCPs and permeabilities are extracted from \cite{Lewis2015}. The pipeline finishes with an LLM proposing modifications to an MCP, increasing its permeability.}
    \label{fig:mc-peptide}
\end{figure}

We implemented the basic building blocks for the pipeline described in \autoref{fig:mc-peptide}, with the most important components being document retrieval, structured data extraction, and in-context learning for design.

\paragraph{Document Retrieval}
An important part of this pipeline is the retrieval of relevant documents. As shown in \autoref{fig:mc-peptide}, the pipeline developed here does not rely on a stand-alone predictive system, but rather aims to leverage existing knowledge in the scientific literature, in an on-demand fashion. This is known as retrieval-augmented generation (RAG) \cite{Lewis2020a}, and one of its key components is a relevance-based retrieval system. By integrating the Semantic Scholar API, a service that provides access to AI-based search of scientific documents, the pipeline is able to create and refine a structured knowledge base from papers.

\paragraph{Structured Data Extraction}
To employ the unstructured data from the previous step, a retrieval-augmented system was designed that retrieves sections of papers and uses them as context for the generation of structured data objects with predefined grammars, which can be used to constrain the decoding of LLMs \cite{Beurer2024}, ensuring that the resulting output follows the same structure and format. This technique also mitigates hallucination, as it prevents the LLM from generating unnecessary and potentially misleading information \cite{Bechard2024}.

\paragraph{In-Context Learning}
The extracted data is then leveraged through the in-context learning capabilities of LLMs \cite{Dong2022}, allowing the models to learn from few data points given as context in the prompt and generate output based on that. This capability has been extensively explored elsewhere \cite{Agarwal2024}. Here we show that, for the specific task of modifying MCPs to improve permeability, LLMs perform well, as assessed by a surrogate random forest model.

\subsection{Conclusions}
We present MC-Peptide, a novel agentic workflow for designing macrocyclic peptides. The system is built from a few main components: data collection, peptide extraction, and peptide generation. We evaluate the peptide permeability of newly generated peptides with respect to the initial structures found in reference articles.

The resulting system shows that LLMs can be successfully leveraged for automating multiple aspects of peptide design, yielding an end-to-end generative tool that researchers can utilize to accelerate and enhance their experimental workflows. Furthermore, this workflow can be extended by adding more specialized modules (e.g., for increasing the diversity of information sources). The modular design of MC-Peptide ensures extendability to more design objectives and input sources, as well as other domains where data is reported in an unstructured fashion in papers, such as materials science and organic chemistry.

The code for this project has been made available at: \url{https://github.com/doncamilom/mc-peptide}.

\newpage
\section{Leveraging AI Agents for Designing Low Band Gap Metal-Organic Frameworks}\label{sec:porevoyant}

\textbf{\textit{Authors: Sartaaj Khan, Mahyar Rajabi, Amro Aswad, Seyed Mohamad Moosavi, Mehrad Ansari
}}

\subsection{Introduction}

\begin{figure}[h!]
    \centering
    \includegraphics[width=0.8\textwidth]{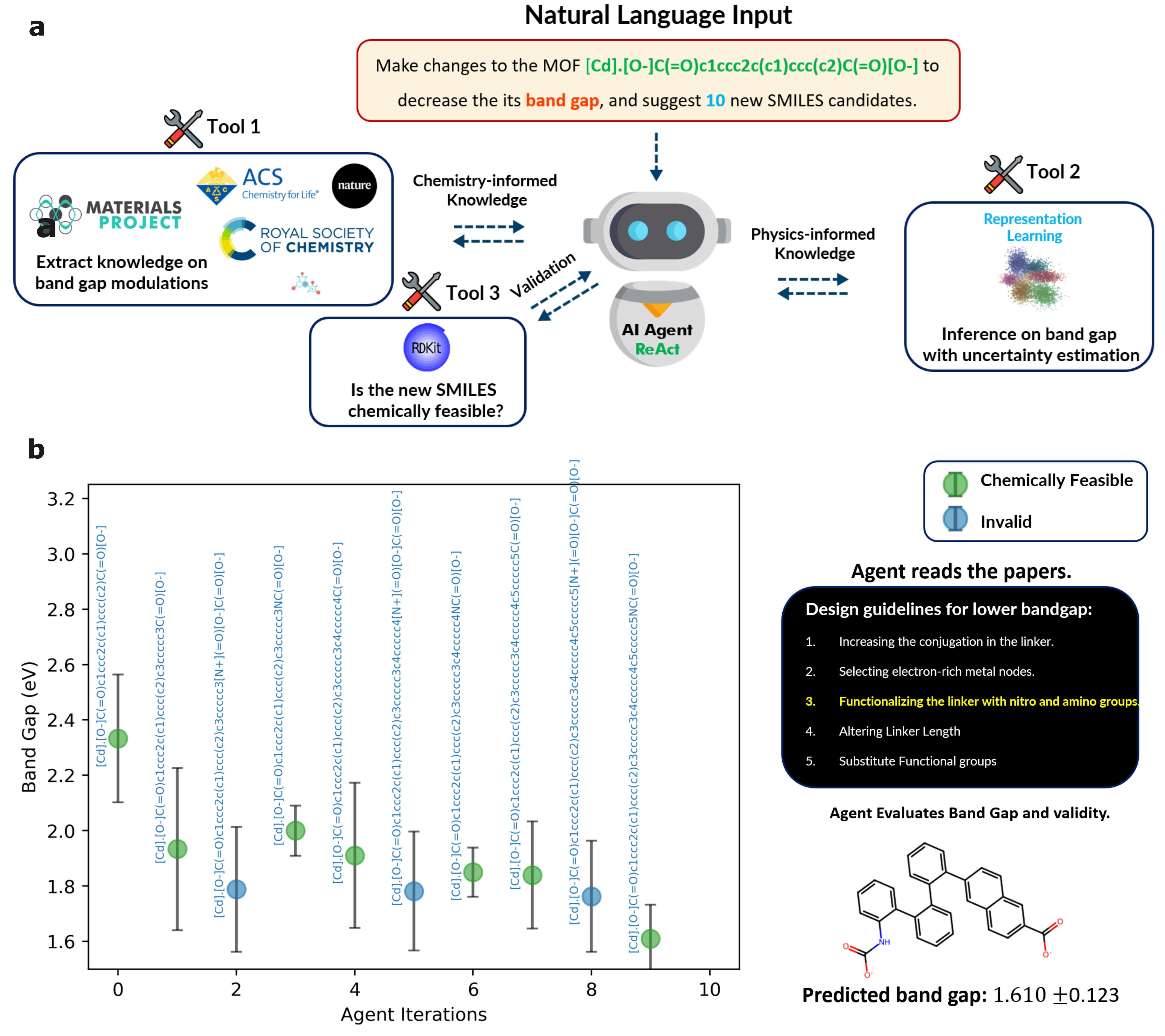} 
    \caption{a) Workflow overview. The ReAct agent looks up guidelines for designing low band gap MOFs from research papers and suggests a new MOF (likely with lower band gap). It then checks validity of the new SMILES candidate and predicts band gap with uncertainty estimation using an ensemble of surrogate fine-tuned MOFormers. b) Band gap predictions for new MOF candidates as a function of agent iterations. Detailed analysis of this methodology applied to other materials and target properties can be found in reference~\cite{Ansari2024}.}
    \label{fig:workflow}
\end{figure}

Metal-organic frameworks (MOFs) are known to be excellent candidates for electrocatalysis due to their large surface area, high adsorption capacity at low CO\textsubscript{2} concentrations, and the ability to fine-tune the spatial arrangement of active sites within their crystalline structure \cite{Li2022a}. Low band gap MOFs are crucial as they efficiently absorb visible light and exhibit higher electrical conductivity, making them suitable for photocatalysis, solar energy conversion, sensors, and optoelectronics. In this work, we aim at using chemistry-informed ReAct \cite{Yao2022} AI Agents to optimize the band gap property of MOFs. The overview of the workflow is presented in \autoref{fig:workflow}a. The agent inputs a textual representation of the initial MOF structure as a SMILES (Simplified Molecular Input Line-Entry System) string representation, and a short description of the property optimization task (i.e., reducing band gap), all in natural language text. This is followed by an iterative closed-loop suggestion of new MOF candidates with a lower band gap with uncertainty assessment, by making adjustments to the initial MOF given a set of design guidelines automatically obtained from the scientific literature.
Detailed analysis of this methodology applied to other materials and target properties can be found in reference~\cite{Ansari2024}.

\subsection{Agent and Tools}

The agent, powered by a large language model (LLM), is augmented with a set of tools allowing for a more chemistry-informed decision-making. These tools are as follows:

\begin{enumerate}
    \item
\textbf{Retrieval-Augmented Generation (RAG):} This tool allows the agent to obtain design guidelines on how to adapt the MOF structure from unstructured text. In specific, the agent has access to a fixed set of seven MOF research papers (see Refs. \cite{Usman2017}-\cite{Lin2012}) as PDFs. This tool is designed to extract the most relevant sentences from papers in response to a given query. It works by embedding both the paper and the query into numerical vectors, then identifying the top k passages within the document that either explicitly mention or implicitly suggest the adaptations to the band gap property for a MOF. The embedding model is OpenAI’s text-ada-002 \cite{Greene2022}. Inspired by our earlier work \cite{Ansari2023}, $k$ is set to 9 but is dynamically adjusted based on the relevant context’s length to avoid OpenAI’s token limitation error.

\item
\textbf{Surrogate Band Gap Predictor:} The surrogate model used is a transformer (MOFormer \cite{Cao2023}) that inputs the MOF as SMILES. This model is pre-trained using a self-supervised learning technique known as Barlow-Twin \cite{Zbontar2021}, where representation learning is done against structure-based embeddings from a crystal graph convolutional neural network (CGCNN) \cite{Xie2018}. This was done against 16,000 BW20K entries \cite{Moosavi2020}. The pre-trained weights are then transferred and fine-tuned to predict the band gap labels taken from 7450 entries from the QMOF database \cite{Rosen2021}. From a 5-fold training, an ensemble of five transformers are trained to return the mean band gap and the standard deviation, which is used to assess uncertainty for predictions. For comparison, our transformer’s mean absolute error (MAE) is approximately 0.467, whereas MOFormer, which was pre-trained on 400,000 entries, achieves an MAE of approximately 0.387.

\item
\textbf{Chemical Feasibility Evaluator:} This tool primarily uses RDKit \cite{Landrum2013} to convert a SMILES string into an RDKit Mol object, and performs several validation steps to ensure chemical feasibility. First, it parses the SMILES string to confirm correct syntax. Next, it validates the atoms and bonds, ensuring they are chemically valid and recognized. It then checks atomic valences to ensure each atom forms a reasonable number of bonds. For ring structures, RDKit verifies the correct ring closure notation. Additionally, it adds implicit hydrogens to satisfy valence requirements and detects aromatic systems, marking relevant atoms and bonds as aromatic. These steps collectively ensure the molecule’s basic chemical validity.
\end{enumerate}

We use OpenAI’s GPT-4 \cite{OpenAI2023a} with a temperature of 0.1 as our LLM and LangChain \cite{Chase2022} for the application framework development (note the choice of LLM is only a hyperparameter and other LLMs can be also used with the agent).

The new MOF candidates and their corresponding inferred band gap are represented in Figure 1.b. The agent starts by retrieving the following design guidelines for low band gap MOFs from research papers: 
1) Increasing the conjugation in the linker. 
2) Selecting electron-rich metal nodes. 
3) Functionalizing the linker with nitro and amino groups.
4) Altering linker length.
5) Substitute functional groups (i.e., substituting hydrogen with electron-donating groups on the organic linker). 
Note that the metal node adaptations were restrained by simply changing the system input prompt. The agent iteratively implements the above strategies and makes changes to the MOF. After each modification, the band gap of the new MOF is assessed using the fine-tuned surrogate MOFormers to ensure a lower band gap. Subsequently, the chemical feasibility is evaluated. If the new MOF candidate has an invalid SMILES string or a higher band gap, the agent reverts to the most recent valid MOF candidate with the lowest band gap.

\subsection{Data and Code Availability}

All code and data used to produce results in this study are publicly available in the following GitHub repository: \url{https://github.com/mehradans92/PoreVoyant}.

\clearpage
\newpage
\section{How Low Can You Go? Leveraging Small LLMs for Material Design}\label{sec:llama3-small-organics}

\textbf{\textit{Authors: Alessandro Canalicchio, Alexander Moßhammer, Tehseen Rug, Christoph Völker
}}

\subsection{Motivation - Leveraging LLMs for Sustainable Construction Material Design}
The construction industry's dependence on limited resources and the high emissions associated with traditional building materials such as Portland cement-based concrete necessitate a transition to more sustainable alternatives \cite{Environment2018}. Geopolymers synthesized from industrial by-products such as fly ash and waste slag represent a promising solution \cite{Provis2015, Gökçe2020, He2013}. However, scaling up these highly complex materials to meet market demand is a major challenge due to the smaller volumes available at diverse sources. Traditionally, it takes years of intensive scientific research to bring a new material to market. To meet the demand for sustainable alternative materials, new resource streams need to be developed much more frequently, making the traditional lengthy process of acquiring specialized material knowledge impractical.

This is where LLMs offer a solution. LLMs are trained on web-scale data and contain deep domain expertise that was previously hidden in scientific institutions but is now accessible to everyone. This knowledge is stored in the LLMs inner representations (hidden states) and is accessible through prompts making best practices from specialized domains easily available. In addition, LLMs can provide information in any format, making it easy to convert complex scientific concepts into workable formulations for direct application in the lab. 

Previous systematic benchmarks have shown that LLMs can keep up with conventional approaches such as Bayesian Optimization (BO) and Sequential Learning (SL) when it comes to designing sustainable alternatives to concrete, so-called alkali-activated concrete \cite{Völker2023}. What sets these models apart from classical design methods is their ability to produce zero-shot designs, meaning they can propose relatively well-functioning formulations without any initial training data. This capability suggests that LLMs can play a crucial role in the initial data collection phase. Essentially, it allows researchers entering a new field to begin new projects at an expert level from the outset, leading to much faster development of viable solutions.

\subsection{Research Gap }
The emergence of a new generation of high-performing, smaller-scale LLMs is expanding the boundaries of feasible application scenarios. Deploying these compact LLMs enables applications in sensitive areas like R\&D. Full control over a local model enhances quality control and repeatability by allowing precise management of versions and system prompts. Additionally, they increase security and data privacy through in-house processing, which reduces the risk of confidential data leakage and other security threats. Moreover, small-scale LLMs improve cost efficiency and reliability by minimizing network dependency and operational costs. This development raises an intriguing research question:

Can current small LLMs retrieve and utilize valuable knowledge from their inner representations, making them effective tools for closed lab networks in material science? 

Investigating this could significantly advance LLM adoption in material science R\&D, especially in sensitive industry settings, potentially transforming experimental workflows and accelerating innovation.

\subsection{Methodology}
To investigate this question, we deployed two small-scale instruction LLMs on consumer hardware, specifically Llama 3 8B and Phi-3 3.8B, both quantized to four bits. Our goal was to evaluate their ability to design alkali-activated concrete with high compressive strength. The LLMs were tasked via system messages to determine four mixture parameters from a predetermined grid of 240 recipes: 1) the blending ratio of fly ash to ground granulated blast furnace slag, 2) the water-to-powder ratio, 3) the powder content, and 4) the curing method. We measured their performance by comparing the compressive strengths of the suggested formulations to previously published lab results \cite{Rao2018}.

\begin{figure}[h!]
    \centering
    \includegraphics[width=0.99\textwidth]{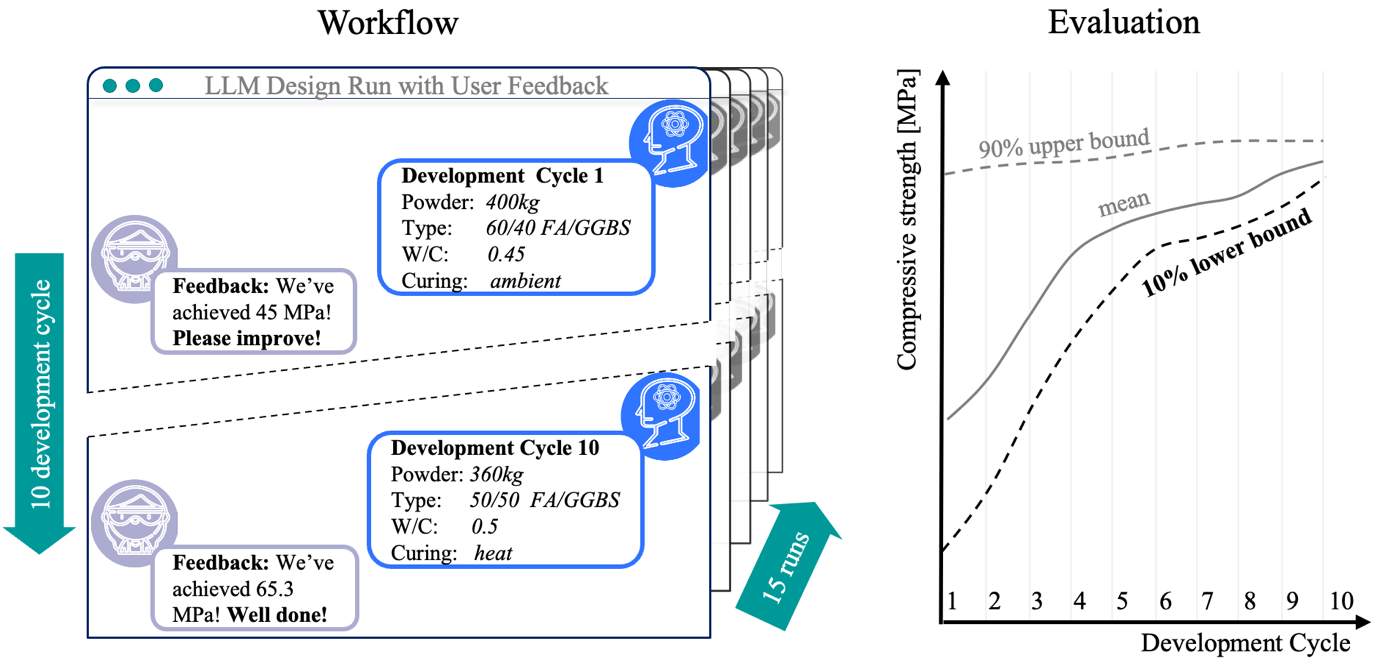} 
    \caption{LLM-based material design workflow (left) and diagram showing evaluation metric (right).
    \label{fig:llama3-smal-orgnics}}
\end{figure}

The workflow and evaluation are shown in \autoref{fig:llama3-smal-orgnics}. Each design run comprised 10 consecutive development cycles, where the LLM suggested a formulation and the user provided feedback. This process was repeated five times for statistical analysis. Additionally, the prompt was rephrased synonymously three times to increase linguistic variability. Two types of contexts were provided: one with detailed design instructions, such as ``reducing the water-to-powder ratio leads to higher strengths,'' and one without additional instructions. In total, 15 design runs were conducted per model and per context. Finally, we assessed the achieved 10\% lower-bound strength, defined as the strength achieved in 90\% of the design runs, and compared this against a random draw as the statistical baseline and SL. 
\subsection{Results and Conclusion }
The results, summarized in \autoref{tab:llama3_material_scientist}, showed that all investigated models generated designs that outperformed the statistical baseline of 28 MPa in the first round and 52 MPa in the final round, demonstrating the surprising effectiveness of small-scale LLMs in materials design. An exception was noted with the Phi-3 model when provided with extensive design context: it produced the same output in each round and failed to produce viable solutions completely after the fifth development cycle, indicating its failure to understand the task. As expected, Llama 3 outperformed the smaller Phi-3 model. Specifically, Llama 3 benefited from additional design context, showing an improvement of more than 10 MPa in the initial rounds.

\begin{table}[h]
\caption{Achieved compressive strength of designs suggested by LLMs. Statistical assessment in terms of 10\% lower bound strength, i.e., the 10\% worst cases.}
\label{tab:llama3_material_scientist}
\begin{tabular}{|l|c|c|}
\hline
\textbf{Model} & \multicolumn{1}{l|}{\textbf{Development Cycle 1}} & \multicolumn{1}{l|}{\textbf{Development cycle 10}} \\ \hline
Random Baseline & 28 MPa & 52 MPa \\ \hline
Phi 3 (No design rules in context & 50 MPa & 58 MPa \\ \hline
Phi 3 (Design rules in context) & 51 MPa & -- \\ \hline
Llama 3 (No design rules in context) & 48 MPa & 60 MPa \\ \hline
Llama 3 (Design rules in context) & 59 MPa & 60 MPa \\ \hline
\end{tabular}
\end{table}

In conclusion, small-scale LLMs performed surprisingly well, with Phi-3 producing results significantly above a random guess, though it faced challenges with more complex prompts. The effectiveness of LLMs in solving design tasks depends on how well material concepts are represented in their hidden states and how effectively these can be retrieved via prompts, giving larger models an advantage. Despite their smaller parameter count and less training data, Phi-3 and Llama 3 demonstrated common sense for domain-specific design tasks, making local deployment a viable option. While 100\% reliability in retrieving sensible information via LLMs is uncertain, small-scale LLMs can generate educated guesses that potentially accelerate the familiarization process with novel materials.

\subsection{Code}
The code used to conduct the experiments is open-source and available here: \url{https://github.com/sandrocan/LLMs-for-design-of-alkali-activated-concrete-formulations}.

\newpage
\section{LangSim}\label{sec:langsim}

\textbf{\textit{Authors: Yuan Chiang, Giuseppe Fisicaro, Greg Juhasz, Sarom Leang, Bernadette Mohr, Utkarsh Pratiush, Francesco Ricci, Leopold Talirz, Pablo A. Unzueta, Trung Vo, Gabriel Vogel, Sebastian Pagel, Jan Janssen
}}

The complexity and non-intuitive user interface of scientific simulation software results in a high barrier for beginners and limits the usage to expert users. In the field of atomistic simulation, the simulation codes are developed by different communities (chemistry, physics, materials science) using different units, file names,and variable names. LangSim addresses this challenge, by providing a natural language interface for atomistic simulation in the field of computational materials science.

Since the introduction of ChatGPT, the application of large language models (LLM) in chemistry and materials science has transitioned from semantic literature search to research agents capable of autonomously executing selected steps of the research process. In particular, in research domains with a high level of automation, like chemical synthesis, the latest research agents already combine access to specialized databases, scientific software for analysis as well as to robots for executing the experimental measurements \cite{Boiko2023a, Bran2024a}. These research agents divide the research question into a series of individual tasks, each addressing one task before combining them with one controlling agent. With this approach, the flexibility of the LLM is reduced, which consequently reduces the risk of hallucinations \cite{Ye2023}.  

In analogy, LangSim (Language + Simulation) is a research agent focused on simulation in the field of computational materials science. LangSim can calculate a series of bulk properties for elemental crystals, like the equilibrium volume and equilibrium bulk modulus. Internally, this is achieved by constructing simulation protocols consisting of multiple simulation steps to calculate one material property. For example, the bulk modulus is calculated by querying the atomistic simulation environment (ASE) \cite{Larsen2017} for the equilibrium crystal structure, optimizing the crystal structure in dependence on the choice of simulation model, and finally evaluating the change of energy over volume change around the equilibrium to calculate the bulk modulus as the second derivative of the change in energy over volume change. The simulation protocols in LangSim are independent of the selected level of theory and can be evaluated with either the effective medium theory model \cite{Jacobsen1996} or the foundation machine-learned interatomic potential MACE \cite{Batatia2024}. Furthermore, to quantify the uncertainty of these simulation results, LangSim also has access to databases with experimental references for these bulk properties.

\begin{figure}[h!]
    \centering
    \includegraphics[width=0.99\textwidth]{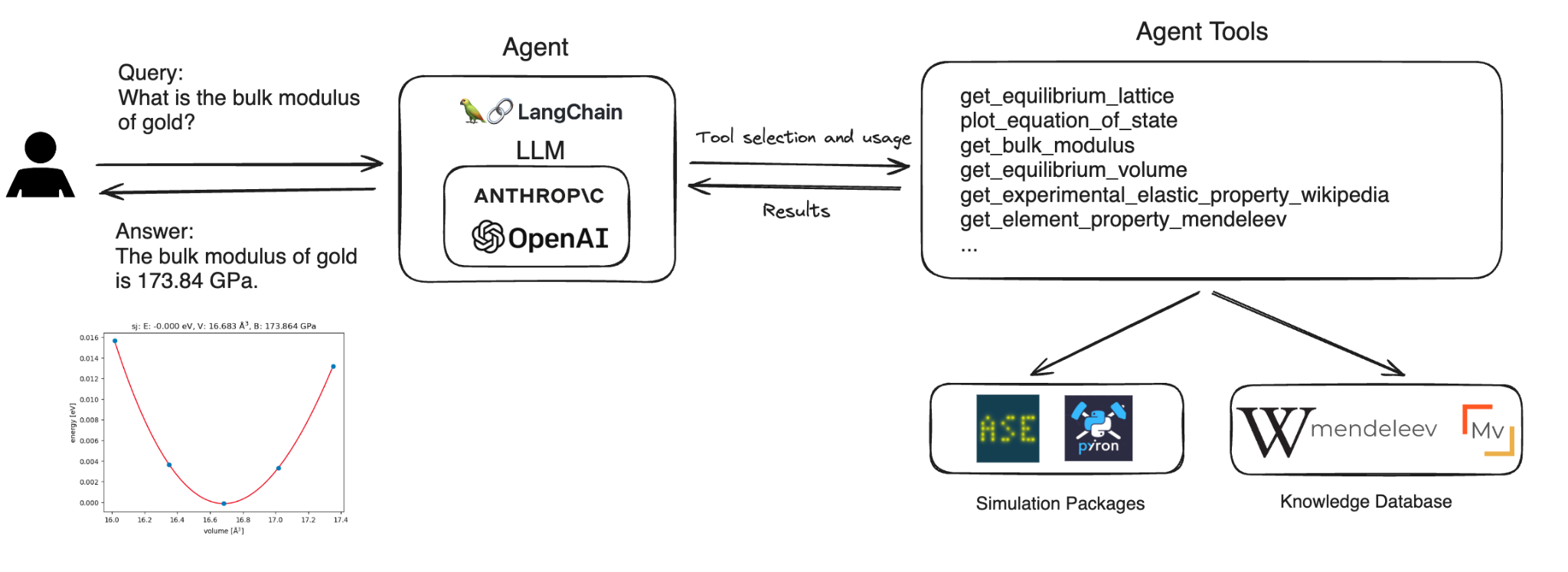} 
    \caption{
    \label{fig:langsim_workflow}}
\end{figure}

The LangSim research agent is based on the LangChain package. This has two advantages: On the one hand, the LangChain package \cite{LangChaina} simplifies the addition of new simulation agents and simulation workflows for LangSim. On the other hand, LangChain is compatible with a wide range of different LLM providers to prevent vendor lock-in. LangSim extends the LangChain framework by providing data types for coupling the simulation codes with the LLM, like a pydantic dataclass \cite{Pydantic} representation of the ASE atoms class and a series of pre-defined simulation workflows to highlight how existing simulation workflows can be implemented as LangChain agents. Once a simulation workflow is represented as a Python function compatible with a simulation framework like ASE, the interfacing with LangSim is as simple as changing the input arguments to LLM-compatible data types indicated by type hints and adding a Docstring as context for the LLM. Abstractly, these LangChain agents can be understood in analogy to the header files in C programming, which define the interfaces for public functions. An example LangSim agent to calculate the bulk modulus is provided below: 

\begin{lstlisting}
from ase.atoms import Atoms
from ase.calc import Calculator
from ase.eos import calculate_eos
from ase.units import kJ 
from langsim import (
    AtomsDataClass, 
    get_ase_calculator_from_str,
)
from langchain.agents import tool

def get_bulk_modulus_function(
    atoms: Atoms, calculator: Calculator
) -> float:
    atoms.calc = calculator
    eos = calculate_eos(atoms)
    v, e, B = eos.fit()
    return B / kJ * 1.0e24
 
@tool
def get_bulk_modulus_agent(
    atom_dict: AtomsDataClass, calculator_str: str
) -> float:
    """
    Returns the bulk modulus of chemical symbol 
    for a given atoms dictionary and a selected 
    model specified by the calculator string in GPa.
    """
    return get_bulk_modulus_function(
        atoms=Atoms(**atom_dict.dict()),
        calculator=get_ase_calculator_from_str(
            calculator_str=calculator_str
        ),
    )

\end{lstlisting}

The example workflow for calculating the bulk modulus highlights how existing simulation frameworks like ASE can be leveraged to provide the LLM with the ability to construct and execute simulation workflows. While for this example, in particular for the case of elemental crystals, it would be possible to pre-compute all combinations and restrict the large language model to a database of pre-computed results, this would become prohibitive for multi-component alloys and an increasing number of simulation models and workflows. At this stage the semantic capabilities of the LLM provide the capability to handle all possible combinations in a systematic way by allowing the LLM to construct and execute the specific simulation workflow when it is requested from the user. 

In summary, the LangSim package provides data classes and utility functions to interface LLMs with atomistic simulation workflows. This provides the LLM with the ability to execute simulations to answer scientific questions like a computational material scientist. The functionality is demonstrated for the calculation of the bulk modulus for elemental crystals.

\subsection{One sentence summaries}
\begin{enumerate}
    \item Problem/Task: Develop a natural language interface for simulation codes in the field of computational chemistry and materials science. Current LLMs, including ChatGPT 4, suffer from hallucination, resulting in simulation protocols that can be executed but fail to calculate the correct physical property with the specified unit. 
    \item Approach: Develop a suite of LangChain agents to interface with the atomic simulation environment (ASE) and corresponding data classes to represent objects used in atomistic simulation in the context of large language models. 
    \item Results and Impact: Developed the LangSim package as a prototype for handling the calculation of multiple material properties using predefined simulation workflows, independent of the theoretical model, based on the ASE framework.
    \item Challenges and Future Work: The current prototype enables the calculation of bulk properties for unaries, the next step is to extend this functionality to multi-component alloys and more material properties.  
\end{enumerate}

\newpage
\section{LLMicroscopilot: assisting microscope operations through LLMs}\label{sec:llmicroscopilot}

\textbf{\textit{Authors: Marcel Schloz, Jose C. Gonzalez
}}

The operation of state-of-the-art microscopes in materials science research is often limited to a selected group of operators due to their high complexity and significant cost of ownership. This exclusivity creates a barrier to broadening scientific progress and democratizing access to these powerful instruments. Presently, operating these microscopes involves time-consuming tasks that demand substantial human expertise, such as aligning the instrument for optimal performance and transitioning between different operational modes to address different research questions. These challenges highlight the need for improved user interfaces that simplify operation and increase the accessibility of microscopes in materials science.

Recent advancements in natural language processing software suggest that integrating large language models (LLMs) into the user experience of modern microscopes could significantly enhance their usability. Just as modern chatbots have enabled users without much programming background to create complex computer programs, LLMs have the potential to simplify the operation of microscopes, thereby making them more accessible to non-expert users \cite{Bauer2024}. Early studies have demonstrated the potential of LLMs in scanning probe microscopy, using microscope-specific external tools for remote access \cite{Diao2024} and control \cite{Liu2024}. Particularly promising is the application of LLMs as agents with access to specific external tools, providing operators with a powerful assistant capable of reasoning based on observations and reducing the extensive hallucinations common in LLM agents. This approach also enhances the accessibility of external tools, eliminating the need for users to learn tool-specific APIs.

\begin{figure}[h!]
    \centering
    \includegraphics[width=0.99\textwidth]{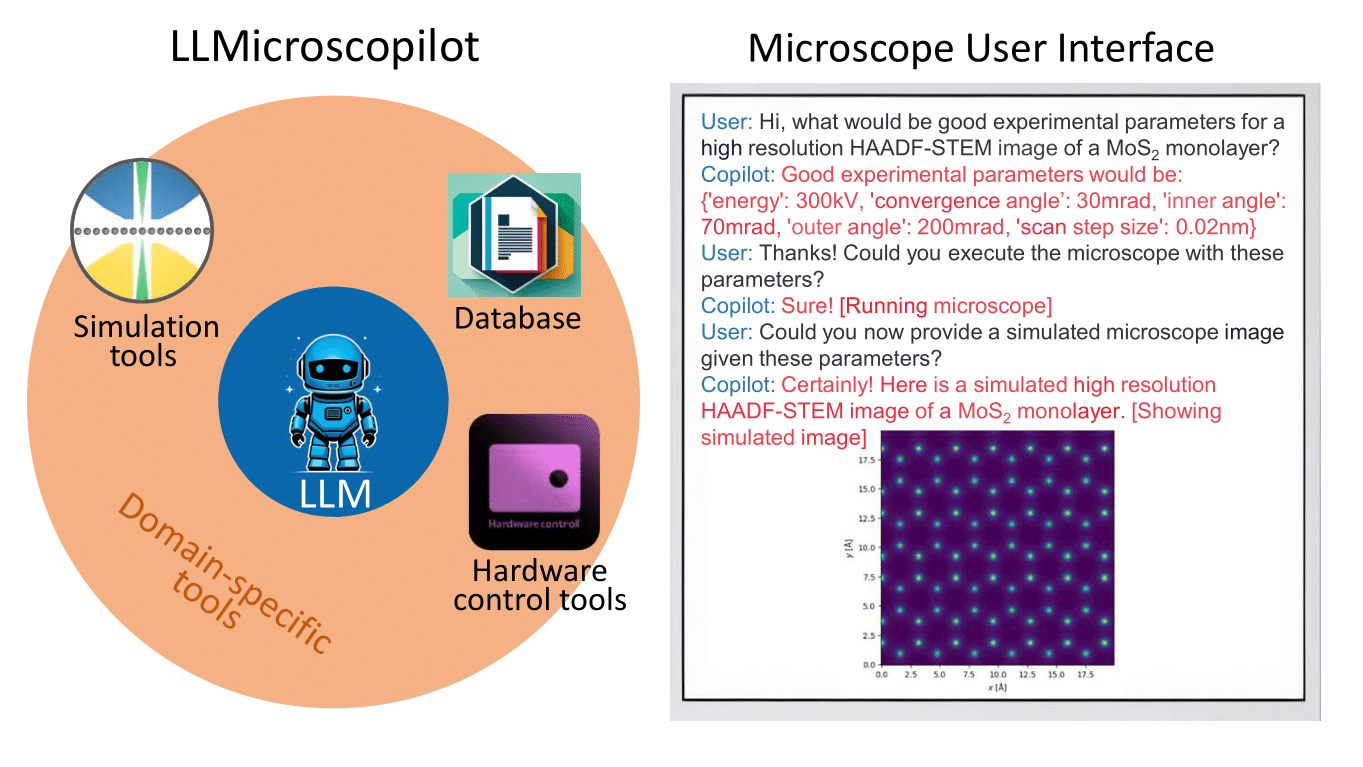} 
    \caption{Schematic overview of the LLMicroscopilot assistant. The microscope user interface allows the user to input queries, which are then processed by the LLM. The LLM executes appropriate tools to provide domain-specific knowledge, support data analysis, or operate the microscope.
    \label{fig:llmicroscopilot}}
\end{figure}

The LLMicroscopilot-team (Jose D. Cojal Gonzalez and Marcel Schloz) has shown that the operation of a scanning transmission electron microscope can be partially performed by the LLM-powered agent "LLMicroscopilot" through access to microscope-specific control tools. \autoref{fig:llmicroscopilot} illustrates the interaction process between the operator and the LLMicroscopilot. The LLMicroscopilot is built on a generally trained foundation model that gains domain-specific knowledge and performance through the provided tools. The initial prototype uses the API of a microscope experiment simulation tool \cite{Madsen2021} to perform tasks such as experimental parameter estimation and experiment execution. This approach reduces the reliance on highly trained human operators, fostering broader participation in materials science research. Future developments of LLMicroscopilot will integrate open-source microscope hardware control tools \cite{Meyer2019} and database-access tools, allowing for Retrieval-Augmented Generation possibilities to improve parameter estimation and data analysis.

\clearpage
\newpage
\section{T2Dllama: Harnessing Language Model for Density Functional Theory (DFT) Parameter Suggestion}\label{sec:t2dllama}

\textbf{\textit{Authors: Chiku Parida, Martin H. Petersen
}}

\subsection{Introduction}

Large language models are now gaining the attention of many researchers due to their capabilities to process human language and perform tasks on which they have not been explicitly trained, making them an invaluable tool for researchers in various fields where the information is in the form of text, like scientific journals, blogs, news articles, and social media posts, etc. This is particularly applicable to the field of chemical sciences, which encounters the challenge of dealing with limited and diverse datasets that are often presented in text format. LLMs have proven their potential in handling these challenges and are progressively being utilised to predict chemical characteristics, optimise reactions, and even independently design and execute experiments \cite{Mirza2024a}.

Here, we used LLM to process published scientific articles to extract simulation parameters and other relevant information about different materials. This will help experimentalists get an idea of optimised parameters for Density Functional Theory (DFT) calculations for the newly discovered material family.

Nowadays, DFT is the most valuable tool to model atomistic materials.  The idea behind DFT is to use Kohn-Sham's equations to approximately solve Schröding's equations for the atomic material at hand. The approximation is done by configuring the electron density for the material at each ionic step, where the ions move in position based on their energy and forces determined by the configured electron density. The biggest part of the approximation is the exchange functional, and depending on the complexity of the exchange functional, the approximation becomes more or less comparable with experimental results \cite{Giustino2014}. When performing a DFT calculation, the question is always what exchange functional and parameters to use, as well as what k-space grid to use. This is material-dependent and will change for different materials. The unoptimized parameters can lead to inaccurate results, resulting in a DFT calculation that fails to describe the relative values and is therefore not comparable to the experimental results \cite{Hafner2008}. For that reason, experimentalists normally collaborate with computational chemists because of their expertise in computational modeling or make due without the atomistic model. Instead, our T2Dllama [talk-to-douments using Llama] framework can be an acceptable solution to give the necessary DFT parameters, and using additional tools on the top of the LLM interface, it can create inputs for atomistic simulations \cite{Kresse1993, Mortensen2005} from the provided structure file by the user.

\subsection{Retrieval Augmented Generation }

\begin{figure}[h!]
    \centering
    \includegraphics[width=0.99\textwidth]{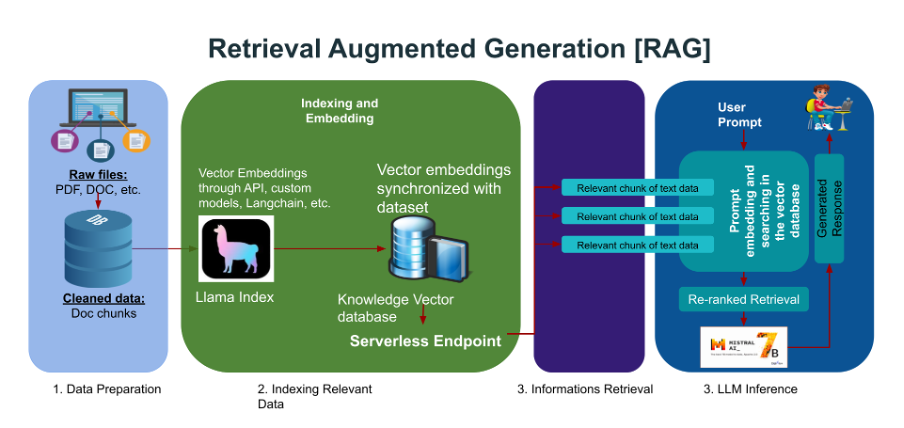} 
    \caption{Retrieval Augmented Generation [RAG] architecture with LLM interface
    \label{fig:t2dllama_fig}}
\end{figure}

The retrieval augmented generation (RAG) technique \cite{Gao2024a} is a popular and efficient technique for generating relevant contextual responses using pre-trained models. It is also less complex as compared to fine-tuning and training LLM from scratch. \autoref{fig:t2dllama_fig} explains our RAG workflow.
The RAG workflow has the following blocks:

\textit{Data Preparation:} We collect open-access scientific journals using Arxiv-api and the necessary filters. Pre-existing licensed scientific journals in the local database can be used for confidential purposes. Then the text documents are processed to create chunks of text.

\textit{Indexing and Embedding:} In this crucial step, we use llama indexing \cite{LlamaIndex} to store the tokenized documents with vector indexes and embeddings, transforming them into knowledge vector databases.

\textit{Information Retrieval:} When the LLM interface receives a human prompt, the model searches the knowledge vector database and retrieves the relevant chunks of stored data.

\textit{LLM Interface:} This is the last step where we communicate with the user. The LLM interface received the prompt from the user and got the information, as explained previously. The retrieved chunks during information retrieval are processed by the pretrained model [Mistral 7B] \cite{Mistral7B}, and the created context is delivered to the user as a response.

\subsection{Summary and outlooks}
The biggest issue here is training a GPT model to predict DFT exchange functional and parameters is filtering the data. There is no consensus about using a specific exchange functionals and parameters for a material, which makes the model confused. A way to avoid this is to only use papers, where DFT calculations are directly comparable with experimental results. This will limit the variability in the data and ensure that the model is trained on reliable information. 
To conclude through a pre-trained GPT model, we are able to predict DFT exchange functionals and Hubbard-U parameters as well as k-point grid by using RAG technique. Further, as we are retrieving information about simulation parameters and exchange-functionals for materials consistent with respect to experiments from the documents , it is important to prioritise relevant and reliable sources and special techniques to ensure the accuracy of the data being extracted. One approach that can be taken involves employing advanced tokenizer techniques specifically designed for scientific notations and terms, so that we can get reliable embedding and vector indexing. This will help to improve the quality of the responses. We still need to do more thorough training as well as a GUI application, but as an initial try in this LLM hackathon we showed that the idea can be realised.

\subsection{Data and Code Availability}
All code and data used in this study are publicly available in the following GitHub repository: \url{https://github.com/chiku-parida/T2Dllama}

\newpage
\section{Materials Agent: An LLM-Based Agent with Tool-Calling Capabilities for Cheminformatics}\label{sec:materials-agent}

\textbf{\textit{Authors: Archit Datar, Kedar Dabhadkar
}}

\subsection{Introduction}

Out-of-the box large language model (LLM) implementations such as ChatGPT, while offering interesting responses, generally provide little to no control over the workflow of the LLM by which the response is generated. In other words, it is easy to get LLMs to say something in response to a prompt, but difficult to get them to do something via an expected workflow. A solution to do this problem is to equip LLMs with tool-calling capabilities; i.e., allow the LLM to generate the response via an available function(s) which is (are) appropriate to answer the prompt. An LLM with tool-calling capabilities, when prompted, typically decides which tool(s) to call (execute) and the order in which to execute them, along with the arguments to pass to them. It then executes these and returns the response. Such a system offers several powerful capabilities such as the ability to query databases to return the latest information and reliably perform mathematical calculations. Such capabilities have been incorporated into ChatGPT via plugins such as Expedia and Wolfram, among others \cite{OpenAIChatGPTPlugins}. In the chemistry literature, recent attempts have been made by researchers such as Smit and coworkers, Schwaller and coworkers, among others \cite{Jablonka2024b, Bran2024}. 

Through Materials Agent, which is an LLM-based agent with tool-calling capabilities for cheminformatics, we seek to build on these attempts to provide a variety of important tool-calling capabilities and build a framework to expand on these. We hope that this can serve to increase LLM adoption in the community and lower the barrier to entry for cheminformatics. Materials Agent is built using the LangChain library \cite{LangChainb}, GPT 3.5-turbo \cite{OpenAIModels} as the underlying LLM, and the user interface is based on the FastDash project \cite{FastDash}. In this demonstration, we have provided tools based on RDKit \cite{RDKita}—a popular cheminformatics library, some custom tools, as well as a Retrieval Augmented Generation (RAG) to allow LLM to interact with documents. The full code is available at \url{https://github.com/dkedar7/materials-agent} and the working application, hosted on Google Cloud Platform (GCP), is available at \url{https://materials-agent-hpn4y2dvda-ue.a.run.app/}. The demonstration video is uploaded on YouTube at \url{https://www.youtube.com/watch?v=_5yCOg5Bi_Q&ab_channel=ArchitDatar}. In the following section, we describe some key use cases.

\subsection{Equipping LLMs with tools based on standard cheminformatics packages (RDKit)}
Common cheminformatics workflows involve obtaining SMILES strings and key properties of molecules. Searching for these data individually can be time consuming. The RDKit CalcMolDescriptors module comes pre-loaded with these data based on SMILES strings. We created a function to select 19 of the most common properties (for ease of visualization) and added it as a tool to the LLM. Not only can the LLM perform the simple task of providing these descriptions when a SMILES string is input, but it can also perform more complicated tasks involving this functionality. As shown in \autoref{fig:materialsagent1} below, given a complex query, the LLM breaks it down into individual tasks, utilizes these tools in the correct sequence, and renders a response. 

\begin{figure}[h!]
    \centering
    \includegraphics[width=0.99\textwidth]{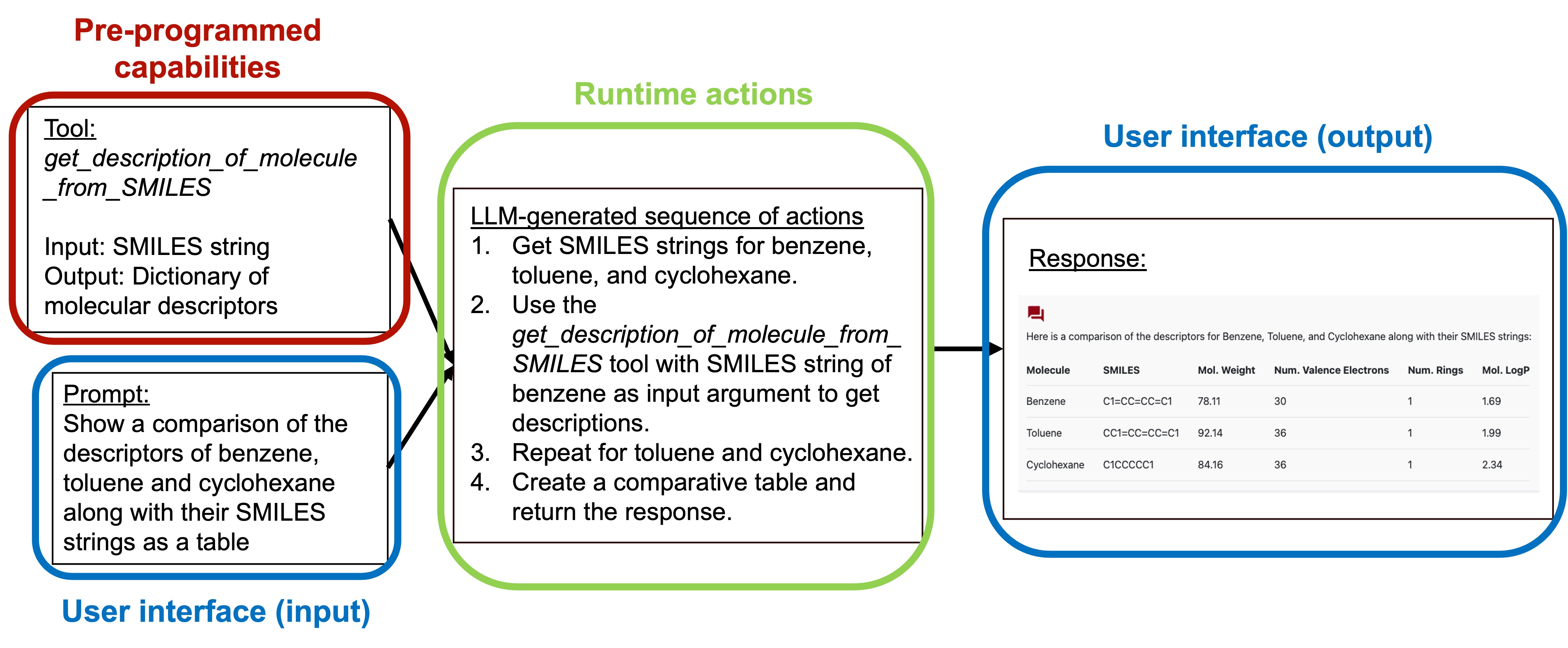} 
    \caption{Workflow of a response generated to a user prompt by Materials Agent using a tool based on RDKit.
    \label{fig:materialsagent1}}
\end{figure}

\subsection{Tool-calling for custom tools: Radial Distribution Function calculation}
\begin{figure}[h!]
    \centering
    \includegraphics[width=0.99\textwidth]{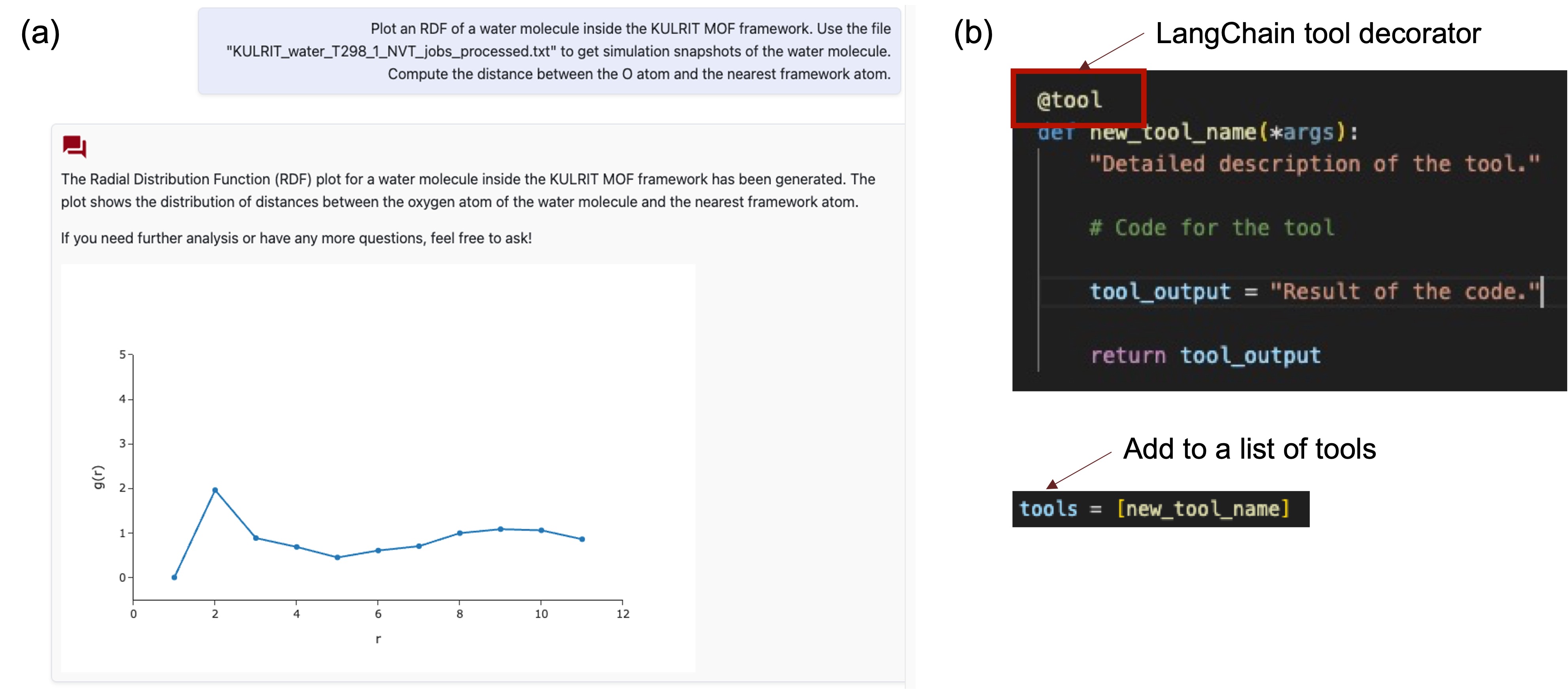} 
    \caption{Niche purpose tools with LLM. (a) Illustration of the RDF computing tool. (b) Code snippet to highlight the ease of transferability of LLMs with tool-calling capabilities.
    \label{fig:materialsagent2}}
\end{figure}

Combining custom tools with LLMs lead to some interesting advantages. For one, they offer the makers of these custom tools an ability to provide their users with a more intuitive natural language-based user interface which can lower the barrier to entry and increase adoption. Other benefits can be that they can be integrated into other workflows and automate more work. To demonstrate this, we have built a tool which computes and plots the radial distribution function (RDF) and integrated it with an LLM. The RDF is an important function commonly used in molecular dynamics and Monte Carlo simulations to understand the statistics of distances between two particles averaged over the simulation. By studying these, one can understand the nature of interactions between these particles. Here, we constructed a custom tool to compute RDF for the distance between a water molecule and the framework atoms in a metal-organic framework (MOF) in a single molecule canonical ensemble Monte Carlo simulation. The inputs are a PDB file of the MOF structure and a TXT file containing snapshots of the location of the water molecule during the simulation. The distance computation also accounts for triclinic periodic boundary conditions which is the accurate way to quantify distances for crystalline systems such as this one. The inputs and outputs for this tool are shown below in \autoref{fig:materialsagent2}(a).

Furthermore, we also stress that this approach is easily scalable and transferable, and adding new tools is exceedingly easy. The reader is encouraged to clone our GitHub repository and experiment with adding new tools to this software package. New tools can be easily added to the \texttt{src/tools.py} file in the repository via the format shown in the code snippet in \autoref{fig:materialsagent2}(b).

\subsection{RAG capabilities}
\begin{figure}[h!]
    \centering
    \includegraphics[width=0.99\textwidth]{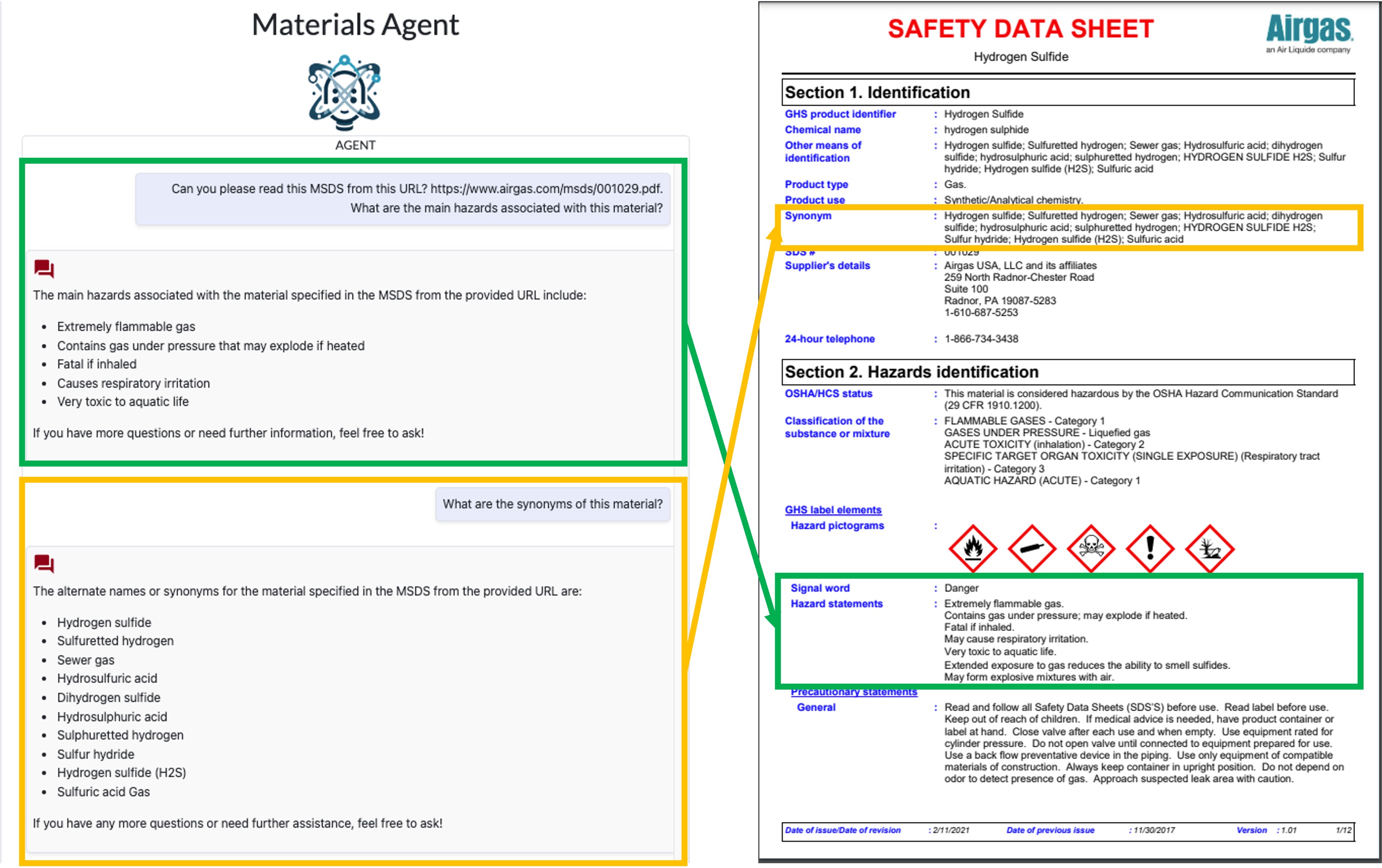} 
    \caption{Illustration of RAG for interacting with a MSDS.
    \label{fig:materialsagent3}}
\end{figure}

Summarizing and asking questions of documents is another common LLM use case. This capability is provided out-of-the box through the EmbedChain library \cite{EmbedChain}. and we have integrated that into Materials Agent for convenience. We demonstrate the utility of this by supplying the LLM with a URL to a materials and safety datasheet (MSDS) and asking questions of it (see \autoref{fig:materialsagent3}).

In future, we aim to expand the toolkit that the LLM is equipped with. For instance, we can add functions built on the publicly available PubChem database \cite{PubChem}. as well as some functions built off of it \cite{PubChemPy}. We also aim to train it on user manuals of commonly used molecular simulations software such as GROMACS \cite{Abraham2024}, RASPA \cite{Dubbeldam2015}, and QuantumEspresso \cite{Giannozzi2009} to assist with setting up molecular simulations. 

Through the experience of building Materials Agent, we realized that, while convenient, such agents cannot replace the need for human vigilance. At the same time, having the development of such an agent will make cheminformatics utilities easier to access for a broader range of users, lower the barrier to entry, and ultimately, accelerate the pace of materials development.

\newpage
\section{LLM with Molecular Augmented Token}

\textbf{\textit{Authors: Luis Pinto, Xuan Vu Nguyen, Tirtha Vinchurkar, Pradip Si, Suneel Kuman
}}

\subsection{Objective}

Our primary objective is to explore how chemical encoders such as molecular fingerprints or embeddings from 2D/3D deep learning models (e.g., ChemBERTa \cite{chithrananda2020chemberta}, UniMol \cite{zhou2022unimol}) can enhance large language models (LLMs) for zero-shot tasks such as property prediction, molecule editing, and generation. We aim to benchmark our approach against state-of-the-art models like LlaSmol \cite{yu2024llasmol} and ChatDrug \cite{liu2023chatgptpowered}, demonstrating the transformative potential of LLMs in the field of chemistry.

\begin{figure}[h!] 
    \centering
    \includegraphics[width=0.7\textwidth]{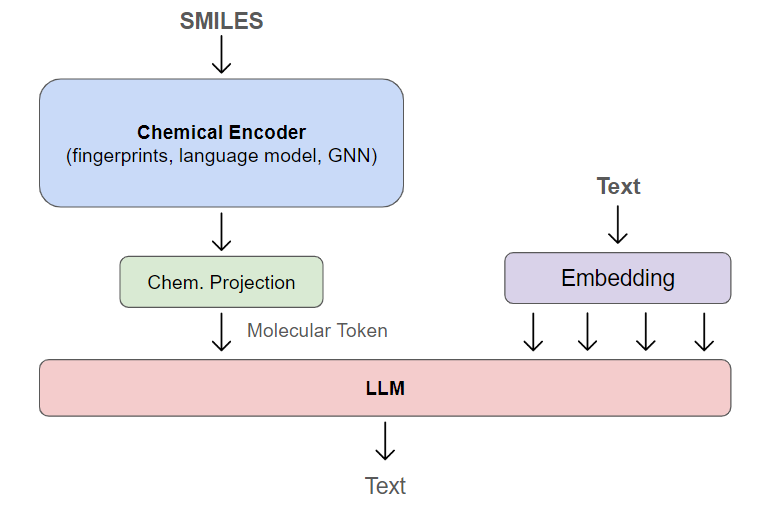} 
    \caption{Workflow for integrating chemical encoders with large language models. Molecular data from SMILES is transformed into molecular tokens and combined with text embeddings for tasks such as property prediction, molecule editing, and generation.}
\end{figure}

\subsection{Methodology}

We identified two key benchmarks to evaluate our approach:
\begin{itemize}
    \item \textbf{LlaSmol:} This benchmark involves fine-tuning a Mistral 7B model \cite{jiang2023mistral7b} on 14 different chemistry tasks, including 6 property prediction tasks. The LlaSmol project demonstrated significant performance improvements over baseline models, both open-source and proprietary, by using the SMolInstruct dataset, which contains over three million samples.
    \item \textbf{ChatDrug:} This framework leverages ChatGPT for molecule generation and editing tasks. It includes a prompt module, a retrieval and domain feedback module, and a conversation module to facilitate effective drug editing. ChatDrug showed superior performance across 39 drug editing tasks, encompassing small molecules, peptides, and proteins, and provided insightful explanations to enhance interpretability.
\end{itemize}

\subsection{Our approach}

We propose to fine-tune a Mistral 7B instruct model to compete against these benchmarks. Due to compute constraints, we were unable to complete the training. Training the model using QLoRA 4-bit \cite{dettmers2023qlora} on 50k property prediction samples requires approximately 5 hours per epoch on a 24GB GPU.

\begin{figure}[hbt!]
    \centering
    \includegraphics[width=0.7\textwidth]{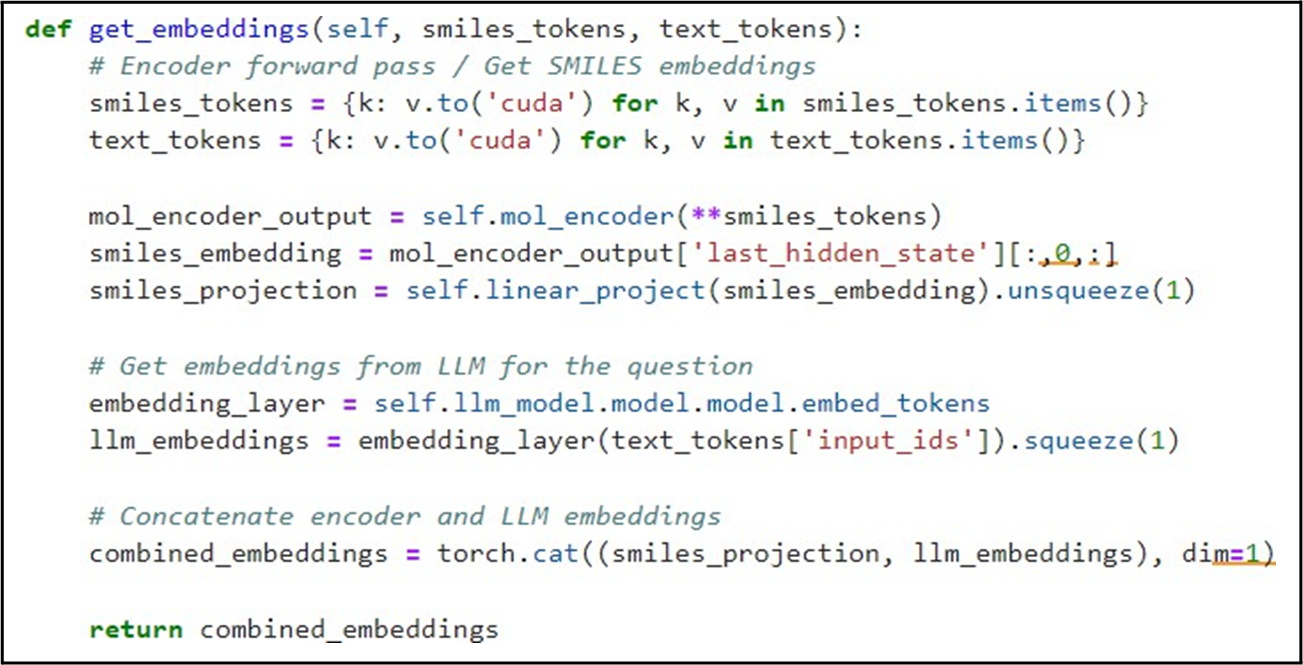} 
    \caption{Modified code of the function get\_embeddings.}
    \label{fig:screenshot-code1}
\end{figure}

\begin{figure}[hbt!]
    \centering
    \includegraphics[width=0.99\textwidth]{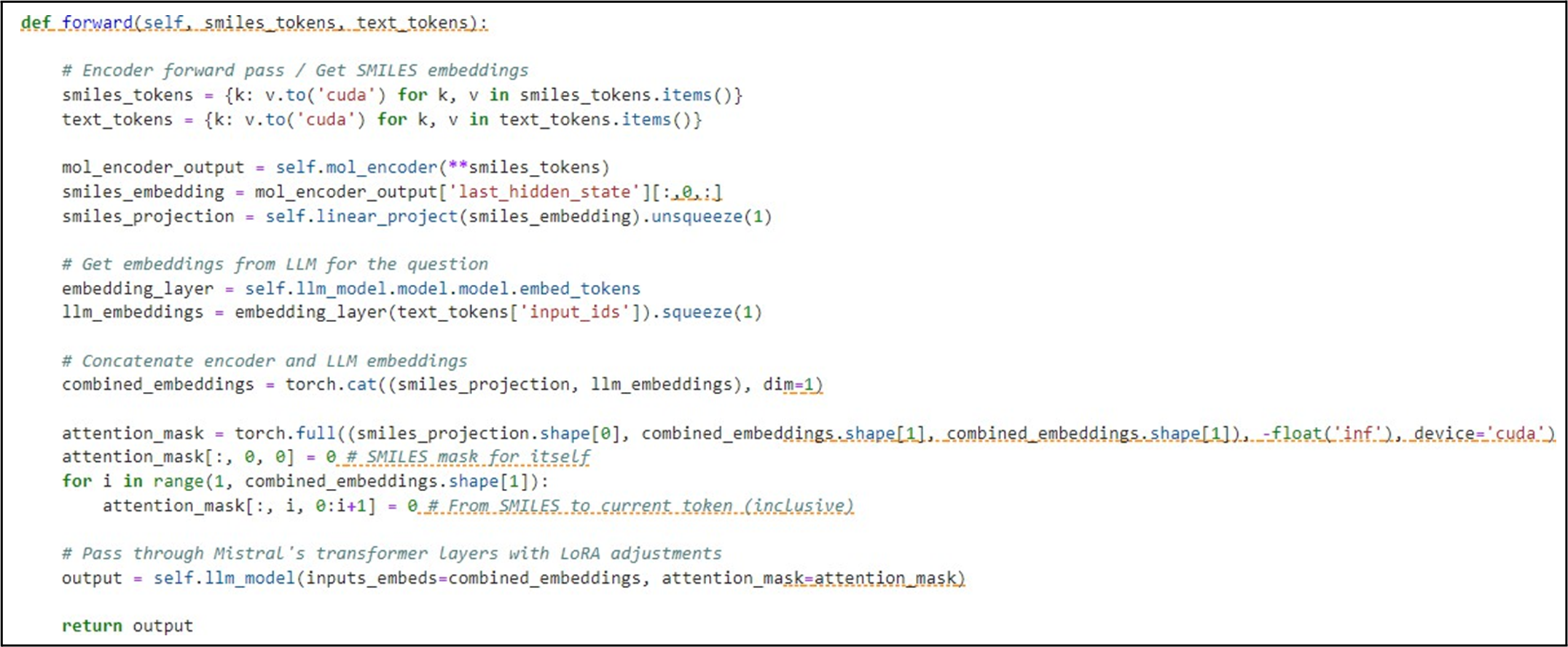} 
    \caption{Modified forward function which allows for the molecular token to be added.}
    \label{fig:screenshot-code2}
\end{figure}

Steps Taken:
\begin{itemize}
    \item \textbf{Data Preparation:} We utilized chemical encoders to transform molecular structures into suitable embeddings for the LLM.
    \item \textbf{Model Modification:} We integrated the embeddings into the LLM's forward function to enrich its input.
    \item \textbf{Fine-Tuning:} We applied QLoRA for efficient training on limited computational resources.
    \item \textbf{Preliminary Results and Ongoing Work:} Although we are still fine-tuning the LLMs, initial results are promising.
    \item \textbf{Code Snippets:} Screenshots in Figures \ref{fig:screenshot-code1} and \ref{fig:screenshot-code2} demonstrate the modifications made to the model code to extract embeddings and implement the forward function of the LLM.
\end{itemize}

\subsection{Conclusion}

Our project underscores the potential of using LLMs enhanced with chemical encoders in materials science and chemistry. By fine-tuning these models, we aim to improve property prediction and facilitate molecule editing and generation, paving the way for future research and applications in this space. The code is available at \url{https://github.com/luispintoc/LLM-mol-encoder}.

\newpage
\section{MaSTeA: Materials Science Teaching Assistant}\label{sec:mastea}

\textbf{\textit{Authors: Defne Circi, Abhijeet S. Gangan, Mohd Zaki
}}
\subsection{Dataset Details}

We take 650 questions from materials science question answering dataset (MaScQA)\cite{Zaki2024}, which required undergraduate-level understanding to solve them. The authors classified them into four types based on their structure: Multiple choice questions (MCQs), Match the following type questions (MATCH), Numerical questions where options are given (MCQN), and numerical questions (NUM): see \autoref{mascqa_dataset}. MCQs are generally conceptual, given four options, out of which mostly one is correct, and sometimes more than one option is also correct. In MATCH, two lists of entities are given, which are to be matched with each other. These questions are also provided with four options, out of which one has the correct set of matched entities. In MCQN, the question has four choices, out of which the correct one is identified after solving the numerical problem stated in the question. The NUM type questions have numerical answers, rounded to the nearest integer or floating-point number as specified in the questions.

To understand the performance of LLMs from a domain perspective, the questions were classified into 14 topical categories \cite{Zaki2024}. The database can be accessed at \url{https://github.com/M3RG-IITD/MaScQA}.

\subsection{Methodology}

Our objective was to automate the evaluation of both open-source and proprietary LLMs on materials science questions from the MaScQA dataset and provide an interactive interface for students to solve these questions. We evaluated the performance of several language models, including LLAMA3-8B, HAIKU, SONNET, GPT-4, and OPUS, across the 14 various categories such as characterization, applications, properties, and behavior.

The evaluation involved:
\begin{itemize}
    \item \textbf{Extracting corresponding values:} For multiple-choice questions, correct answer options were extracted using regular expressions to compare model predictions against the correct choices.
    \item \textbf{Prediction verification:} For numerical questions, the predicted value was checked against a specified range or exact value. For multiple-choice questions, the predicted answer was verified against the correct option or the extracted corresponding value.
    \item \textbf{Calculating accuracy:} Accuracy was calculated for each question type and topic, and the overall accuracy across all questions was computed.
\end{itemize}

The results of the evaluation are summarized in \autoref{evaluation on mascqa}, which presents the accuracy of the various models for different question types and topics. The \textit{\textbf{opus}} variant of Claude consistently outperformed the others, achieving the highest accuracy in most categories. GPT-4 also showed strong performance, particularly in topics related to material processing and fluid mechanics.

Our interactive web app, MaSTeA (Materials Science Teaching Assistant), developed using Streamlit, allows easy model testing to identify LLMs` strengths and weaknesses in different materials science subfields. The results suggest that there is significant room for improvement to enhance the accuracy of language models in answering scientific questions. Once these models become more reliable, MaSTeA could be a valuable tool for students to practice answering questions and learn the steps to get to the answer. By analyzing LLM performance, we aimed to guide future model development and pinpoint areas for improvement.

Our code and application can be found at:
\begin{itemize}
    \item \url{https://github.com/abhijeetgangan/MaSTeA}
    \item \url{https://mastea-nhwpzz8fehvc9b3n5bhzya.streamlit.app/}
\end{itemize}

\begin{table}[h!]
    \centering
    \caption{Sample questions from each category: (a) multiple choice question (MCQ), (b) matching type question (MATCH), (c) numerical question with multiple choices (MCQN), and (d) numerical question (NUM). Correct answers are in bold.}
    \begin{tabular}{|m{7.5cm}|m{7.5cm}|}
        \hline
        Floatation beneficiation is based on the principle of & A peak in the X-ray diffraction pattern is observed at $2\theta = 78^\circ$, corresponding to $\{311\}$ planes of an fcc metal, when the incident beam has a wavelength of 0.154 nm. The lattice parameter of the metal is approximately \\
        \textbf{(A) Mineral surface hydrophobicity} & (A) 0.6 nm \\
        (B) Gravity difference & \textbf{(B) 0.4 nm} \\
        (C) Chemical reactivity & (C) 0.3 nm \\
        (D) Particle size difference & (D) 0.2 nm \\
        \hline
        (a) Multiple choice question (MCQ) & (c) Numerical question with multiple choices (MCQN) \\ \hline
        Match the composite in Column I with the most suitable application in Column II. & The third peak in the X-ray diffraction pattern of a face-centered cubic crystal is at $2\theta$ value of $45^\circ$, where $2\theta$ is the angle between the incident and reflected rays. The wavelength of the monochromatic X-ray beam is 1.54 Å. Considering first-order reflection, the lattice parameter (in Å) of the crystal is? (Round off to two decimal places) \\
        Column I: (P) Glass fibre reinforced plastic, (Q) SiC particle reinforced Al alloy, (R) Carbon-carbon composite, (S) Metal fibre reinforced rubber: & \textbf{Ans. 5.64 to 5.73} \\
        Column II: (1) Missile cone heads, (2) Commercial automobile chassis, (3) Airplane wheel tyres, (4) Car piston rings, (5) High performance skate boards: & \\
        (A) P-4, Q-5, R-1, S-2 (B) P-3, Q-5, R-2, S-4 & \\
        \textbf{(C) P-5, Q-4, R-1, S-3} (D) P-4, Q-2, R-3, S-1 & \\
        \hline
        (b) Matching type question (\textbf{MATCH}) & (d) Numerical question (\textbf{NUM}) \\
        \hline
    \end{tabular}
\label{mascqa_dataset}
\end{table}

\begin{table}[h!]
\centering
\caption{Accuracy of Language Models by Topic}
\scalebox{1.0}{
\begin{tabular}{lcccccc}
\hline
\textbf{Topic}                 & \textbf{\# Questions} & \textbf{LLaMA-3-8b} & \textbf{Haiku} & \textbf{Sonnet} & \textbf{OPUS} & \textbf{GPT4} \\ \hline
Thermodynamics        & 114   & 37.72  & 47.37  & 55.26  & \textbf{73.68}  & 57.02  \\ 
Atomic structure       & 100   & 32     & 40     & 49     & \textbf{64}     & 59     \\ 
Mechanical behavior    & 96    & 22.92  & 41.67  & 52.08  & \textbf{71.88}  & 43.75  \\ 
Material manufacturing & 91    & 43.96  & 57.14  & 56.04  & \textbf{80.22}  & 68.13  \\ 
Material applications  & 53    & 52.83  & 64.15  & 77.36  & \textbf{92.45}  & 86.79  \\ 
Phase transition       & 41    & 31.71  & 46.34  & \textbf{65.85}  & 70.73  & 63.41  \\ 
Electrical properties  & 36    & 33.33  & 25     & 55.56  & \textbf{72.22}  & 44.44  \\ 
Material processing    & 35    & 48.57  & 54.29  & 74.29  & \textbf{88.57}  & \textbf{88.57}  \\
Transport phenomena    & 24    & 37.5   & \textbf{70.83}  & 58.33  & 87.5   & 62.5   \\
Magnetic properties    & 15    & 26.67  & 46.67  & 46.67  & \textbf{66.67}  & 60     \\ 
Material characterization & 14    & 78.57  & 57.14  & 85.71  & \textbf{92.86}  & 71.43  \\ 
Fluid mechanics        & 14    & 21.43  & 50     & 57.14  & 78.57  & \textbf{85.71}  \\ 
Material testing       & 9     & 77.78  & 66.67  & \textbf{100}    & \textbf{100}    & \textbf{100}    \\
Miscellaneous          & 8     & 62.5   & 62.5   & 62.5   & \textbf{75}     & 62.5   \\ \hline
\end{tabular}
\label{evaluation on mascqa}
}
\end{table}

\clearpage
\newpage
\section{LLMy-Way}\label{sec:llmy-way}

\textbf{\textit{Authors: Ruijie Zhu, Faradawn Yang, Andrew Qin, Suraj Sudhakar, Jaehee Park, Victor Chen}}

\subsection{Introduction}

In the academic realm, researchers frequently present their work and that of others to colleagues and lab members. This task, while essential, is fraught with difficulties.
For example, below are three challenges:
\begin{enumerate}
    \item Reading and understanding research papers: Comprehending the intricacies of a research paper can be daunting, particularly for interdisciplinary subjects like materials science.
    \item Creating presentation slides: Designing slides that effectively communicate the content requires significant effort, including remaking slides, sourcing images, and determining optimal text and image placement.
    \item Tailoring to the audience: Deciding on the appropriate level of technical vocabulary and the number of slides needed to fit within a given time limit adds another layer of complexity.
\end{enumerate}

\begin{figure}[h!]
    \centering
    \includegraphics[width=0.99\textwidth]{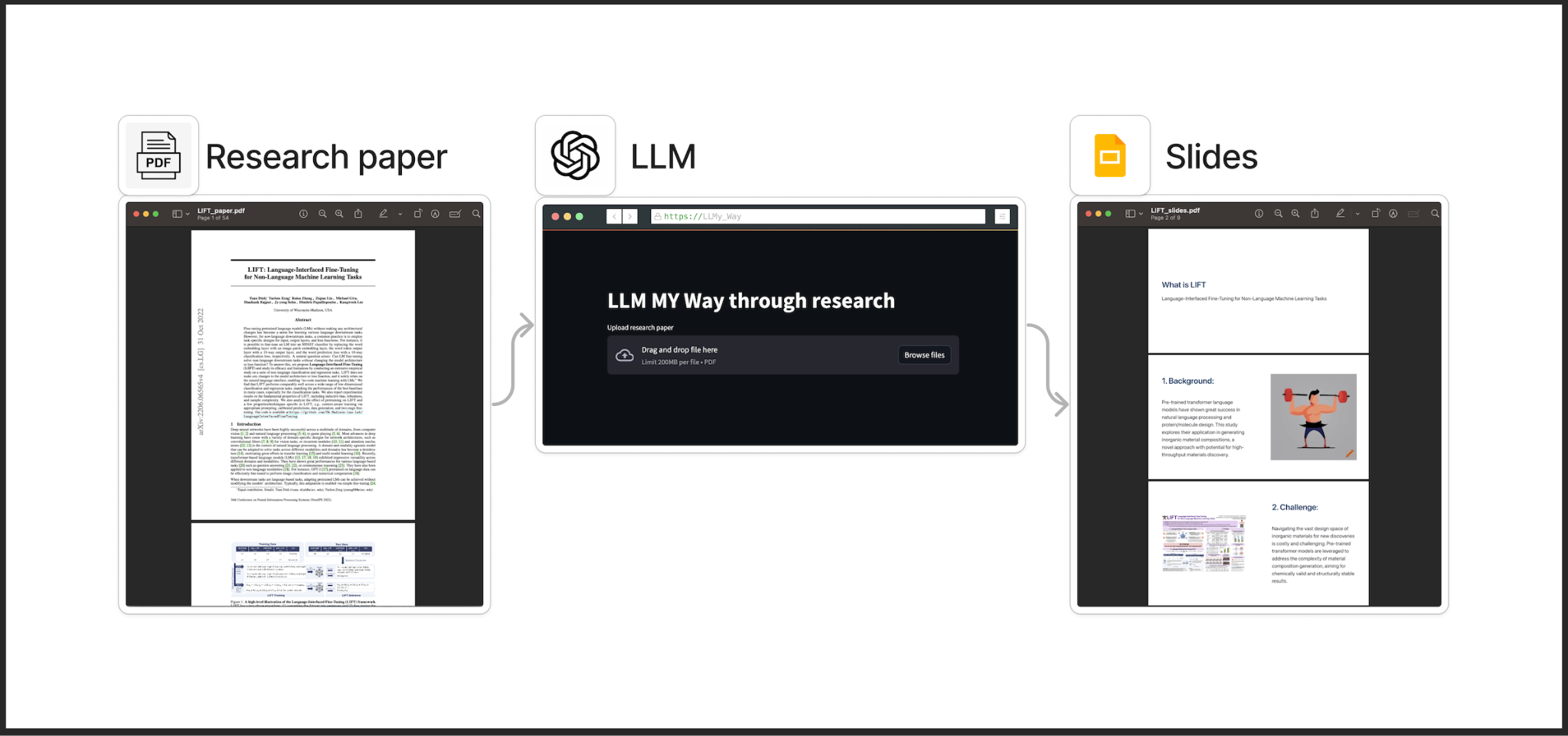} 
    \caption{
    \label{fig:llmyway}}
\end{figure}

These challenges can be effectively addressed using large language models, which can streamline and automate the text summarization and slide creation process. LLMy Way leverages the power of GPT-3.5-turbo to automate the creation of academic slides from research articles.
The methodology involves primarily three steps:

\subsection{Structured Summarization}
We prompt the large language model to generate a structured summary of the research paper, adhering to the typical sections of an academic paper: background, challenge, current status, method, result, conclusion, and outlook. Each section is summarized in under a certain number of words to ensure brevity and clarity.

Example Prompt:
Read the paper, and summarize in \{num\_words\} words each about the following:\textbackslash n\textbackslash n1. Background\textbackslash n2. Challenge\textbackslash n3. Current status\textbackslash n4. Method\textbackslash n5. Result\textbackslash n6. Conclusion\textbackslash n7. Outlook\textbackslash n

\subsection{Slide Generation}
To create slides, we format the language model's output in a specific manner, using symbols to denote slide breaks. This output is then parsed and converted into a Markdown file, where additional images and text formatting are applied as needed. The formatted Markdown file is subsequently transformed into PDF slides using Marp.
Example output formatting:
\begin{lstlisting}
# Background
Summary of the background here.
---
# Challenge
Summary of the challenge here.
---
\end{lstlisting}

\subsection{Customization for Audience and Time Limit}
LLMy Way allows customization based on the target audience's expertise level (expert or non-expert) and the presentation time limit (e.g., 10 minutes, 15 minutes). This information is incorporated into the initial generation phase, ensuring that the content and slide count are appropriately tailored. This feature is to be implemented.

\subsection{Conclusion}
LLMy Way represents a significant step forward towards the automation of academic presentation preparation. By leveraging the structured nature of scientific papers and the capabilities of advanced large language models, our tool addresses common pain points faced by researchers. The current implementation of our framework can be summarized into three consecutive steps. First, the research paper is parsed by LLM, which is summarized into predefined sections. Next, the summarized texts are converted into Markdown. Finally, Marp is used to generate the final slide deck in the PDF format. The current implementation of our framework uses GPT-3.5-turbo, but it can be adapted to other language models as needed. We also support the output format of LaTex to fit the needs of many researchers. Future work will focus on further refining the tool, incorporating user feedback, and exploring additional customization options.

\newpage
\section{WaterLLM: Creating a Custom ChatGPT for Water Purification Using Prompt-Engineering Techniques}\label{sec:waterllm}

\textbf{\textit{Authors: Viktoriia Baibakova, Maryam G. Fard, Teslim Olayiwola, Olga Taran
}}
\subsection{Introduction}
Drinking water pollution is a growing environmental problem that, ironically, comes from an increase in industrial production of new materials and chemicals. Common pollutants include heavy metals, perfluorinated compounds, microplastic particles, excreted medicinal drugs from hospitals, agricultural runoff and many others \cite{Mishra2023}. The communities that suffer the most from water contamination often lack the plumbing infrastructure necessary for centralized water analysis and  treatment. Decentralized and localized water treatments, based on resources available to the communities, can alleviate the problem. Since the resources can vary greatly, from well equipped analytical facilities to low-cost DIY solutions, a knowledge base that can rapidly provide information relevant for specific situations is needed.  Here we show a prototype Large Language Model (LLM) chatbot that can take a variety of inputs about possible contaminants (ranging from detailed LC/MS analysis to general description of the situation) and propose the best solution to the water treatment for the particular case based on  contaminant composition, cost and resources availability. We employed a Retrieval Augmented Generation (RAG) capability of ChatGPT to answer questions about possible water treatments based on the latest scientific literature .

For this project, we focused on advanced oxidation procedures (AOPs) for water purification from microplastics (MPs). In recent times, MPs have received significant attention globally due to their widespread presence in various species’ bodies, environmental media, and even bottled drinking water, as frequently documented \cite{Li2022}. Numerous trials have been conducted and reported on the use of AOPs for breaking down diverse persistent microplastics as wastewater treatment methodologies. However, there remains a lack of guidelines on selecting the most suitable and cost effective treatment method based on the characteristics of the contaminant, maximum removal percentage of MP. 

The complexity of existing research on AOPs for MPs can be tackled with LLMs enhanced with RAG. RAG allows to augment LLM’s knowledge and achieve state of the art results on knowledge-intensive tasks. One straightforward modern way to implement LLM with RAG is through configuring a custom chatGPT. We uploaded current scientific papers under the “Knowledge” for RAG and tailored chatbot performance with prompt-engineering techniques such as grounding, context, and chain-of-thought reasoning to ensure that it delivers accurate, detailed, and useful information.

\subsection{Grounding}

To make sure that chatbot provides accurate and scientifically valid answers, we loaded the latest research on microplastic pollution remediation for RAG and implemented grounding in chat prompt. To collect the data, we gathered the initial set of 10 review articles from the expert in the field that talk about water purification using Advanced Oxidation Process. Then, we manually found 112 scientific articles discussing the specific treatment procedure. This way, our dataset has studies on different pollution sources like laundry, hospitals, industry, different pollutant types and their descriptive characteristics like size, shape, color, and different treatment methods like Ozonation, Fenton Reaction, UV, Heat-Activated Persulfate. We merged all papers into 8 pdfs to meet chatGPT “Knowledge” restrictions in files number, size and length. With grounding, we aim to anchor the chatbot's responses in concrete and factual knowledge. We explicitly asked chatGPT to avoid giving wordy broad generalized answers and to provide concise scientific details using the files uploaded under Knowledge.

\subsection{Context}

We provided context for the chatbot to understand and respond appropriately to user queries. We specified that Chatbot has expert-level knowledge on MPs and water purification strategies from MPs and other contaminants. We defined a user as a technician with basic knowledge on chemical engineering that needs to choose and apply a purification method. We set that the communication between Chatbot and User should be in the form of interactive dialog. It means that Chatbot should ask follow-up questions from the user and should finally return an accurate purification protocol with all the details that can be reproduced in the experiment.

\subsection{Chain-of-Thought Reasoning}

To further ensure that the chatbot considers all necessary aspects before providing a solution, we employed the chain-of-thought reasoning by breaking down the problem-solving process into sequential steps: 
Get the source of contamination.
Ask what the pollutant particle is and suggest evaluation tests if it is unknown.
Ask if the characteristics of pollutants are known: size, shape, type.
Suggest the most effective purification approach that will get rid of the largest percentage of the pollutants and estimate the price. Adjust if it is too expensive.
Inquiry about the post-treatment analysis. If unsatisfactory, go to the previous step.
End of conversation: Chatbot should provide a table with all used purification methods and their parameters for established contaminants.

These techniques allowed to boost the custom GPT performance, and WaterLLM demonstrated performance approved by the expert in the field.

\begin{figure}[h!]
    \centering
    \includegraphics[width=0.8\textwidth]{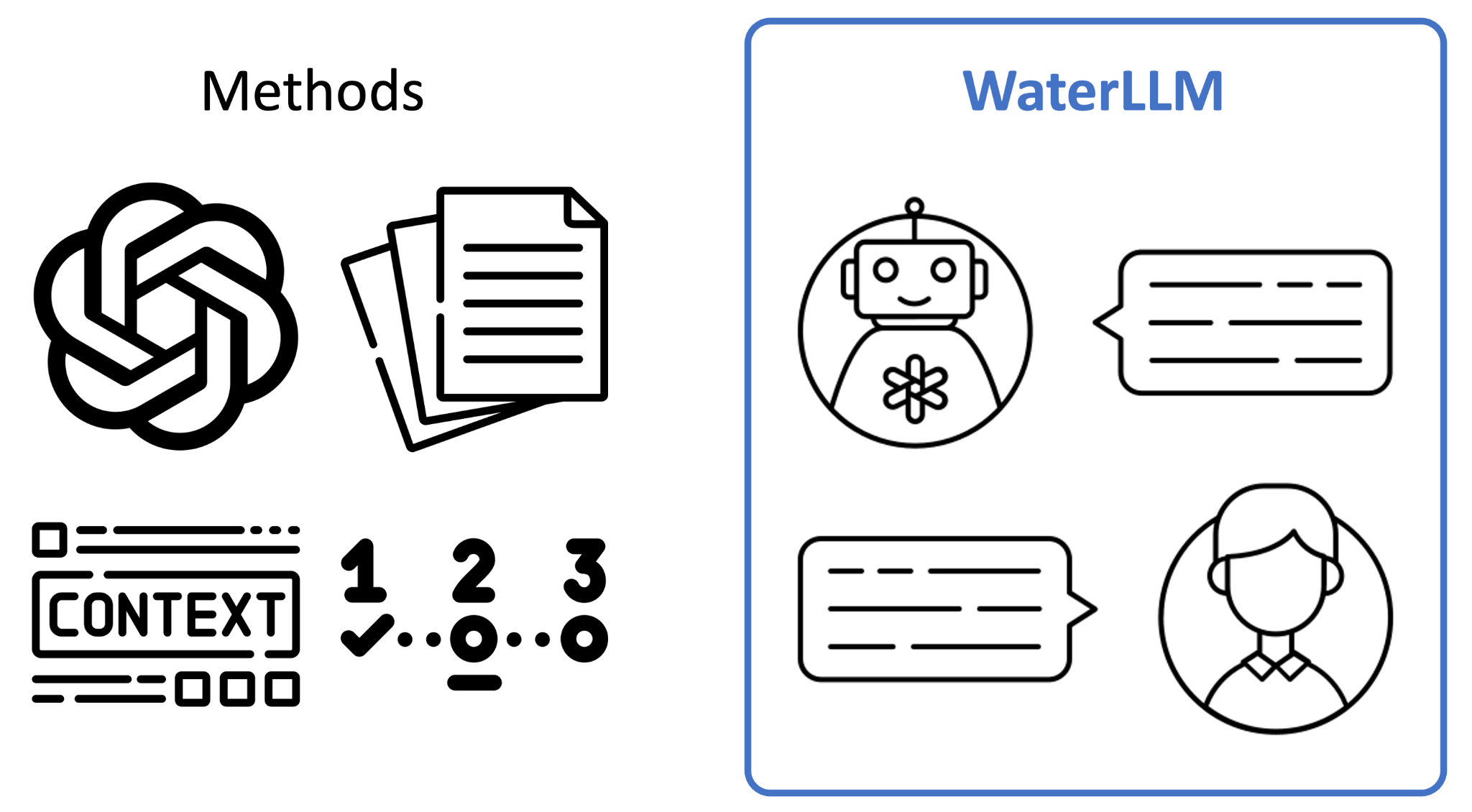} 
    \caption{WaterLLM approach: custom chatGPT with RAG from scientific papers, context and chain-of-thought allowed for interactive dialog with the user anchored to science.
    \label{fig:waterllm1}}
\end{figure}

\begin{figure}[h!]
    \centering
    \includegraphics[width=0.99\textwidth]{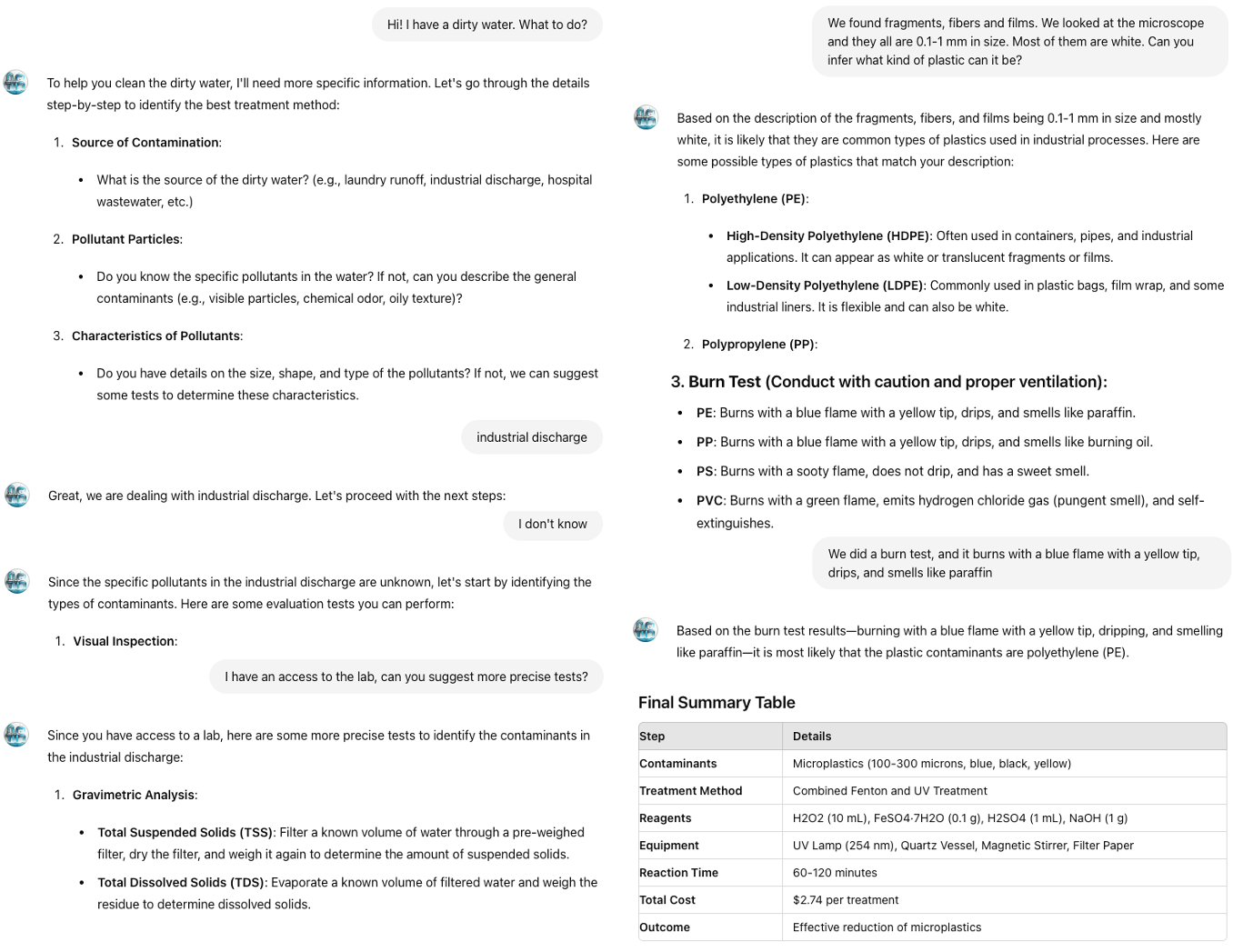} 
    \caption{Sample of the WaterLLM communication with the User.
    \label{fig:waterllm2}}
\end{figure}

\clearpage
\newpage
\section{yeLLowhaMMer: A Multi-modal Tool-calling Agent for Accelerated Research Data Management}\label{sec:yellowhammer}

\textbf{\textit{Authors: Matthew L. Evans, Benjamin Charmes, Vraj Patel, Joshua D. Bocarsly
}}

As scientific data continues to grow in volume and complexity, there is a great need for tools that can simplify the job of managing this data to draw insights, increase reproducibility, and accelerate discovery. Digital systems of record, such as electronic lab notebooks (ELN) or laboratory information management systems (LIMS), have been a great advancement in this area. However, complex tasks are still often too laborious, or simply impossible, to accomplish using graphical user interfaces alone, and any barriers to streamlined data management often lead to lapses in data recording.

As developers of the open-source \emph{datalab} \cite{Evans2024a} ELN/LIMS, we explored how large language models (LLMs) can be used to simplify and accelerate data handling tasks in order to generate new insights, improve reproducibility, and save time for researchers. Previously, we made progress toward this goal by developing a conversational assistant, named Whinchat \cite{Jablonka2023}, that allows users to ask questions about their data. However, this assistant was unable to take action with a user’s data. Here, we developed yeLLowhaMmer, a multimodal large language model (MLLM)-based data management agent capable of taking free-form text and image instructions from users and executing a variety of complex scientific data management tasks.

Our agent is powered by a low-cost commercial MLLM (Anthropic’s Claude 3 Haiku) used within a custom agentic infrastructure that allows it to write and execute Python code that interacts with \emph{datalab} instances via the \texttt{datalab-api} package. In typical usage, a yeLLowhaMmer user might instruct the agent: “Pull up my 10 most recent sample entries and summarize the synthetic approaches used.” In this case, the agent will attempt to write and execute Python code using the \emph{datalab} API to query for the user’s samples in the \emph{datalab} instance and write a human-readable summary. If the code it generates gives an error (or does not give sufficient information), the agent can iteratively rewrite the program until the task is accomplished successfully.

Furthermore, we leveraged the powerful multimodal capabilities of the latest MLLMs to allow for prompts that include visual cues. For example, a user may upload an image of a handwritten lab notebook page and ask that a new sample entry be added to the \emph{datalab} instance. The agent uses its multimodal capabilities to “read” the lab notebook page (even if it is a messy/unstructured page), adds structure to the information it finds by massaging it into the form requested by the \emph{datalab} JSON schema, then writes a Python snippet to ingest the new sample into the \emph{datalab} instance. Notably, we found that even the inexpensive, fast model we used (Claude 3 Haiku) was able to perform sufficiently well at this task, while larger models may be explored in the future to allow for more advanced applications (though with slower speed and greater cost). We believe the capabilities demonstrated by yeLLowhaMmer show that MLLM agents have the potential to greatly lower the barrier to advanced data handling in experimental materials and chemistry laboratories. This proof-of-concept work is accessible on GitHub at \href{https://github.com/bocarsly-group/llm-hackathon-2024}{bocarsly-group/llm-hackathon-2024}, with ongoing work at \href{https://github.com/datalab-org/yellowhammer}{datalab-org/yellowhammer}.

\begin{figure}[h!]
    \centering
    \includegraphics[width=0.8\textwidth]{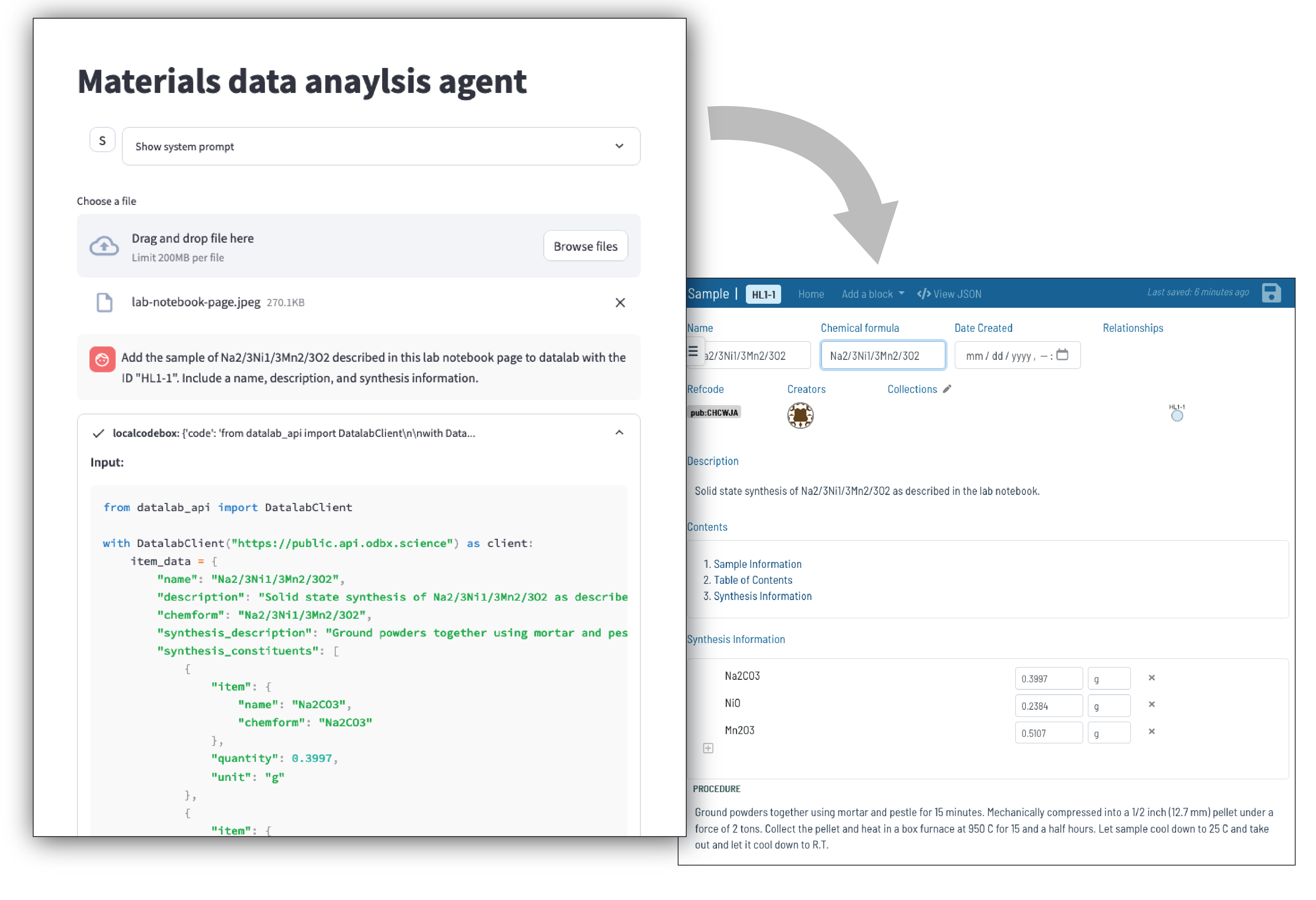} 
    \caption{The yeLLowhaMmer multimodal agent can be used for a variety of data management tasks. Here, it is shown automatically adding an entry into the \emph{datalab} lab data management system based on an image of a handwritten lab notebook page.}
    \label{fig:yellowhammer}
\end{figure}

yeLLowhaMmer was built upon several open-source software packages. The codebox-api Python package was used to set up a local code sandbox that the agent has access to in order to safely read and save files, install Python packages, and run the code generated by the model. The \texttt{datalab-api} Python package was used to interact with \emph{datalab} instances. An MLLM-compatible tool was designed to allow the model to use function-calling capabilities, write, and execute code within the sandbox. LangChain was used as a framework to interact with the commercial MLLM APIs and build the agentic loop. Streamlit was used to build a responsive GUI to show the conversation and upload/download the files from the codebox. A customized Streamlit callback was written to display the code and files generated by the agent in a user-friendly manner.

An interesting challenge in the development of yeLLowhaMmer was the creation of a system prompt that would enable the agent to reliably generate robust code using the \texttt{datalab-api} package, which is a recent library not included in the training of the commercial models at the time of writing. Initially, we copied the existing documentation for the \texttt{datalab-api} into the system prompt, but we found that the code generated by the model was not working very well. Instead, it was helpful to produce a simplified version of the documentation that removed extraneous information and gave a few concrete examples of scripts. Additionally, we provided an abridged version of the \emph{datalab} schemas in the JSON Schema format in the system prompt, which was necessary for the generation of compliant data to be inserted into \emph{datalab}.

Overall, the yeLLowhaMmer system prompt amounts to around 12,000 characters (corresponding to about 3200 tokens using Claude’s tokenizer). Given the large context windows of the current generation of MLLMs (e.g., 200k tokens for Claude 3 Haiku), this size of prompt is feasible for fast, extended conversations involving text, generated code, and images. In the future, we envision that library maintainers may wish to increase the utility of their libraries by maintaining two parallel sets of documentation: the standard human-readable documentation, and an abridged \texttt{agents.txt} (or \texttt{llms.txt} -- \url{https://llmstxt.org/}) file that can be used by ML agents to write high-quality code using that library.

Going forward, we will undertake further prototyping to incorporate MLLM-based agents more tightly into our data management workflows, and to ensure that data curated or modified by such an agent will be appropriately `credited' by, for example, visually demarcating AI-generated content, and providing UI pathways to verify or `relabel' such data in an efficient manner. Finally, we emphasize the great progress made within the last year in MLLMs themselves, which are now able to handle audio and video content in addition to text and images. These will allow MLLM agents to use audiovisual data in real-time to provide new user interfaces. Based on these promising developments, we believe that data management platforms are well-placed to help bridge the divide from physical to digital data recording.

\clearpage
\newpage
\section{LLMads}\label{sec:llmads}

\textbf{\textit{Authors: Sarthak Kapoor, José M. Pizarro, Ahmed Ilyas, Alvin N. Ladines, Vikrant Chaudhary
}}

Parsing raw data into a structured (meta)data standard or schema is a major Research Data Management (RDM) topic. While defining F.A.I.R. (Findable, Accessible, Interoperable, and Reusable) metadata schemas are the key to RDM, these empty fields must be populated. This is typically done in two ways: 
\begin{itemize}
    \item Fill a schema manually using electronic or physical lab notebooks, or 
    \item Create scripts that read the input/output raw files and parse them into the data schema. 
\end{itemize}

The first option is used in a lab setting where data is entered in real-time as it is generated. The second option is used when data files are available, albeit in an incompatible format to fill the schema directly. These can be measurement files coming from instruments or files generated from simulations. Specific parsers for each raw file type can transfer large amounts of data into schemas, making them essential for automation and big-data management. However, implementing parsers for all the possible raw files to a fill schema can be laborious and time-consuming. It requires expert knowledge of the structure of the raw files and regular maintenance to keep up with new versions of raw files. In this work, we attempted to substitute parsers with Large Language Models (LLMs).

We investigated whether LLMs can be used to parse data into structured schemas, thus relieving the need for coding parsers. 
As an example, we used raw files from X-ray diffraction (XRD) measurements from three different instrument vendors (Bruker, Rigaku, and Panalytical). We defined a Pydantic model for our structured data schema and used the pre-trained Mixtral-8x7b from Groq. The data schemas are provided to the LLM using the function-calling mechanism. The schema is constructed by defining a Pydantic base model class and their fields or attributes with well-defined types and descriptions. The LLM tries to extract data for each variable from the raw files that matches these descriptions. Considering the size of the raw files and the token limitation of LLMs, we decided to create the following workflow:

\begin{itemize}
    \item Break the raw file contents into chunks.
    \item Prompt the LLM with the initial chunk.
    \item Generate a response from the LLM and populate the schema.
    \item Prompt the LLM with the next chunk along with the previous response.
\end{itemize}

We found that when populating the schema, the LLM was correctly extracting the values in cases where the data types were \texttt{float} and \texttt{str}. This was the case for the \texttt{XRDSettings} class. However, when parsing values with a data type \texttt{list[float]}, the LLM was often unable to extract the data in the expected format. This occurred when populating \texttt{XRDResults} class with the intensities data. The LLM output included non-numeric characters like \texttt{\textbackslash n} or \texttt{\textbackslash t}, along with hallucinated data values. By providing the previous response along with the new chunk of data in the prompts, we incorporated some degree of context. We found that using smaller chunk sizes led to the rapid replacement of the populated data. Sometimes, the correct data was replaced by hallucinated values.

We used LangChain to build our models and prompt generators. The code is openly available on Github: \url{https://github.com/ka-sarthak/llmads}. Our work uses prompt engineering and function-calling. Future work into tuning the model temperature and fine-tuning could be explored to combat hallucination. Our work also indicates a need for human intervention to verify if the schema was filled correctly and to correct it when necessary. Nevertheless, our prompting strategy proves to be a valuable tool as it manages to initialize the schema properly for non-vectorial fields, all at the minimal effort of providing a structured schema and the raw files.

\newpage
\section{NOMAD Query Reporter: Automating Research Data Narratives}\label{sec:nomad-query-reporter}

\textbf{\textit{Authors: Nathan Daelman, Fabian Schöppach, Carla Terboven, Sascha Klawohn, Bernadette Mohr
}}

Materials science research data management (RDM) platforms and structured data repositories contain large numbers of entries, each composed of property-value pairs. Users query these repositories by specifying property-value pairs to find entries matching specific criteria.
While this guarantees that all returned entries have at least the queried properties, they do not provide context or insights into the structure and variety of other data present in them.

Traditionally, it is up to the data scientist to examine the returned properties and interpret the overall response. 
To assist with this task, we use a LLM to create context-aware reports based on the properties and their meanings.
We build and tested our ``Query Reporter'' \cite{Scheidgen2024} prototype on the NOMAD \cite{NomadQueryReporter} repository, which stores heterogeneous materials science data, including metadata on scientific methodologies, atomistic structures, and materials properties.

We developed the NOMAD Query Reporter \cite{Scheidgen2024} as a proof-of-concept. It fetches and analyzes entries and produces a summary of the used methodology and standout results. It does so in a scientific style, lending itself as the basis for a “methods” section in a journal article. Our agent uses a retrieval-augmented generation (RAG) approach \cite{Gao2024b}, which enriches an LLM’s knowledge of external DB data without performing retraining or fine-tuning. To allow its safe application on private and unpublished data, we use a self-hosted Ollama instance running Meta’s Llama3 70B \cite{Touvron2023} model.

\begin{figure}[hbt!]
    \centering
    \includegraphics[width=0.8\textwidth]{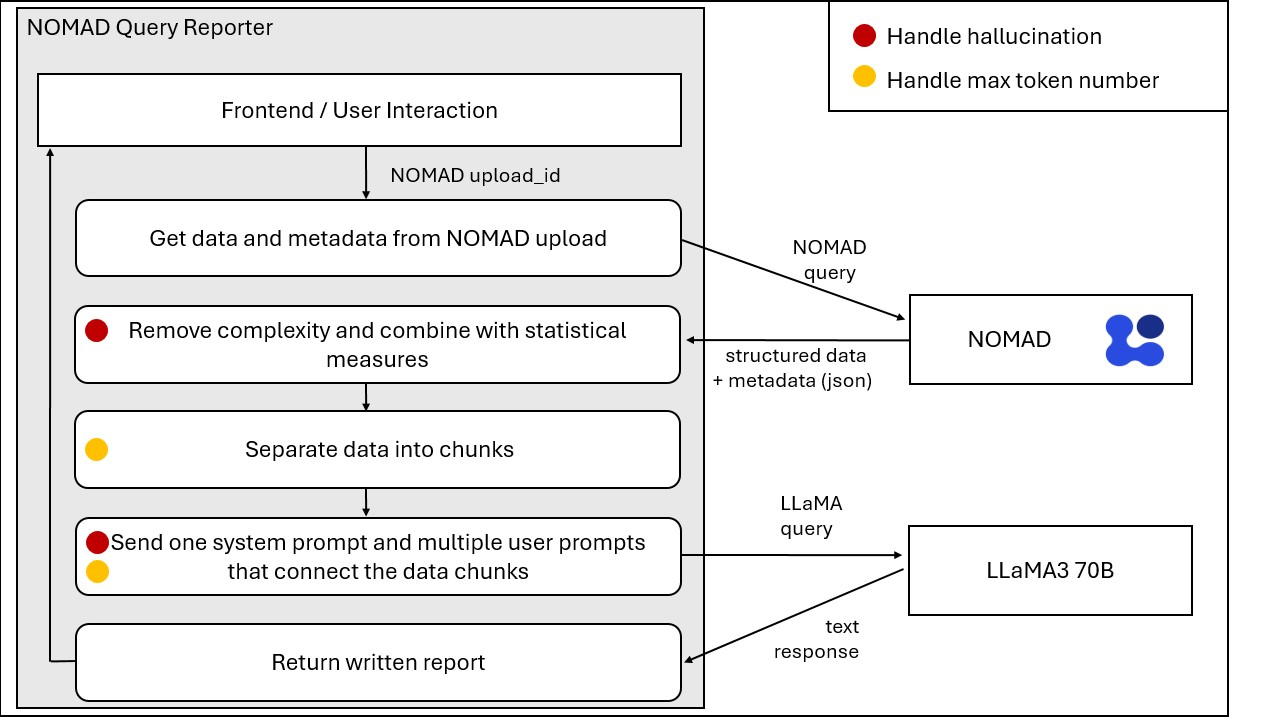} 
    \caption{Flowchart of the Query Reporter usage, including the back-end interaction with external resources, i.e., NOMAD and Llama. Intermediate steps managing hallucinations or token limits are marked in red and orange, respectively.
    \label{fig:nomad_query_reporter}}
\end{figure}

We tested the agent on publicly available data from NOMAD. To manage Llama3’s context window of 8,000 tokens, the entries are collected as rows into a Pandas dataframe. Each row (e.g., entry) is individually passed on to the LLM via the chat-completion API. Instead of a single message, it accepts a multi-turn conversation that simulates several demarcated roles. We use the “system” and “user” roles of the chat to reinforce the retention of parts of the previous summary. This approach generally conforms to the Naive RAG category in Gao et al.’s classification \cite{Gao2024b}. For a step-by-step overview, see \autoref{fig:nomad_query_reporter}.

We used two kinds of data for testing: (a) homogeneous, property-value pairs of computational data; and (b) heterogeneous text typed properties of solar cell experiments, often formatted as in-text tables. We engineered different prompts for each kind. The agent performed better on the homogeneous than the heterogeneous data. Here, summaries would often suffer from irrelevant threads, or even hallucinations. We theorize that homogeneous data maps more consistently onto our predefined dataframe columns, which aids the LLM in interpreting follow-up messages. Still, we could not improve the performance for heterogeneous data within the hackathon.

In short, the NOMAD Query Reporter demonstrates that the combined approach of filtering and RAG can effectively summarize collections of limited-size, structured data directly stored in research data repositories, allowing for automated drafting of methods and setups for publications at a consistent level of quality and style. These results suggest applicability for other well-defined materials science APIs, such as the OPTIMADE standard \cite{Evans2024}. Follow-up work includes investigating the impact of Advanced RAG strategies \cite{Gao2024b}.

\subsection{Acknowledgements}
N.~D., S.~K., and B.~M. are members of the NFDI consortium FAIRmat, funded by the German Research Foundation (DFG, Deutsche Forschungsgemeinschaft) in project 460197019. F.~S. acknowledges funding received from the SolMates project, which has been supported by the European Union’s Horizon Europe research and innovation program under grant agreement No 101122288. C.~T. is supported by the German Federal Ministry of Education and Research (BMBF, Bundesministerium für Bildung und Forschung) in the framework of the project Catlab 03EW0015A.

\newpage
\section{Speech-schema-filling: Creating Structured Data Directly from Speech}\label{sec:speech-schema-filling}

\textbf{\textit{Authors: Hampus Näsström, Julia Schumann, Michael Götte, José A. Márquez
}}

\subsection{Introduction}
\begin{figure}[h!]
    \centering
    \includegraphics[width=0.99\textwidth]{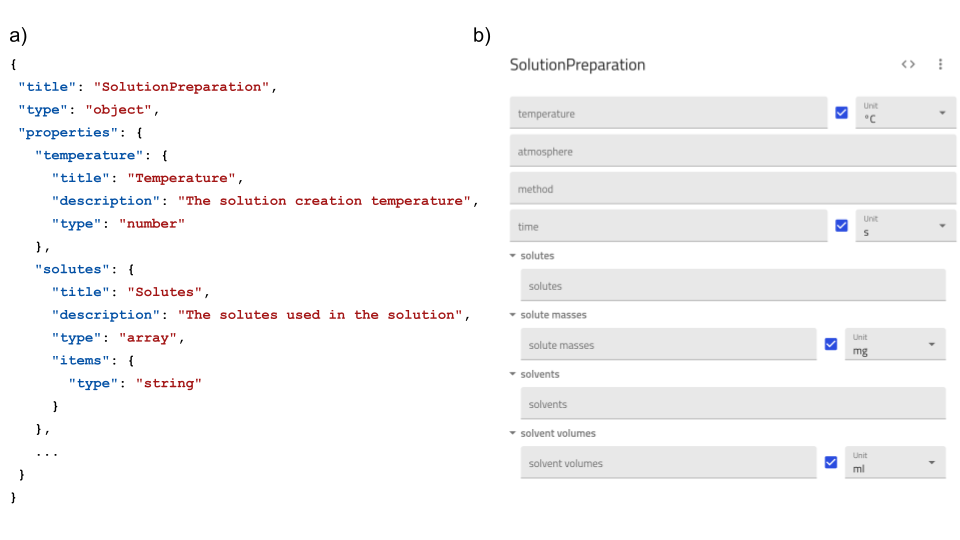} 
    \caption{a) Part of a JSON Schema defining a data structure for a solution preparation. b) The schema converted to an ELN form in NOMAD \cite{Scheidgen2023}.
    \label{fig:nomad_json_schema}}
\end{figure}

As the amount of materials science data being created increases, so do the efforts to make this data Findable, Accessible, Reusable, and Interoperable (FAIR) \cite{Wilkinson2016}. One pragmatic approach to creating FAIR data is by defining so-called data schemas for the various types of data being recorded. These schemas can then be used in both input tools like electronic lab notebooks (ELNs) and storage solutions like data repositories to create structured data. One widely adopted standard for writing data schemas is the so-called JSON Schema \cite{JSON}. JSON Schema allows us to define objects, such as, for example, a solution preparation experiment in the lab, with properties such as temperature and a list of solutes and solvents (see \autoref{fig:nomad_json_schema}a). These schemas can then be used to create forms in an ELN like NOMAD \cite{Scheidgen2023} (see \autoref{fig:nomad_json_schema}b). However, in a lot of lab situations, such as when working inside a glovebox, it is difficult to i) navigate the ELN and select the right form and ii) actually fill in the form with experimental data. In our experience, this usually results in users having to record their data later from memory or even in data not being recorded at all.

We propose a solution for this using LLMs to:
\begin{itemize}
    \item Converting spoken language in the lab into text using advanced speech recognition technologies, such as OpenAI's Whisper. 
    \item Based on the text select and fill the appropriate schema, enabling accurate data capture without manual text entry.
\end{itemize}

\subsection{Speech recognition}
The first step of converting speech into structured data is to record the speech and transcribe it into text. There are multiple options for recording audio, and the best solution depends on both the hardware and the operating system used. We use the \texttt{Recognizer} class from the Python package \texttt{SpeechRecognition} \cite{SpeechRecognition} to perform the recording only, while leaving the actual speech recognition to OpenAIs Whisper. There is an open version of this available for download through the Python package \texttt{openai-whisper} \cite{OpenaiWhisper}.

\subsection{Text to structured data}
Once the speech has been converted to text we need to use this text to i) select the appropriate schema and ii) fill the schema.
For this, we make use of a common feature in LLMs called “function calling”, “API calling”, or “tool use”.
This has been developed to allow LLMs to make valid API calls and, since OpenAPI is validated by it, uses JSON Schema to define the possible “functions”, or “tools”, that can be outputted.
Since we use JSON Schema to define our data schemas we can simply supply these as the tools for the LLM to use.
One way to do this in Python is to write a Pydantic schema for each of the data schemas and then convert this to a JSON Schema.
For the Llama model we used, the python package \texttt{langchain-experimental} \cite{LangChainc} has an \texttt{OllamaFunctions} class that can be imported from the \texttt{llms.ollama\_functions} sub-package.
This class can then be used to instantiate the model and add the valid data schemas.
Here is an example using LangChain and Llama3:

\begin{lstlisting}
from langchain_experimental.llms.ollama_functions import OllamaFunctions
model = OllamaFunctions(model="llama3:70b", base_url='...', format='json')
model = model.bind_tools(
   tools=[
       {
           "name": "solution_preparation",
           "description": "Schema for solution preparation",
           "parameters": SolutionPreparation.schema(),
       },
       {
           "name": "powder_scaling",
           "description": "Schema for powder scaling",
           "parameters": Scaling.schema(),
       }
   ],
)
\end{lstlisting}

Where \texttt{SolutionPreparation} and \texttt{Scaling} are Pydantic models for our desired data schemas. Finally, this can be chained together with a prompt template and used to process the transcribed audio from before:

\begin{lstlisting}
prompt = PromptTemplate.from_template(...)
chain_with_tools = prompt | model
response = chain_with_tools.invoke(transcribed_audio)


For langchain the selected schema can be retrieved from:
schema = response.additional_kwargs['function_call']['name']
And the filled instance from:
instance = json.loads(
response.additional_kwargs['function_call']['arguments'])
\end{lstlisting}

For LangChain the selected schema can be retrieved from:

\begin{lstlisting}
schema = response.additional_kwargs['function_call']['name']
\end{lstlisting}

And the filled instance from:

\begin{lstlisting}
instance = json.loads(
response.additional_kwargs['function_call']['arguments'])
\end{lstlisting}

A detailed example of the text-to-structured data can be found as an iPython notebook on \url{github.com/hampusnasstrom/speech-schema-filling} together with an implementation of the audio recording and transcribing using Whisper. In conclusion, we believe that LLMs can be useful in labs where traditional ELNs are hard to operate by transcribing speech, selecting appropriate schemas, and filling the schemas to ultimately create structured data directly from speech.

\subsection{Acknowledgements}
H.N., J. S., and J. A. M. are part of the NFDI consortium FAIRmat funded by the Deutsche Forschungsgemeinschaft (DFG, German Research Foundation) – project 460197019.

\clearpage
\newpage
\section{Leveraging LLMs for Bayesian Temporal Evaluation of Scientific Hypotheses}\label{sec:lk99-hypothesis}

\textbf{\textit{Authors: Marcus Schwarting
}}

\subsection{Introduction}
Science is predicated on empiricism, and a scientist uses the tools at their disposal to gather observations that support the veracity of their claims.
When one cannot gather firsthand evidence for a claim, they must rely on their own assessment of evidence presented by others.
However, the scientific literature on claim is often large and dense (particularly for those without domain expertise), and the scientific consensus on the veracity of a claim may drift over time.
In this work we consider how large language models (LLMs), in conjunction with temporal Bayesian statistics, can rapidly provide a more holistic view of a scientific inquiry.
We demonstrate our approach on the hypothesis that the material LK-99, which went viral after its discovery in April 2023, is in fact a room-temperature superconductor.

\subsection{Background}

Scientific progress requires a researcher to iteratively update their prior understanding based on new observations. 
The process of updating a statistical prior based on new information is the backbone of Bayesian statistics and is routinely used in scientific workflows \cite{Settles2009}.
Under such a model, a hypothesis has an inferred probability $P_H \in (0,1)$ of being true and will never be completely discarded ($P_H=0$) or accepted ($P_H=1$) but may draw arbitrarily close to these extrema.
This inferred probability from a dataset is also a feature of assessing the power of a claim using statistical hypothesis tests \cite{Lehmann1986}, which are commonly used across most scientific disciplines.

Modelling the veracity of a scientific claim as a probability also stems from the work of philosopher Karl Popper \cite{Popper2005}.
Popper posits that a scientific claim should be falsifiable, such that it can be refuted by empirical evidence.
If a scientific claim cannot be refuted, Popper argues that it is not proven, but gains further credibility from the scientific community.
Scientific progress is not made by proving claims, but instead by hewing away false claims through observation until only those that withstand repeated scrutiny remain.

The history of science is littered with discarded hypotheses.
Some of these claims were believed to be true for centuries before being dismissed (including geocentrism, phrenology, energeticism, and spontaneous generation).
In this work, we focus on a claim by Lee, Kim, and Kwon that they had successfully synthesized a room-temperature superconductor, which they called LK-99 \cite{Lee2023}.
Such a material would necessitate altering the existing theory of superconductors at a fundamental level and would enable innovations that are currently beyond reach.
LK-99 went viral in summer 2023 \cite{Chang2023}, but replication efforts quickly ran into issues \cite{Garisto2023}.
Since the initial claim was made in April 2023, roughly 160 works have been published, and the scientific consensus now appears established: LK-99 is not a room-temperature superconductor.

\subsection{Methods}
Our dataset consists of the 160 papers on Google Scholar, ordered by publication date, that are deemed relevant to the hypothesis ``LK-99 is a room-temperature superconductor.''
For each paper abstract in our dataset, we construct an LLM prompt as follows to perform a zero-shot natural language inference operation \cite{Liu2021}:

\begin{displayquote}
Given a Hypothesis and an Abstract, determine whether the Hypothesis is an ‘entailed’, ‘neutral’, or ‘contradicted’ by the Abstract. \textbackslash n Hypothesis: \{claim\} \textbackslash n Abstract: \{abstract\}
\end{displayquote}

We then wrote a regular expression to check the LLM response to make an assertion for each publication: “entailment,” “neutrality,” or “contradiction.” 
We use Llama2 to make these assertions on all 160 papers, which complete in under five minutes (on a desktop with an Nvidia GTX-1070 GPU).

Next, we constructed a temporal Bayesian model where, starting from an initial Gaussian prior $N(\mu_P,\sigma_P^2)$ that models the likelihood of accepting the hypothesis, we can update with a Gaussian likelihood $N(\mu_L,\sigma_L^2)$.
Our likelihood is a Gaussian designed so that for a given publication, “entailment” acceptance probability higher, “contradiction” pushes the acceptance probability lower, and “neutrality” leaves the acceptance probability the same.
Our likelihood probability is further weighted according to the impact factor of the journal, with an impact factor floor set to accommodate publications that have not passed peer review.
Retroactively, we could also weight by the number of citations, however we omit this feature in our analysis since this is inherently a post-hoc metric and would violate our temporal assessment.
We update our Gaussian prior using the equations \cite{Murphy2007}:

$$\mu_P \leftarrow \biggr( \frac{1}{\sigma_P^2} + \frac{1}{\sigma_L^2} \biggr)^{-1} \biggr( \frac{\mu_P}{\sigma_P^2} + \frac{\mu_L}{\sigma_L^2} \biggr);\; \sigma_P^2 \leftarrow \biggr( \frac{1}{\sigma_P^2} + \frac{1}{\sigma_L^2} \biggr)^{-1}$$

We are also able to specify the initial probability associated with the hypothesis ($\mu_P$), as well as how flexible we are in changing our perspective based on new evidence ($\sigma_P^2$).
We select two initial probabilities: 50\% and 20\%, and fix standard deviations $\sigma_P^2$ and $\sigma_L^2$.
Assuming either “contradiction” or “entailment”, the update due to $\mu_L$ then scales linearly with the publication impact factor.
Finally, we can compare our temporal probability assessment with probabilities provided by the betting platform Manifold Markets \cite{ManifoldMarkets2024}, where players bet on the outcome of a successful replication of the LK-99 publication results.

\subsection{Results}
\begin{figure}[h!]
    \centering
    \includegraphics[width=0.99\textwidth]{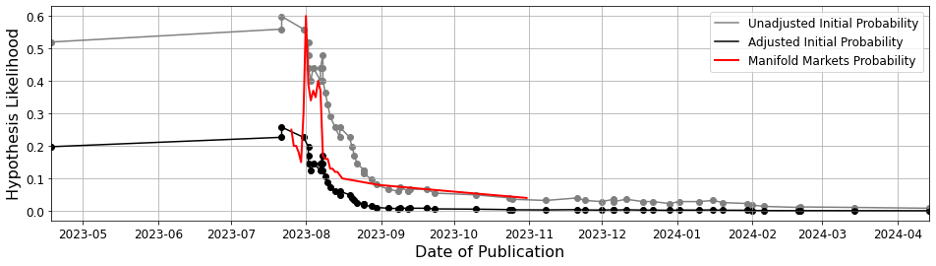} 
    \caption{Likelihood of accepting the hypothesis “LK-99 is a room-temperature superconductor” via three approaches, from April 15, 2023 to April 15, 2024.
    The unadjusted initial probability (set to 50\%) is shown in gray, the adjusted initial probability (set to 20\%) is shown in black, and the probability according to the Manifold Markets online betting platform is shown in red.
    \label{fig:LK99_temporal}}
\end{figure}

We find that our temporal Bayesian probabilities, with an adjusted initial prior, can mirror the Manifold Markets probabilities with two interesting divergences.
While the initial probability starts at around 20\% for both, the temporal Bayesian approach never goes above 30\%.
By contrast, the betting market, following the hype and virality of the LK-99 publication, reaches a peak at around 60\%.
Furthermore, while the betting market has a long tail of low probability starting in mid-August 2023, our approach more quickly disregards the hypothesis based on a continuing accumulation of studies showing that the LK-99 findings could not be replicated.
Our temporal Bayesian model with the adjusted initial prior reaches a probability below 1\% by mid-September 2023, but never entirely dismisses the chance that the hypothesis is true.
\autoref{fig:LK99_temporal} showcases these results.

While we specifically select initial probabilities or 20\% and 50\%, both trajectories end with a steadily shrinking probability of accepting the hypothesis.
Our initial prior probability of 50\% mimics an unbiased observer with no knowledge about whether the hypothesis should be accepted or rejected. A prior probability of 20\% could be considered a reasonable guess for an observer biased by a baseline understanding and suspicion of the claims and their corresponding evidence.
Such an initial guess is admittedly subjective, as is the degree to which new information affects one’s inherent biases.
We treat these settings as presets, however these are trivial for others to configure and assess based on their background.
Finally, for claims with established scientific consensus, our approach is guaranteed to asymptotically approach that consensus where the rate of convergence varies according to these initial presets.

\subsection{Conclusion}

Carefully validating a claim based a body of scientific literature can be a time-consuming and challenging prospect, especially without domain expertise.
In this work, we demonstrate how a claim might be evaluated using a temporal Bayesian model based on a literature evaluation using natural language inference.

We show how our aggregated literature predictions allow us to quickly reject the hypothesis that LK-99 is a room-temperature superconductor.
In the future, we hope to apply this approach to other scientific claims, including those with debates that are ongoing and as well as claims have an established scientific consensus.
In general, we hope this approach will allow a researcher to quickly measure the scientific community’s confidence in a claim, as well as aid the public in assessing both the veracity of a claim and the change in confidence driven by continued experimentation and observation.

\newpage
\section{Multi-Agent Hypothesis Generation and Verification through Tree of Thoughts and Retrieval Augmented Generation}\label{sec:hyp}

\textbf{\textit{Authors: Aleyna Beste Ozhan, Soroush Mahjoubi
}}
\subsection{Introduction}

\begin{figure}[h!]
    \centering
    \includegraphics[width=0.99\textwidth]{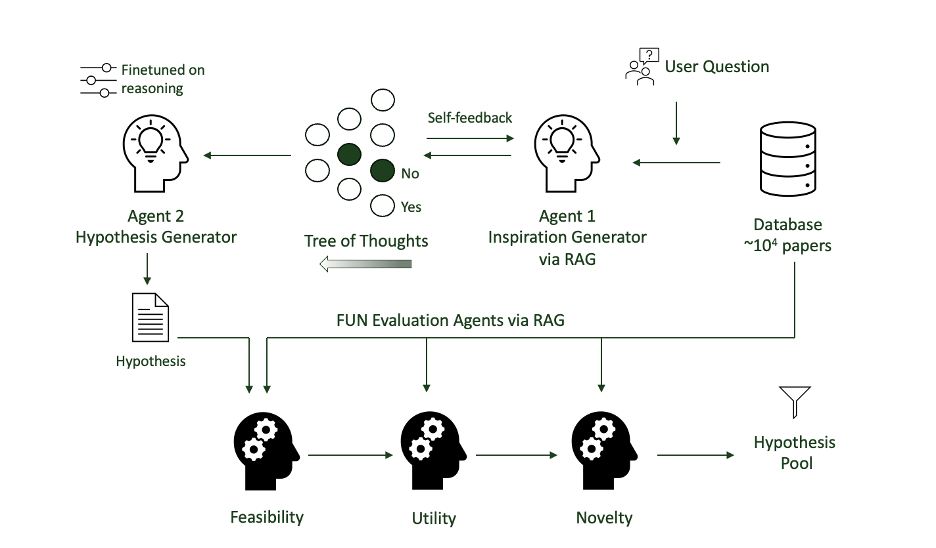} 
    \caption{Multi-Agent Hypothesis Generation and Verification Pipeline
    \label{fig:multi_agent_hyp_gen}}
\end{figure}

Our project, developed during the “LLM Hackathon for Applications in Materials and Chemistry,” aims to accelerate scientific hypotheses and enhance the creativity of scientific inquiry.
We propose using a multi-agent system of specialized large language models (LLMs) to streamline and enrich hypothesis generation and verification in materials science. This approach leverages diverse, fine-tuned LLM agents collaborating to generate and validate novel hypotheses more effectively. While similar pipelines have been proven useful in the social sciences \cite{Yang2023}, to the best of our knowledge, this work marks the first adaptation of such an approach to hypothesis generation in materials science.
As illustrated in \autoref{fig:multi_agent_hyp_gen}, The system includes agents such as a background provider, an inspiration generator, a hypothesis generator, and three evaluators.
Each agent plays a crucial role in formulating and assessing hypotheses, ensuring only the most viable and compelling ideas are developed.
This innovative approach fosters an environment conducive to scientific inquiries.

\subsection{Methodology}
\paragraph{Background Extraction}
The background extraction module is designed to search through a vast database for relevant information directly related to the user’s query. This module employs advanced embedding-based retrieval techniques to identify and extract pertinent corpus. As new papers and findings are added to the repository, the system dynamically updates, ensuring the use of the most current and relevant information.

\paragraph{Inspiration Generator Agent}
The inspiration generator agent leverages extensive background data to effectively formulate inspirations using a Retrieval Augmented Generation (RAG) mechanism.
Serving as the strategic core of the hypothesis generation process, it draws inspiration from a broad spectrum of sources to spawn diverse hypotheses, similar to the branching structure of a "Tree of Thoughts (ToT)" \cite{Yao2024}.
The agent samples "k" candidates as possible solutions, evaluates their effectiveness through self-feedback, and votes on the most promising candidates.
The selection is narrowed down to "b" promising options per step, with this structured approach helping the agent systematically refine its solutions.

\paragraph{Hypothesis Generator}
Based on the background information and the inspirations, this module generates meaningful research hypotheses.
It is fine-tuned on reasoning datasets, such as Atlas, which encompasses various types of reasoning including deductive and inductive reasoning, cognitive biases, decision theory, and argumentative strategies \cite{AtlasUnified}.

\subsection{Evaluator Agents}
Once a hypothesis is generated, a RAG mechanism is used to fetch relevant abstracts from the dataset to evaluate the hypothesis. The evaluation is based on three critical aspects adopted from existing literature for the materials science domain:

\textit{Feasibility:} Assesses whether the hypothesis is realistically grounded and achievable based on current scientific knowledge and technology.

\textit{Utility:} Evaluates the practical value of the hypothesis, considering its potential to solve problems, enhance experimental design, or lead to beneficial exploratory paths.

\textit{Novelty:} Measures the uniqueness and originality of the hypothesis, encouraging the generation of innovative ideas that advance scientific understanding.

\subsection{Case Study}

Our case study focuses on sustainable practices in concrete design at the material supply level. We  processed 66,000 abstracts related to cement and concrete, converting them into a Chroma vector database using the sentence transformer all-MiniLM-L6-v2 \cite{MiniLML6}. This model maps abstracts to a 384-dimensional dense vector space. We then queried the embedding-based retrieval system with questions related to material-level solutions for sustainability in concrete, retrieving 10,000 relevant abstracts.

\textit{Example Query:} "How can we incorporate industrial wastes and byproducts as supplementary cementitious materials to promote sustainability while maintaining its fresh and hardened properties such as strength?"

\subsection{Results}
In the initial phase of our ``Tree of Thoughts" structure, we generated approximately 5,000 inspirations.
These inspirations were refined to around 1,000 through a distillation step.
The hypothesis generator, GPT-3.5 Turbo, fine-tuned on 13,000 data points from the AtlasUnified/Atlas-Reasoning dataset, produced one hypothesis per inspiration.
The evaluation process involved three agents assessing feasibility, utility, and novelty (FUN) using embedding-based retrieval to identify relevant abstracts.
For enhanced precision, GPT-4 was employed during the evaluation stages.
Ultimately, hypotheses that withstood all evaluation stages were included in the hypothesis pool.
Out of the initial 1,000 hypotheses, 243 passed the feasibility filter, 175 were deemed useful, and only 12 were found to be highly novel.
The 12 hypotheses deemed feasible, novel, and useful are listed in \autoref{tab:hypothesis_pool}.

\subsection{Future Directions: Adaptability to Other Material Systems and Cross-Domain Applications}
To apply it to another material system, the first background query would be modified to target the new material of interest, and a database relevant to that material system would be employed.
Also, this framework is not limited to materials science; it can be applied across various domains.
For example, ideas generated from civil engineering could inspire hypotheses in materials science.
A background provider querying civil engineering databases might produce inspirations that, when evaluated by our multi-agent system, lead to innovative hypotheses in materials science.
Similarly, within the domain of materials science, inspirations generated based on concrete research could be used to develop hypotheses for other materials, such as ceramics or composites.
This cross-pollination of ideas can foster creativity and drive breakthroughs by applying concepts from one domain to another.


\begin{table}[h!]
\caption{Final hypothesis pool for the study of Section~\ref{sec:hyp}}
\label{tab:hypothesis_pool}
\begin{tabular}{|p{0.6cm}|p{15cm}|}
\hline
\textbf{No.} & \textbf{Hypothesis} \\ \hline
\textbf{1} & 
Incorporating Stainless Steel (SS) micropowder from additive manufacturing into cement paste mixtures can improve the mechanical strength and durability of the mixture, with an optimal addition of 5\% SS micropowder by volume. 
\\ \hline
\textbf{2} & 
The use of synthesized zeolites in self-healing concrete can significantly improve the durability 
and longevity of concrete structures. 
\\ \hline
\textbf{3} & 
The utilization of municipal solid waste (MSW) in cement production by integrating anaerobic digestion and mechanical-biological treatment to produce refuse-derived fuel (RDF) for cement kilns can reduce environmental impacts, establish a sustainable waste-to-energy solution, and create a closed-loop process that aligns waste management with cement production for a more sustainable future.
\\ \hline
\textbf{4} & 
The use of smart fiber-reinforced concrete systems with embedded sensing capabilities can revolutionize infrastructure monitoring and maintenance by providing real-time feedback on structural health, leading to safer and more resilient built environments.
\\ \hline
\textbf{5} & 
The use of advanced additives or nanomaterials in geopolymer well cement can enhance its mechanical properties and durability, leading to more reliable CO2 sequestration projects.
\\ \hline
\textbf{6} & 
The use of carbonated steel slag as an aggregate in concrete can enhance the self-healing performance of concrete, leading to improved durability and longevity.
\\ \hline
\textbf{7} & 
The synergistic effect of combining different pozzolanic materials with varying particle sizes and reactivities can lead to the development of novel high-performance concrete formulations with superior properties.
\\ \hline
\textbf{8} & 
Smart bio concrete incorporating bacterial silica leaching exhibits superior strength, durability, and reduced water absorption capacity compared to traditional concrete.
\\ \hline
\textbf{9} & 
Novel eco-concrete formulation developed by combining carbonated-aggregates with other sustainable materials like volcanic ash or limestone powder can create a carbon-negative concrete with superior mechanical strength, durability, and thermal conductivity.
\\ \hline
\textbf{10} & 
The use of nano-enhanced steel fiber reinforced concrete (NSFRC) will result in a significant improvement in the mechanical properties, durability, and crack resistance of concrete structures compared to traditional steel fiber reinforced concrete.
\\ \hline
\textbf{11} & 
The combined addition of silica fume (SF) and nano-silica (NS) can further enhance the sulphate and chloride resistance to higher than possible with the single addition of SF or NS.
\\ \hline
\textbf{12} & 
The utilization of oil shale fly ash (OSFA) in concrete production can be optimized to develop sustainable and high-performance construction materials.
\\ \hline
\end{tabular}
\end{table}

\newpage
\section{ActiveScience}\label{sec:activescience}

\textbf{\textit{Authors: Min-Hsueh Chiu
}}

\subsection{Introduction}
Humans have conducted material research for thousands of years, yet the vast chemical space, estimated to encompass up to 10\textsuperscript{60} compounds, remains unexplored. Traditional research methods often focus on incremental improvements, making the discovery of new materials a slow process. As data infrastructure has developed, data mining techniques have increasingly accelerated material discovery. However, three significant obstacles hinder this process: First, the availability and sparsity of data present a major challenge. Comprehensive and high-quality datasets are essential for effective data mining, yet materials science suffer from limited data availability. Second, each database typically consists of specific types of quantitative properties, which may not fully meet researchers' needs. This fragmentation and specialization of databases can impede the holistic analysis necessary for breakthrough discoveries. Third, scientists usually focus on certain materials and related articles, decreasing the likelihood of deeply exploring diverse literature that reports potential materials or applications not yet utilized in their specific field. This siloed approach further limits the scope of discovery and innovation.

Unlike the intrinsic properties found in databases, scientific articles provide unstructured but higher- level information, such as applications, material categories, and properties that might not be explicitly recorded in databases. Additionally, these texts include inferences and theories proposed by domain experts, which are crucial for guiding research directions. The challenge lies in automatically extracting and digesting this unstructured text into actionable insights. This process typically requires experienced experts, creativity, and a measure of luck to identify the next desirable candidate. These challenges motivate the potential of utilizing large language models to parse scientific reports and integrate the extracted information into knowledge graphs, thereby constructing high-level insights.

\begin{figure}[h!]
    \centering
    \includegraphics[width=\textwidth]{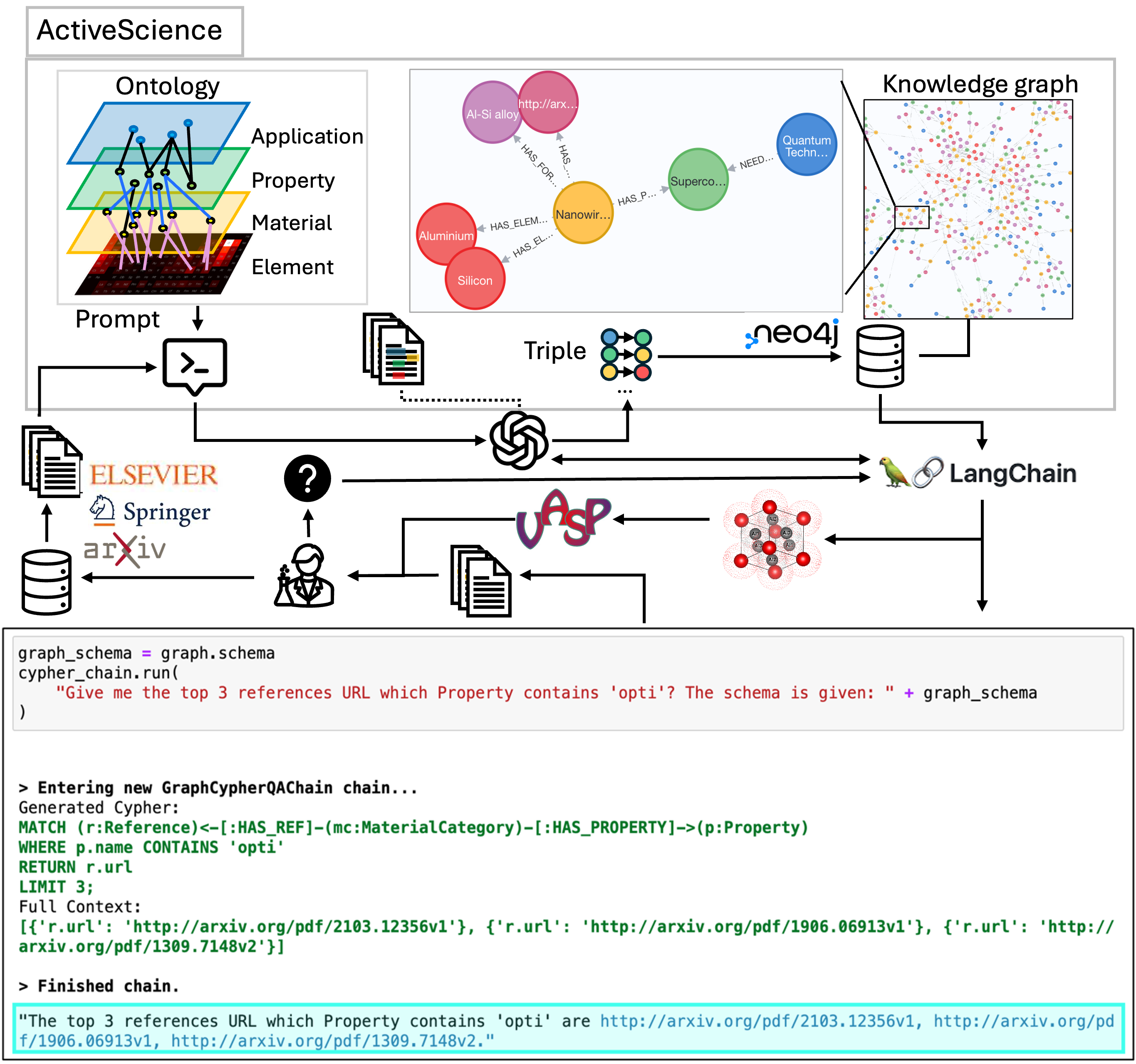} 
    \caption{A Schematic illustration of ActiveScience architecture and its potential applications. Code snippet demonstrating the use of LangChain.
    \label{fig:activescience1}}
\end{figure}

\subsection{Approach}
The Python-based ActiveScience framework consists of three key functionalities: data source API, large language model, and graph database. LangChain is employed for downstream applications within this framework. The schematic pipeline is illustrated in Figure 1.Notably, this framework is not restricted to the specific packages or APIs used in this demonstration; alternative tools that provide the required functionality and input/output can also be integrated.

ArXiv APIs were used to retrieve scientific report titles, abstracts, and URLs. This demonstration focused on reports related to alloys. Consequently, the string “cat:cond-mat.mes-hall AND ti:alloy” was queried in the ArXiv APIs, which returned the relevant articles. GPT-3.5 Turbo models was used to access the large language model. The system's role was defined as: “You are a material science professor and want to extract information from the paper's abstract.” The provided prompt was: “Given the abstract of a paper, can you generate a Cypher code to construct a knowledge graph in Neo4j? ...” along with the designated ontology schema. The generated Cypher code is then input into Neo4j, which is used to ingest the entity relationships, store the knowledge graph, and provide robust visualization and querying interfaces.

\subsection{Results}
With the implemented knowledge graph, GraphCypherQAChain module from LangChain was emploied to perform retrieval-augmented generation.
For instance, when asked, “Give me the top 3 references URLs where the Property contains 'opti'?” GraphCypherQAChain automatically generates a Cypher query according to the designated schema, executes it in Neo4j, and ultimately returns the relevant answer, as shown in the right bottom box in \autoref{fig:activescience1}.
Although this demonstration used a simple question, more complex queries can be processed using this framework.
However, handling such queries effectively will require more refined prompting techniques.

\subsection{Conclusion and future works}
The work demonstrates the pipeline of integrating large language model, knowledge graph, and a question- answering interface. Several aspects of this framework can be enhanced. Firstly, using domain-specific language models in the materials domain could improve entity and relationship recognition. Secondly, enhancing entity resolution and data normalization could lead to a more concise and informative knowledge graph, thereby improving the quality of answers. Thirdly, designing more effective prompting strategies, such as chain-of-thought prompting, could enhance the quality of both answers and code generation.

\subsection{Data and Code Availability}
All code and data used in this study are publicly available in the following GitHub repository: \url{https://github.com/minhsueh/ActiveScience}

\newpage
\section{G-Peer-T: LLM Probabilities For Assessing Scientific Novelty and Nonsense}\label{sec:g-peer-t}

\textbf{\textit{Authors: Alexander Al-Feghali, Sylvester Zhang}}

Large language models (LLMs) and foundation models have garnered significant attention lately due to their natural language programmability and potential to parse high-dimensional data from reactions to the scientific literature \cite{Jablonka2024c}. While these models have demonstrated utility in various chemical and materials science applications, we propose leveraging their designed strength in language processing to develop a first pass peer review system for materials science research papers \cite{Boiko2023}.

Traditional-gram tests such as BLEU or ROUGE, as well as X-of-thought LLM-based evaluations, are not sensitive enough for creativity or diversity in scientific writing \cite{Liu2023}. Our approach utilizes the learned probabilistic features to establish a baseline for typical scientific language in materials science, based on fine-tuning on materials science abstracts through a historical train-test split. New abstracts are scored by their weighted-average probabilities, identifying those that deviate from the expected norms, flagging both possibly innovative or potentially nonsensical works.

As a proof-of-concept in this direction, we fine-tuned two models: OPT (6.7B) and TinyLLama (1.1B) using the Huggingface PEFT library's Low Rank Adapters (LoRA) to access the log probabilities of the abstracts, which is not typically accessible for modern API services \cite{Mangrulkar2022, Zhang2024, Zhang2022}. Our results come with the usual caveats for small models with small computational costs.

We curated a dataset of 6000 abstracts from PubMed, published between 2017--2020, focusing on Materials Science and Chemistry \cite{NCBI2024}. The models were fine-tuned over 200 steps using this dataset. We compared highly cited papers ($>$200 citations) with those of average citation counts. Our preliminary findings suggest that higher-cited papers exhibit less ``typical'' language use, with mean log probabilities of -2.24 ± 0.32 for highly cited works compared to -1.79 ± 0.3 for average papers. However, the calculated p-value of 0.07 indicates that these results are not statistically significant at the conventional 0.05 level.

Full training with more steps on larger models, as well as more experimentation and method optimization, would yield more reliable results and be of modern relevance. Our documented code with step-by-step instructions is available in the repository \cite{Al-Feghali2024}.

\newpage

\section{ChemQA: Evaluating Chemistry Reasoning Capabilities of Multi-Modal Foundation Models}\label{sec:chemqa}

\textbf{\textit{Authors: Ghazal Khalighinejad, Shang Zhu, Xuefeng Liu
}}

\subsection{Introduction}

Current foundation models exhibit impressive capabilities when prompted with text and image inputs in the chemistry domain. However, it is essential to evaluate their performance on text alone, image alone, and a combination of both to fully understand their strengths and limitations. In chemistry, visual representations often enhance comprehension. For instance, determining the number of carbons in a molecule is easier for humans when provided with an image rather than SMILES annotations. This visual advantage underscores the need for models to effectively interpret both types of data. To address this, we propose ChemQA—a benchmark dataset containing problems across five question-and-answering (QA) tasks. Each example is presented with isomorphic representations: visual (images of molecules) and textual (SMILES). ChemQA enables a detailed analysis of how different representations impact model performance.

We observe from our results that models perform better when given both text and visual inputs compared to when they are prompted with image-only inputs. Their accuracy significantly decreases when provided with only visual information, highlighting the importance of multimodal inputs for complex reasoning tasks in chemistry.

\subsection{ChemQA: A Benchmark Dataset for Multi-Modal Chemical Understanding}

Inspired by the existing work of IsoBench \cite{Fu2024} and ChemLLMBench \cite{Guo2023}, we create a multimodal question-and-answering dataset on chemistry reasoning, ChemQA \cite{Zhu2024} containing five QA tasks:

\begin{itemize}
    \item \textbf{Counting Numbers of Carbons and Hydrogens in Organic Molecules:} adapted from the 600 PubChem molecules created by \cite{Guo2023}, evenly divided into validation and evaluation datasets.
    \begin{itemize}
        \item Example: Given a molecule image or its SMILES notation, identify the number of carbons and hydrogens.
    \end{itemize}

    \item \textbf{Calculating Molecular Weights in Organic Molecules:} adapted from the 600 PubChem molecules created by \cite{Guo2023}, evenly divided into validation and evaluation datasets.
    \begin{itemize}
        \item Example: Given a molecule image or its SMILES notation, calculate its molecular weight.
    \end{itemize}

    \item \textbf{Name Conversion: From SMILES to IUPAC:} adapted from the 600 PubChem molecules created by \cite{Guo2023}, evenly divided into validation and evaluation datasets.
    \begin{itemize}
        \item Example: Convert a given SMILES string or a molecule image to its IUPAC name.
    \end{itemize}

    \item \textbf{Molecule Captioning and Editing:} inspired by \cite{Guo2023}, adapted from the dataset provided in \cite{Edwards2022}, following the same training, validation, and evaluation splits.
    \begin{itemize}
        \item Example: Given a molecule image or its SMILES notation, find the most relevant description of the molecule.
    \end{itemize}

    \item \textbf{Retro-synthesis Planning:} inspired by \cite{Guo2023}, adapted from the dataset provided in \cite{Irwin2022}, following the same training, validation, and evaluation splits.
    \begin{itemize}
        \item Example: Given a molecule image or its SMILES notation, find the most likely reactants that can produce the molecule.
    \end{itemize}
\end{itemize}

\begin{figure}[h!]
    \centering
    \includegraphics[width=0.8\textwidth]{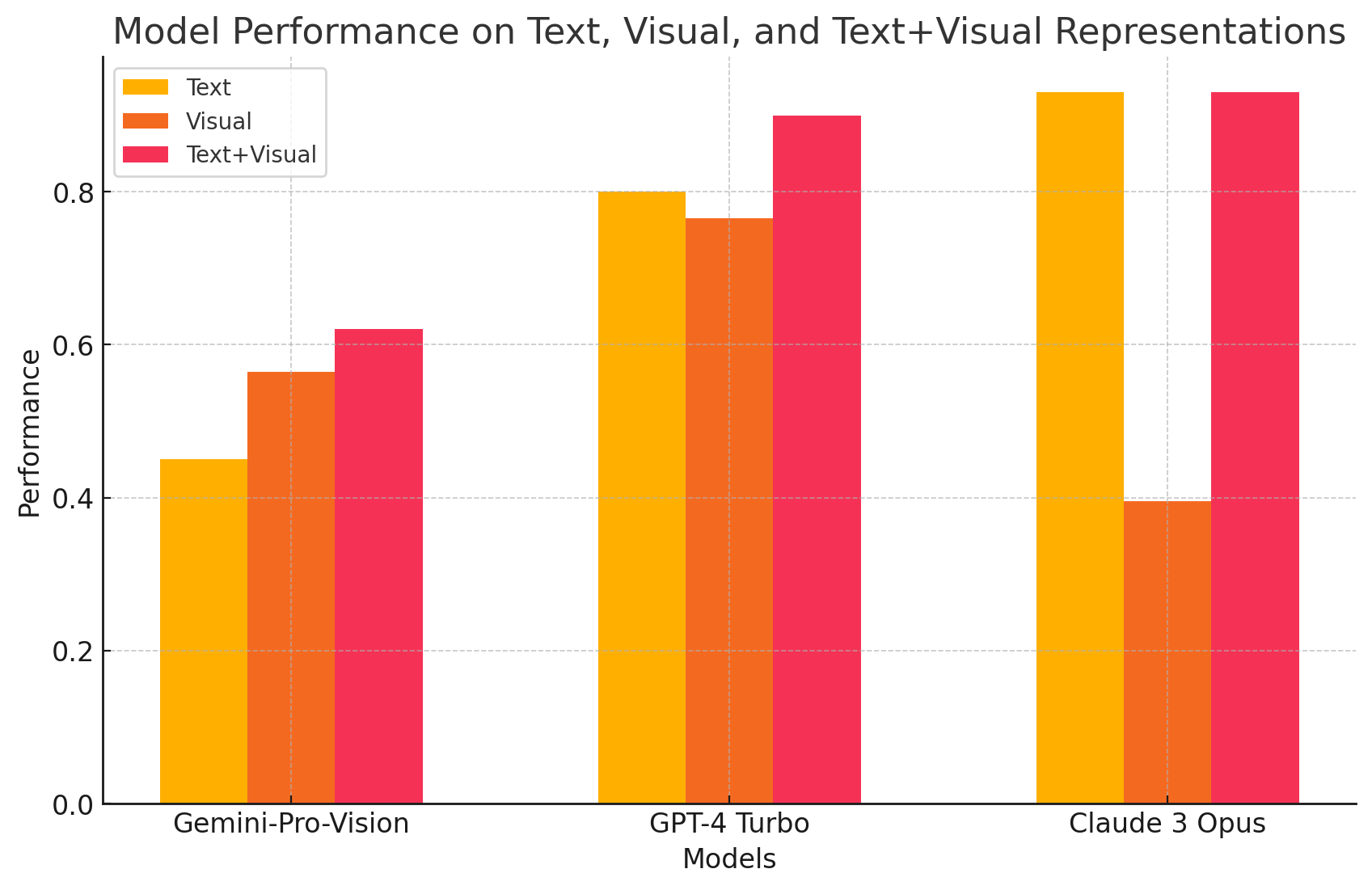} 
    \caption{Performance of Gemini Pro, GPT-4 Turbo, and Claude3 Opus on text, visual, and text+visual representations. The plot shows that models achieve higher accuracy with combined text and visual inputs compared to visual-only inputs.}
    \label{fig:performance}
\end{figure}

\subsection{Evaluating Foundation Model Performances on ChemQA}

The model evaluation results are shown in \autoref{fig:performance}. According to the plot, models perform better with text and visual inputs combined, while their accuracy drops when given only visual inputs. Claude performs well in text-only tasks, whereas Gemini and GPT-4 Turbo perform better with visual or combined inputs. This highlights the importance of evaluating models with different input modalities to understand their full capabilities and limitations.

\subsection{Conclusion and Future Work}

The evaluation of multimodal language models in the chemistry domain demonstrates that the integration of both text and visual inputs significantly enhances model performance. This suggests that for complex reasoning tasks in chemistry, multimodal approaches are more effective than relying on a single type of input. Future work should continue to explore and refine these multimodal strategies to further improve the accuracy and applicability of AI models in specialized fields such as chemistry.

\newpage
\section{LithiumMind - Leveraging Language Models for Understanding Battery Performance}\label{sec:lithium-heart}

\textbf{\textit{Authors: Xinyi Ni, Zizhang Chen, Rongda Kang, Sheng-Lun Liao, Pengyu Hong, Sandeep Madireddy}}

\subsection{Introduction}

In this project, we explore multiple applications of Large Language Models (LLMs) in analyzing the performance of lithium metal batteries. Central to our investigation is the concept of Coulombic efficiency (CE), a critical measure in battery technology that assesses the efficiency of electron transfer during a battery's charge and discharge cycles. Improving the CE is essential for advancing the adoption of high-energy-density lithium metal batteries, which are crucial for next-generation energy storage solutions. A key focus of our analysis is the role of the liquid electrolyte engineering, which is instrumental in determining the battery's cyclability and improving the CE.

We explored two methods of integrating LLM as a supplemental tool in battery research. First, we utilize LLMs as information extractors to distill structural knowledge essential for the design of electrolytes from the vast amount of paper. Second, we introduce an LLM-powered chatbot specifically tailored to respond to inquiries related to lithium metal batteries. We believe these applications of LLMs may potentially enhance our capabilities in battery innovation, streamlining research processes and increasing the accessibility of specialized knowledge. Our code is available at: \url{https://github.com/KKbeckang/LiGPT-Beta}.

\subsection{Investigation 1: Structural Knowledge Extraction}

\paragraph{Motivation}

In this project, we utilize LLM to extract CE-electrolyte pairs from over 70 papers. A recent study \cite{Dataset_S01} (Dataset S01) provides a reliable database about the relationship between Coulombic Efficiency (CE) and battery electrolytes, along with a list of relevant papers. The data was filtered and cleaned by human experts. We managed to automate the data collection procedure through LithiumMind and aimed to discover more relationships from the paper list. The pipeline is describe in the left column of Figure 1.

\paragraph{Method}

The pipeline consists of the following steps:
\begin{itemize}
    \item \textbf{Parse PDF:} The raw papers were saved as PDF files and loaded into text using a PDF parser.
    \item \textbf{Retrieve Relevant Context:} Instead of extracting information directly from the whole paper, we ingest the documents into a vector datastore. We combined three domain-specific embedding models - MaterialBERT, ChemBERT, and OpenAI - together as a powerful retriever through the LangChain LOTR module. The retriever finds 10 text chunks for each paper that are most relevant to Coulombic Efficiency.
    \item \textbf{CE Extraction:} We defined the schema for extracting key information, including Coulombic Efficiency and the solvent and salt of the electrolyte. We provided few-shot in-context instruction to GPT-4 Turbo JSON mode. The extracted output is saved in JSON format.
\end{itemize}

\paragraph{Preliminary Results}

The LLM extracted 334 CE-  electrolyte pairs in 71 papers, while the original paper found 152 pairs. Since it is difficult to verify our results without the help of human experts, we filtered the extracted pairs through the Coulombic Efficiency and found 46 matches to the original dataset. We classified these verifiable data into three categories: \textit{correct} (the extracted data exactly matches the human-labeled data); \textit{incorrect} (the types or amounts of the solvent/salt do not match the labels); and \textit{unknown} (the extracted data provides more details than the human-labeled data). For example, the label only shows the types of the salts, but the extracted data contains not only the correct types but also the mix ratio of different salts. The results are shown in \autoref{tab:results}:

\begin{table}
\centering
    \caption{Results}
\label{tab:results}
\begin{tabular}{|c|c|c|c|}
\hline
\textbf{Correct} & \textbf{Incorrect} & \textbf{Unknown} \\
72.4\% & 23.4\% & 4.2\% \\
\hline
\end{tabular}
\end{table}

\begin{figure}[h!]
    \centering
    \includegraphics[width=0.8\textwidth]{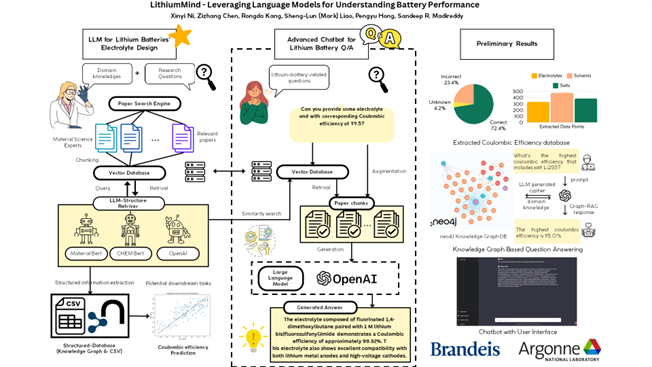} 
    \caption{Summary of our LLM hackathon project.}
\end{figure}

\paragraph{Future Work}

One major challenge is the low recall of the extracted information, as only 46 out of 152 labeled pieces of information were retrieved. Upon investigating the papers, we found that much of the Coulombic Efficiency was recorded in tables and figures, which were dropped during PDF parsing. It is necessary to introduce multimodal LLMs to further investigate those papers.

\subsection{Investigation 2: Advanced Chatbot for Lithium Battery Q/A}

In this exploration, we utilized a curated collection of research papers focused on lithium metal batteries to construct a vector database and developed a chatbot employing a Retrieval-Augmented Generation (RAG) framework. Our Q/A pipeline includes two types of answer strategies: General Q/A and Knowledge-graph-based Q/A. The pipeline is described in Figure 1.

\paragraph{General Q/A Building}

In this section, we detail our comprehensive Q/A pipeline designed for the exploration of lithium metal battery-related research. We began by sourcing and downloading 71 research papers pertinent to lithium metal batteries. The information extracted from these papers was encoded into vectors and stored in Chroma databases. To best reflect the specialized language of the field, we created two distinct databases: one utilizing MaterialsBERT for materials science content, and another using ChemBERT for chemical context.

During the retrieval phase, we employ LOTR (MergerRetriever) to enhance the precision of document retrieval. Upon receiving a user query, the system retrieves relevant document segments from each database. It then removes any redundant results from the merged outputs and selects the top 10 most pertinent document chunks. Finally, both the selected context and the user query are processed by GPT-4 Turbo to generate an informed and accurate response. This pipeline exemplifies a robust integration of state-of-the-art technologies to facilitate advanced research interrogation and knowledge discovery in the domain of lithium metal batteries.

\paragraph{Knowledge Graph-based Q/A Building}

We built a knowledge graph with the extracted information using Neo4j. The knowledge graph consists of four node types:

\begin{itemize}
    \item \textbf{Solvent node:} properties include name, SMILES, density, and weight.
    \item \textbf{Salt node:} properties include name, SMILES, and weight.
    \item \textbf{Electrolyte node:} properties include name and Coulombic Efficiency.
    \item \textbf{Reference node:} properties include title, content, and page number and,
\end{itemize}

three edge types:

\begin{itemize}
    \item \textbf{(electrolyte)-[:VOLUME]} ->(solvent)
    \item \textbf{(electrolyte)-[:CONCENTRATION]} ->(salt)
    \item \textbf{(electrolyte)-[:CITED]} ->(reference)
\end{itemize}

The built knowledge graph can be accessed and visualized using the Neo4j web application.

\paragraph{Knowledge Graph Enhanced Question Answering}

By using GraphCypherQAChain in LangChain, LLMs can generate Cypher queries to solve users' questions. This capability allows the LLMs to address user queries by filtering and leveraging relationships between data points, which is particularly valuable in complex domains such as lithium battery technology. This integration ensures that our RAG pipeline is adept at handling domain-specific questions and excels in scenarios where understanding the interconnections within data is crucial.

\clearpage
\newpage
\section{KnowMat: Transforming Unstructured Material Science Literature into Structured Knowledge}\label{sec:knowmat}

\textbf{\textit{Authors: Hasan M. Sayeed, Ramsey Issa, Trupti Mohanty, Taylor Sparks}}

\subsection{Introduction}
The rapid growth of materials science has led to an explosion of scientific literature, often presented in unstructured formats that pose significant challenges for systematic data extraction and analysis. To address this, we developed KnowMat, a novel tool designed to transform complex, unstructured material science literature into structured knowledge. Leveraging advanced Large Language Models (LLMs) such as GPT-3.5, GPT-4, Llama 3, KnowMat automates the extraction of critical information from scientific texts, converting them into structured JSON formats. This tool not only enhances the efficiency of data processing but also facilitates deeper insights and innovation in material science research.
KnowMat's user-friendly interface allows researchers to input material science papers, which are then parsed and analyzed to extract key insights. The tool's versatility in handling multiple input files and its capacity for customization through sub-field specific prompts make it an indispensable resource for researchers aiming to streamline their workflow. Additionally, KnowMat's ability to integrate with other tools and platforms, along with its support for various LLMs, ensures that it remains adaptable to the evolving needs of the research community.

\subsection{Method}
\paragraph{Data Parsing and Extraction}
The KnowMat workflow begins with parsing unstructured text from material science literature using a tool called Unstructured \cite{Unstructured}. This tool reads the input file, separates out the sections, and stores everything in a machine-readable format. This initial step involves identifying relevant sections and extracting pertinent information related to material compositions, properties, and experimental conditions.
\paragraph{Customizable Prompts}
KnowMat provides field-specific prompts for several fields, and it offers the flexibility for users to customize these prompts further or create their own prompts for new fields. This feature ensures that the extracted data is both relevant and comprehensive, tailored to the specific needs of the researcher. The interface allows users to define the scope and focus of the extraction process effectively.
\paragraph{Integration and Interoperability}
To enhance usability and interoperability, KnowMat supports seamless integration with other tools and platforms. Extracted results can be easily exported in CSV format, enabling straightforward data sharing and further analysis. The tool's flexibility extends to its compatibility with various LLMs, including both subscription-based models like GPT-3.5 and GPT-4 \cite{OpenaiModels}, and open-source models like Llama 3 \cite{Llama3}. This ensures that researchers can select the most suitable LLM for their specific requirements.

\begin{figure}[h!]
    \centering
    \includegraphics[width=0.99\textwidth]{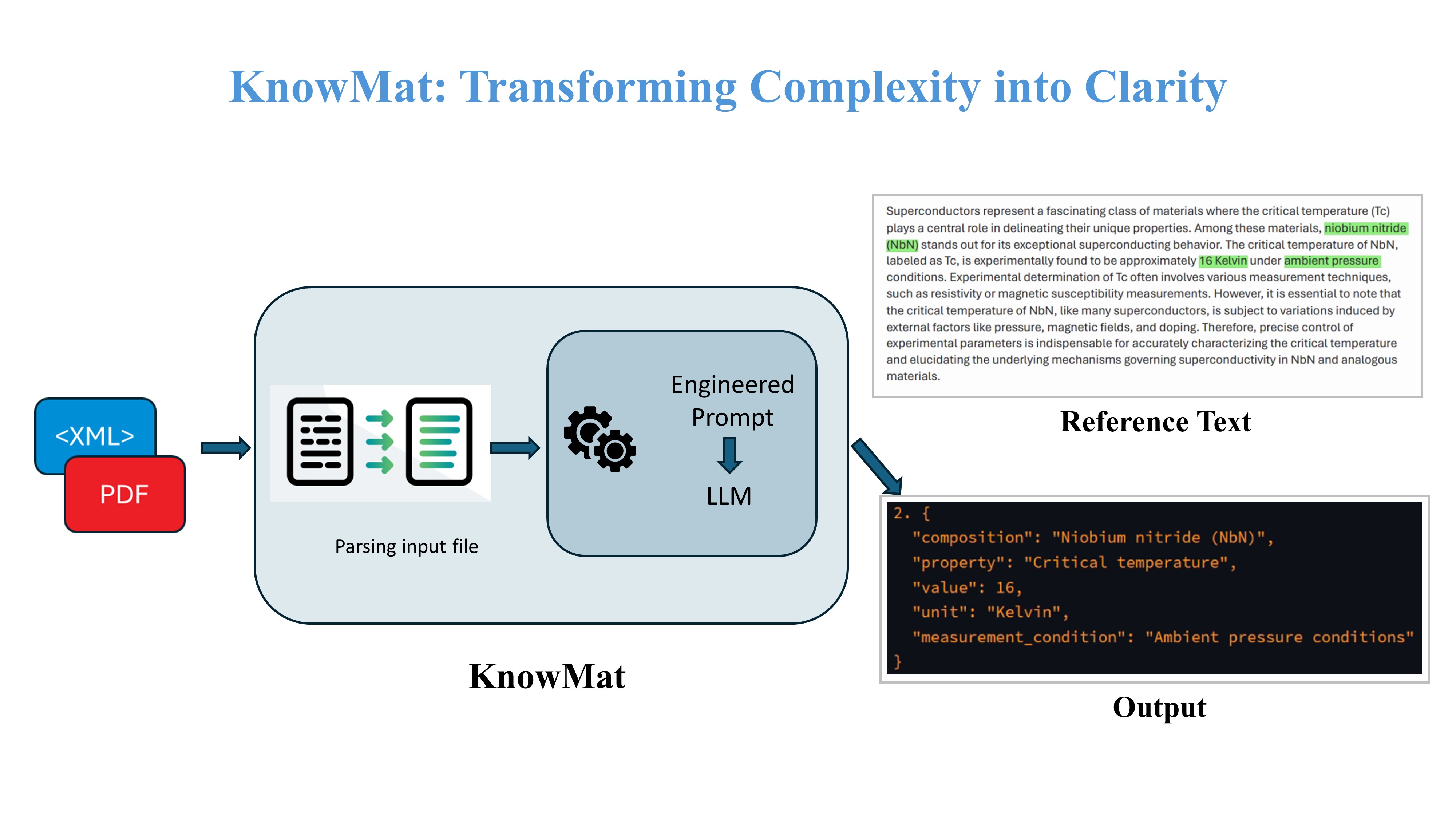} 
    \caption{KnowMat Workflow. The graphical abstract illustrates the KnowMat workflow, which begins with parsing input files in XML or PDF formats. The Large Language Model (LLM) powered by engineered prompts processes the reference text to extract key information, such as material composition, properties, and measurement conditions, and converts it into a structured JSON output.
    \label{fig:knowmat_workflow}}
\end{figure}

\subsection{Preliminary Results}
In initial tests, KnowMat demonstrated promising results in the efficiency and accuracy of data extraction from material science literature. The tool successfully parsed and structured information from multiple papers, converting complex textual data into actionable insights. For instance, in a study involving superconductors, KnowMat accurately extracted detailed information on compositions, critical temperatures, and experimental conditions, presenting them in a structured JSON format.

\subsection{Future Development}
Looking ahead, future developments for KnowMat include enhancing the editing capabilities for field-specific prompts, improving the handling of multiple input files in a single operation, and expanding output options to further enhance flexibility and usability. Continuous improvement and expansion of field-specific prompts will ensure that KnowMat remains a valuable tool for researchers across various domains of material science.
In conclusion, KnowMat represents a significant advancement in the field of knowledge extraction from scientific literature. By converting unstructured material science texts into structured formats, it provides researchers with powerful tools to unlock insights and drive innovation in their fields.

\subsection{Data and code availability}
\url{https://github.com/sparks-sayeed/LLMs_for_Materials_and_Chemistry_Hackathon}

\newpage
\section{Ontosynthesis}\label{sec:ontosynthesis}

\textbf{\textit{Authors: Qianxiang Ai, Jiaru Bai, Kevin Shen, Jennifer D'Souza, Elliot Risch}}

\subsection{Introduction}
Organic synthesis is often described in unstructured text without a standard taxonomy, which makes synthesis data less searchable and less compatible with downstream data-driven tasks (e.g., retrosynthesis, condition recommendation) compared to structured records. The specificities and writing styles of these descriptions also vary, ranging from simple sentences about chemical transformations to long paragraphs that include procedure details. This leads to ambiguities, unidentified missing information, challenges in comparing different syntheses, and can impede proper experimental reproduction.

In last year's hackathon, we fine-tuned an open-source LLM to extract data structured in the Open Reaction Database schema from synthesis procedures \cite{Ai2024}. While this method has proved to be successful for patent text, it relies on existing unstructured-structured data pairs and does not generalize well to non-patent writing styles. The dependency of fine-tuning on existing data makes it less useful, especially when considering new writing styles, or newly developed data structures or ontologies are preferred.

In this project, we explore the potential of LLMs in structured data extraction without fine-tuning. Specifically, given an ontology (formally defined concepts and relationships) for organic synthesis, we aim to extract structured information as Resource Description Framework (RDF) graphs from unstructured text using LLMs with zero-shot prompting. RDF is a World Wide Web Consortium (W3C) standard that serves as a foundation for the Semantic Web and expressing meaning and relationships between data. While LLMs can create ontologies on the fly for a given piece of text, “grounding” to a pre-specified ontology allows standardizing the extracted data and reasoning with existing axioms. The extraction workflow is wrapped in a web application which also allows visualization of the extracted results. We showcased the capability of our application with case studies where RDFs were extracted from reaction descriptions of different complexities and specificities. 

\subsection{Information extraction workflow}
OpenAI’s GPT model (\texttt{gpt-4-turbo-2024-04-09}) is used for structured information extraction through its \texttt{ChatComplete} API. The prompt for an individual extraction task consists of two parts:
\begin{enumerate}
    \item The \texttt{System} prompt: The given ontology in OWL format based on which the RDF graph is defined, along with supporting instructions on the task. The full prompt template is included in the project GitHub repository \cite{Ontosynthesis2024}. Two ontologies were used in this study:
    \begin{enumerate}
    \item OntoReaction: A reaction ontology previously used in distributed self-driving laboratories \cite{Bai2024};
    \item Synthesis Operation Ontology: A process chemistry ontology designed for common operations used in organic synthesis \cite{AiGithub}.
    \end{enumerate}
    \item The \texttt{User} prompt: The unstructured text from which the RDF graph is extracted.
\end{enumerate}

The final OpenAI API request is a combination of the \texttt{User} and \texttt{System} prompts:

\begin{lstlisting}
from openai import OpenAI
client = OpenAI(api_key=api_key)
completion = client.chat.completions.create(
   model="gpt-4-turbo",
   messages=[
      {"role": "system", "content": system_prompt,},
      {"role": "user", "content": unstructured_data.text}
   ],
)
\end{lstlisting}

Alternatively, one can use GPT assistants and provide the \texttt{System} prompt as a base Instruction through OpenAI’s web user interface. 

\subsection{Application}
We collected a set of eight reaction descriptions taken from patents, journal articles (main text or supporting information), and lab notebooks. Each of them is assigned a complexity rating and a specificity rating using three-point scales. Based on these test cases, we found our workflow was able to produce valid RDF graphs representing the chemical reactions, even for multi-step reactions including many elements. Expert inspections indicate the resulting RDF graphs better represent the unstructured text when OntoReaction is used as the target ontology compared to the larger Synthesis Operation Ontology (the latter contains more classes and object properties).

Since the extracted data is in RDF format, they can be readily visualized using interactive graph libraries. Using \texttt{dash-cytoscape} \cite{DashCytoscape},  we created an interface application to the extraction workflow. The interface allows submitting unstructured text as input to the extraction workflow with a user-provided OpenAI API key, retrieving and interactively visualizing the extracted knowledge graph, as well as displaying the extracted RDF text. A file-based caching routine is used to store previous extraction results. All code and test cases are available in the project GitHub repository \cite{Ontosynthesis2024}. 

\begin{figure}[h!]
    \centering
    \includegraphics[width=0.99\textwidth]{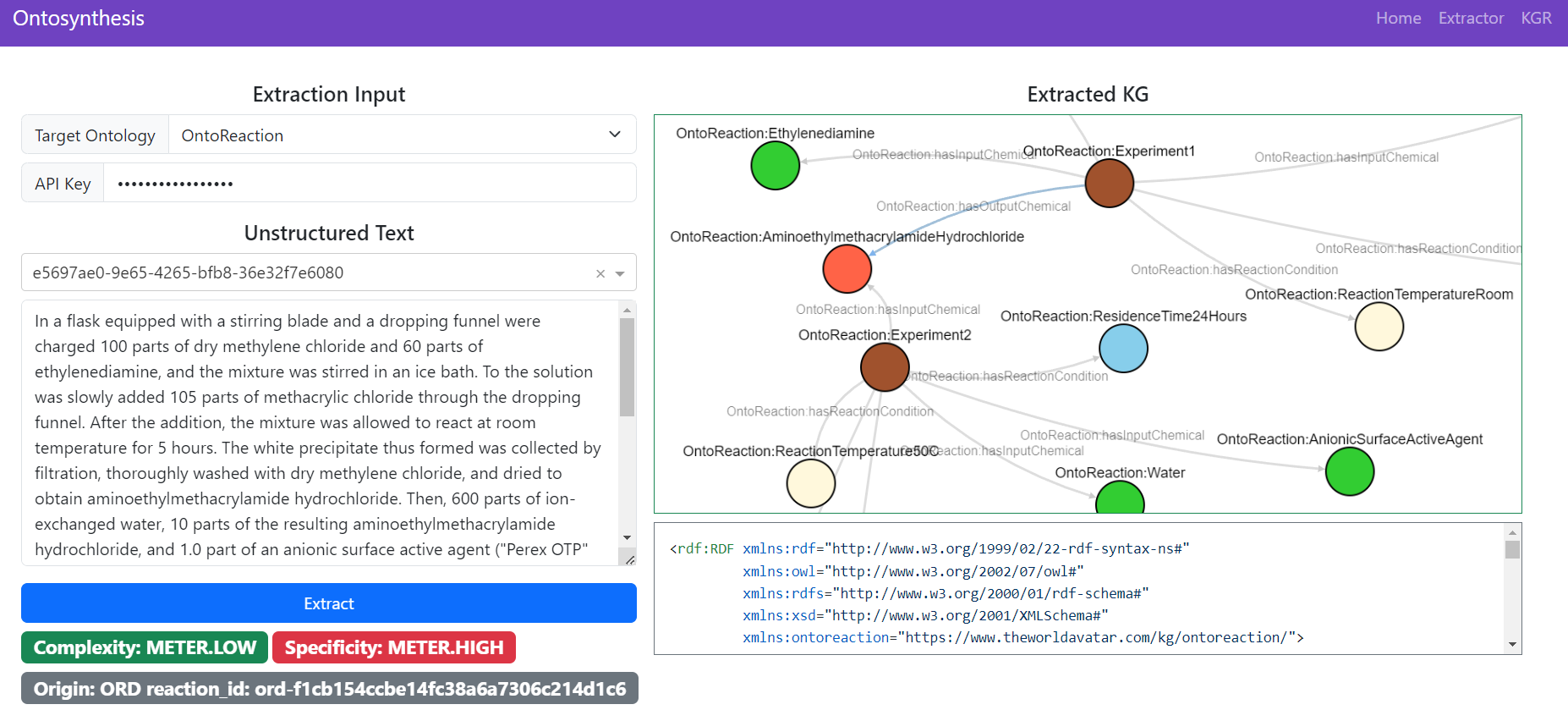} 
    \caption{
    \label{fig:ontosynthesis_workflow}}
\end{figure}

\subsection{Acknowledgements}
Q.A. acknowledges support by the National Institutes of Health under award number U18TR004149. The content is solely the responsibility of the authors and does not necessarily represent the official views of the National Institutes of Health. J. D. acknowledges the SCINEXT project (BMBF, German Federal Ministry of Education and Research, Grant ID: 01lS22070).

\newpage
\section{Knowledge Graph RAG for Polymer Simulation}\label{sec:kg-rag-polymers}

\textbf{\textit{Authors: Jiale Shi, Weijie Zhang, Dandan Tang, Chi Zhang
}}

\begin{figure}[h!]
    \centering
    \includegraphics[width=0.99\textwidth]{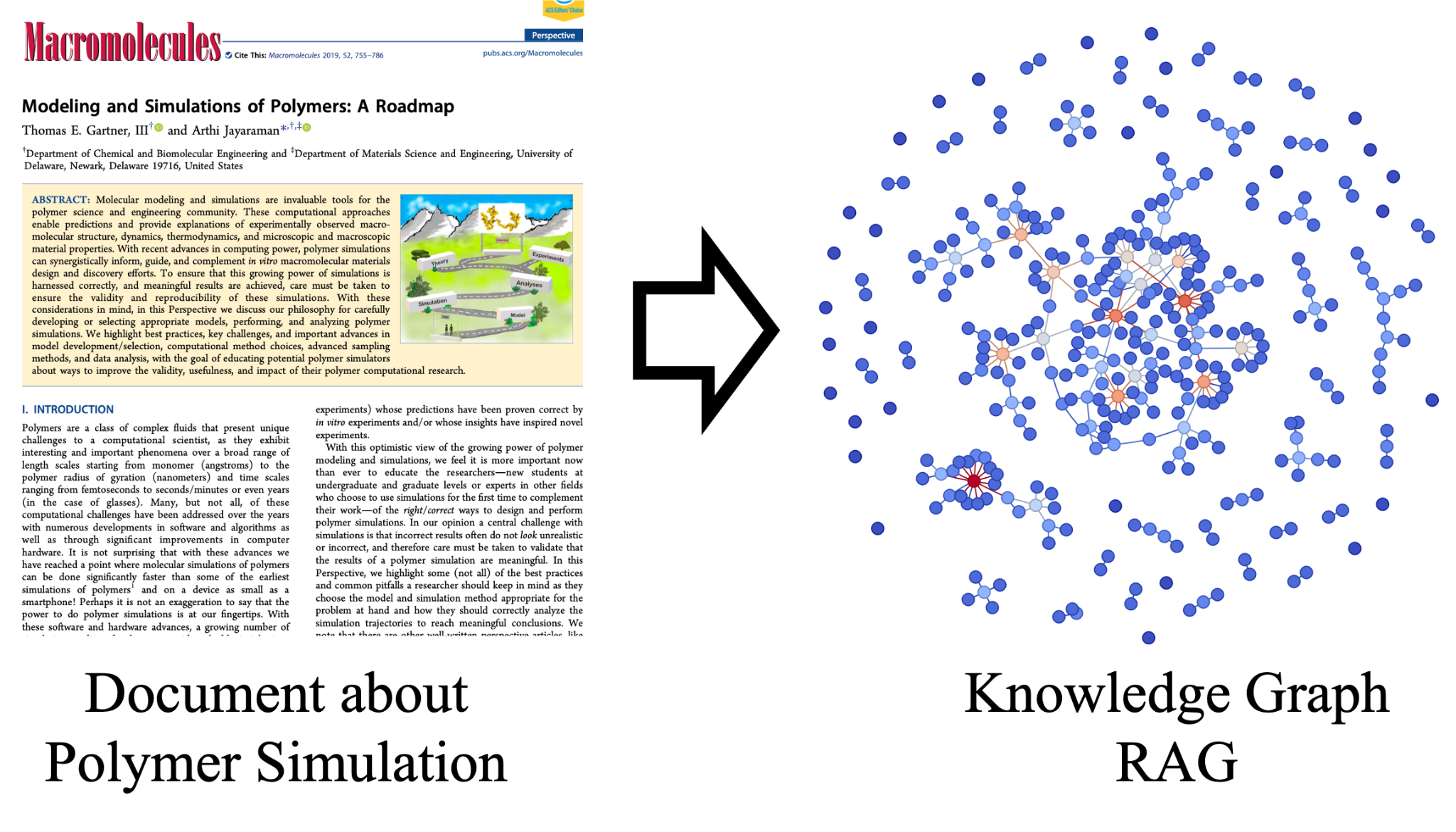} 
    \caption{Creating Knowledge Graph Retrieval-Augmented Generation (KGRAG) for Polymer Simulation.
    \label{fig:polymerRAG}}
\end{figure}

Molecular modeling and simulations have become essential tools in polymer science and engineering, offering predictive insights into macromolecular structure, dynamics, and material properties. However, the complexity of polymer simulations poses challenges in model development/selection, computational method choices, advanced sampling techniques, and data analysis. While literature \cite{Gartner2019} provides guidelines to ensure the validity and reproducibility of these simulations, these resources are often presented in massive, unstructured text formats, making it difficult for new learners to systematically and comprehensively understand the correct approaches for model selection, simulation execution, and post-processing analysis. Therefore, this study presents a novel approach to address these challenges by implementing a Knowledge Graph Retrieval-Augmented Generation (KGRAG) system for building an AI chatbot focused on polymer simulation guidance and education.

Our team utilized the large language model GPT-3.5 Turbo \cite{OpenAI2024b} and Microsoft's GraphRAG \cite{GraphRAG2024, Edge2024} framework to extract polymer simulation-related entities and relationships from unstructured documents, constructing a comprehensive KGRAG, as shown in \autoref{fig:polymerRAG}, where the nodes are colored by their degrees. Those nodes with high degrees include “Polymer Simulation”, “Atomistic Model,” “CG Model,” “Force Field,” and “Polymer Informatics,” which are all the keywords about polymer simulation and modeling, illustrating the effective performance of entity extraction. We run the query engineering of KGRAG to ask questions. For comparative analysis, we also implemented a baseline Retrieval-Augmented Generation (RAG) system using LlamaIndex \cite{LlamaIndex2024}. Upon comparing the responses from the baseline RAG and KGRAG by human experts, we found that the KGRAG demonstrates substantial improvements in question-answering performance when analyzing complex and high-level information about polymer simulation. This improvement is attributed to KGRAG's ability to capture semantic concepts and uncover hidden connections within the data, providing more accurate, logical, and insightful responses compared to traditional vector-based RAG methods.

This study contributes to the growing field of data-driven approaches in polymer science by offering a powerful tool for knowledge extraction and synthesis. Our KGRAG system shows promise in enhancing the understanding of massive unstructured polymer simulation guidance in the relevant literature, potentially improving the validity and reproducibility of these polymer simulations, and accelerating the development of new polymer simulation methods. We found that the quality of prompts is crucial for effective entity extraction and knowledge graph construction. Therefore, our future work will focus on optimizing prompts for entity extraction, relationship building, and knowledge graph construction to further improve the system's performance and applicability in polymer simulation research.

\subsection{Data availability}
Repository: \url{https://github.com/shijiale0609/KG-RAG-LLM-Polymers }
 
\subsection{Author}
\begin{itemize}
    \item Jiale Shi, Department of Chemical Engineering, Massachusetts Institute of Technology, Cambridge, Massachusetts 02139, United States (jialeshi@mit.edu)
    \item Weijie Zhang, Department of Chemistry, University of Virginia, Charlottesville, Virginia 22904, United States (shv9px@virginia.edu)
    \item Dandan Tang, Department of Psychology, University of Virginia, Charlottesville, Virginia 22904, United States (gux8df@virginia.edu)
    \item Chi Zhang, Department of Automation Engineering, RWTH Aachen University, Aachen, North Rhine-Westphalia, 52062, Germany (chi.zhang2@rwth-aachen.de)
\end{itemize}

\newpage
\section{Synthetic Data Generation and Insightful Machine Learning for High Entropy Alloy Hydrides}\label{sec:hea-synthetic-data}

\textbf{\textit{Authors: Tapashree Pradhan, Devi Dutta Biswajeet
}}

The generation of synthetic data in materials science and chemistry is traditionally performed by machine learning interatomic potentials (MLIPs) that approximate the first principle functional form used to compute wave functions of known electronic configurations \cite{Focassio2024}. Synthetic data is a component of the active-learning feedback loop that is utilized to retrain these potentials. The purpose of this active-learning component is to incorporate a wider range of physical conditions into the potential’s application domain. However, the initial cost of data generation to train the MLIPs is a major setback for complex chemistries like in the case of high entropy alloys (HEAs).

The potential application of high entropy alloys in hydrogen storage \cite{Marques2021} demands acceleration in the computation of surface and mechanical properties of the alloys by better approximation of the potential landscape. The use of Large Language Models (LLMs) in the generation of synthetic data to tackle this bottleneck problem poses an alternative to the traditional MLIPs. LLMs like ChatGPT, Llama3, Claude, and Gemini \cite{Jain2022} learn from text-embeddings of the training data, capturing inherent trends between semantic or numerical relationships of the text and making them suitable for learning certain complex relationships in materials physics that might be present in the language itself.

The current work aims to build LLM applications working in conjunction with an external database of high entropy alloy hydrides via Retrieval-Augmented Generation (RAG) \cite{Lewis2020b} to populate synthetic data for predictive modeling later. The inbuilt RAG feature of GPT-4 enables us to write a few prompts to make a synthetic data generator utilizing a custom dataset of HEA hydrides. The work also utilizes OpenAI’s API \cite{OpenAI2020} and the text-embedding-3-large model \cite{OpenAI2023} to configure custom generators that can be fine-tuned via prompts for synthetic data generation.

\begin{figure}[h!]
    \centering
    \includegraphics[width=0.8\textwidth]{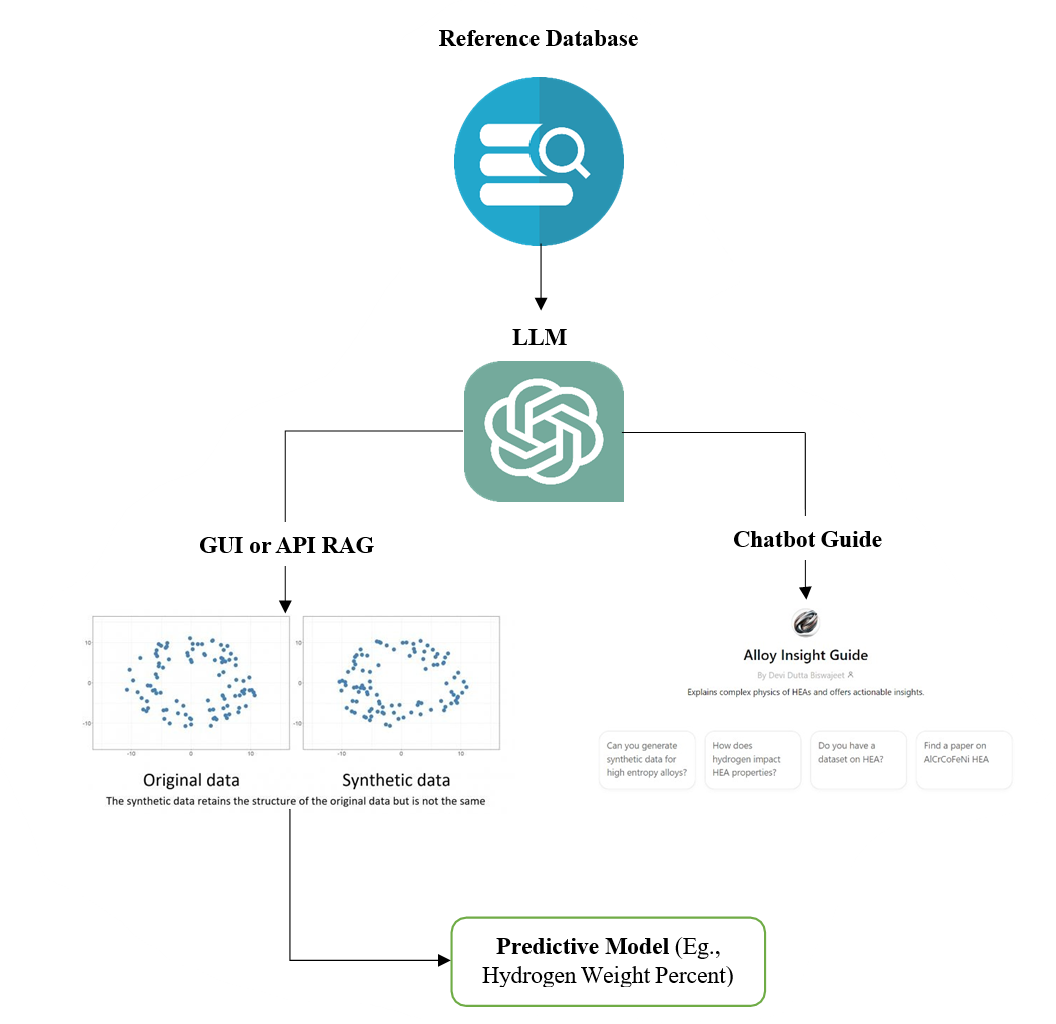} 
    \caption{Insightful machine learning for HEA hydrides}
    \label{fig:HEA}
\end{figure}

The development of the entire product is aimed at a web-based application that allows users to upload their datasets and instruct the GPT model to generate more entries that can serve as training data for predictive ML models like hydrogen capacity predictors. The term “Insightful Machine Learning” refers to a sequential pipeline starting with (a) a reference database that serves as the retrieval component of an LLM, (b) the generation of synthetic data and features, and (c) getting insights from a chatbot on physics of the problem having multiple retrieval components inclusive of the predictive model. \autoref{fig:HEA} shows the flowchart of the pipeline which is currently at the prototype stage under development. The current code to generate synthetic data is available for use and modification.

\subsection{Future Directions}

The proposed pipeline shown in Figure 1 requires a validation stage which is essential for active learning. The ongoing work involves the search for novel validation techniques taking inspiration from recently published works on information theory.

The future direction of the work is to complete the validation phase and have a product utilizing the pipeline for high entropy hydrides that can accelerate the search and discovery process without performing complex first principle calculations for data generation and training.

\clearpage
\newpage
\section{Chemsense: Are large language models aligned with human chemical preference?}\label{sec:chemsense}

\textbf{\textit{Authors: Martiño Ríos-García, Nawaf Alampara, Mara Schilling-Wilhelmi, Abdelrahman Ibrahim, Kevin Maik Jablonka}}

\subsection{Introduction}
Generative AI models are revolutionizing molecular design by learning from vast chemical datasets to create novel compounds \cite{Bilodeau2022}. The challenge in molecular design goes beyond just creating new structures. A key issue is ensuring the designed molecules have specific useful properties, e.g., high solubility, high synthesizability, etc. LLMs that can be conditioned on the desired property seem to be a promising solution \cite{Jablonka2024d}. However, current frontier models often lack an innate chemical understanding, which can lead to impractical or even dangerous molecular designs \cite{Mirza2024b}.

A factor that distinguishes many successful chemists is their chemical intuition. This intuition, for instance, describes the preference for certain compounds that are not grounded in knowledge that can be easily conveyed but rather by tacit knowledge accumulated over years of experience. If models could possess this chemical intuition, they would be more useful for real-world scientific applications.

In this work, we introduce ChemSense, in which we explore how well frontier models are aligned with human experts in chemistry. By aligning AI with human knowledge and preferences, one can aim to create tools that can propose feasible, safe, and desirable molecular structures, bridging the gap between theoretical capabilities and practical scientific needs. Moreover, ChemSense would help us in understanding the emergent alignment of frontier models with dataset scale and size.

\subsection{Related work}
Yang et al.\ (2024) investigated the ability of LLMs to predict human characteristics in their social life and showed that LLMs encountered great difficulties in predicting these. Mirza et al.\ (2024) benchmarked LLMs on various chemistry tasks; however, good performance in that benchmark does not guarantee alignment with human-expert intuitions. Chennakesavalu et al.\ (2024) aligned LLMs to generate low energy stable molecules with externally specified properties, they noticed huge improvement upon alignment compared to the base model.


\begin{figure}[h!]
    \centering
    \includegraphics[width=0.8\textwidth]{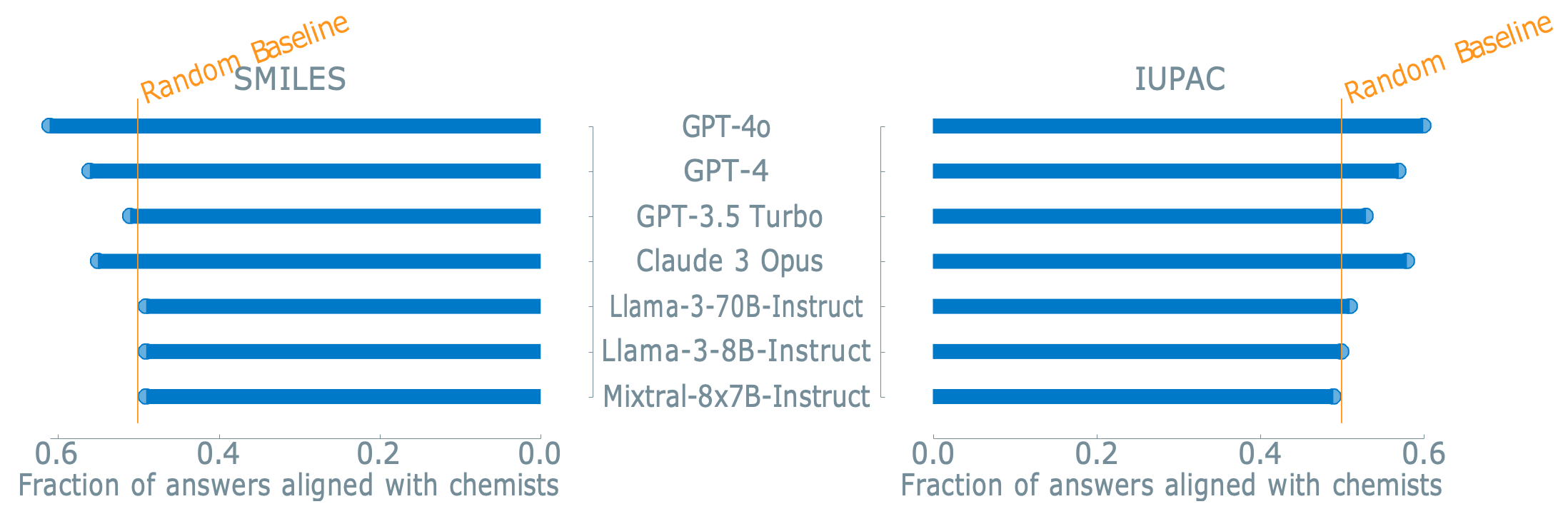} 
    \caption{Comparison of the alignment of the different LLMs with the SMILES (left) and IUPAC (right) molecular representations. For both representations, note that the random baseline is at a fixed value of 0.5 since all the questions were binary and the datasets are balanced.
    \label{fig:chemsense}}
\end{figure}

\subsection{Materials and Methods}
\paragraph{Chemical preference dataset}
The test set used during all the experimentation was constructed with 900 samples from the dataset published by Choung et al. (2023). This dataset contains the preferences of 35 Novartis medical chemists, resulting in more than 5000 question-answer pairs about organic synthesis. To choose the 900 samples, only questions in which both molecules could be converted into the IUPAC names (using the \href{https://cactus.nci.nih.gov/chemical/structure}{chemistry name resolver}) were selected. In that way, we ensure some homogeneity in the test dataset.

\paragraph{LLM evaluation}
To study the chemical preference of the actual LLMs, some of the models performing best on the ChemBench benchmark \cite{Mirza2024b} were prompted with a simple instruction prompt inspired by the question that was asked to collect the original data from the scientists of Novartis. Additionally, to study how molecular representations could affect the model, each of the 900 questions was given to the model using different molecular representations.

To ensure the correct format of the answers, OpenAI models as well as Claude-3 were constrained to answer only “A” or “B” using the \href{https://github.com/jxnl/instructor}{Instructor} package. On the other hand, the Llama models and Mixtral 8Bx7 used (in an unconstrained setup) through \href{https://console.groq.com/docs/quickstart}{Groq API service} and instead further prompted to encourage them only to answer “A” or “B”.

\subsection{Results}
\paragraph{Comparison of chemical preference of LLMs}
We compare the accuracy of the preference prediction of the different models and representations (Figure 1). The accuracy of all models ranges from 49\% to 61\% where 50\% is the accuracy one would obtain for a random choice. The GPT-4o model achieves the highest accuracy of all models and performs best with the SMILES and IUPAC representations. This might be explained by the widespread use of both of the representations and, therefore, a high occurrence in the training data of the model. We observe the same trend in the GPT-4 and GPT-3.5 Turbo predictions. For the other LLMs, representation seems to have a smaller impact on the accuracy with values that presumably are random.

\subsection{Discussion and Conclusions}
Our work shows that while some LLMs might show sparks of chemical intuition, they are mostly unaligned with expert intuitions, as their performance currently is only marginally better than random guessing. Interestingly, similar to previous work \cite{Jablonka2024d}, we find that one obtains different performance in different molecular representations, which might be attributable to different prevalences in the training data.
Our research demonstrates that preference tuning is a promising and underexplored approach to enhance models for chemists. Addressing model biases is crucial for ensuring fair and accurate predictions. By tackling these challenges, we can develop large language models (LLMs) that are both powerful and practical for chemists and materials scientists.

\newpage
\section{GlossaGen}\label{sec:glossagen}

\textbf{\textit{Authors: Magdalena Lederbauer, Dieter Plessers, Philippe Schwaller}}

\begin{figure}[h!]
    \centering
    \includegraphics[width=0.99\textwidth]{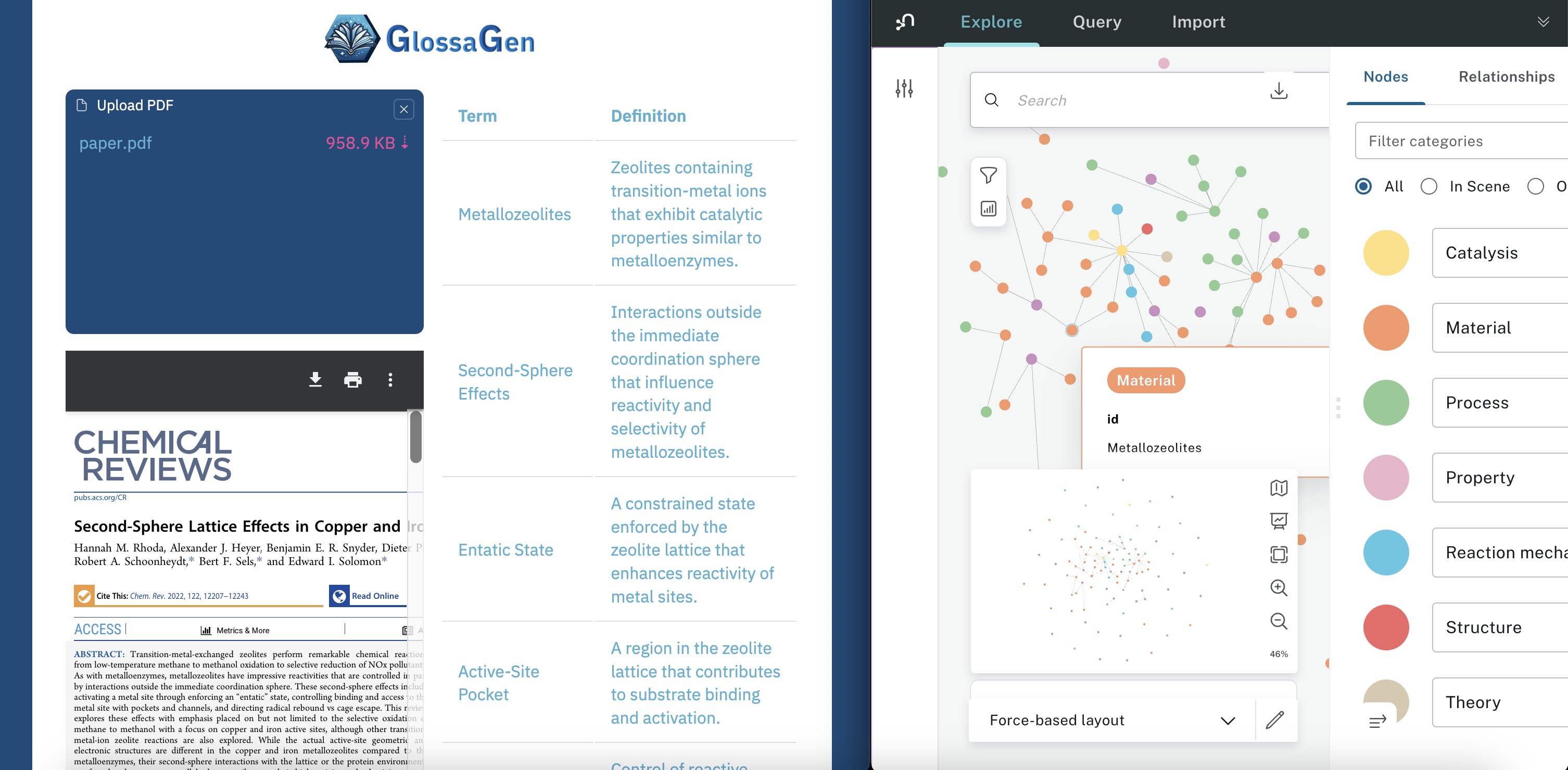} 
    \caption{Overview of (left) the graphical user interface (GUI) protoype and (right) the generated Neo4J knowledge graph (right).
Our results demonstrate that LLMs can greatly accelerate glossary creation, increasing the likelihood that authors will include a helpful glossary without the need for tedious manual effort. Additionally, an analysis of our test case by a zeolite domain expert revealed that LLMs produce good results, with about 70\% - 80\% of explanations requiring little to no manual changes.
    \label{fig:GGEN1}}
\end{figure}

Academic articles, particularly reviews, and grant proposals would greatly benefit from a glossary explaining complex jargon and terminology. However, the manual creation of such glossaries is a time-consuming and repetitive task. To address this challenge, we developed GlossaGen, an innovative framework that leverages large language models to automatically generate glossaries from PDF or TeX files, streamlining the process for academics. The generated glossary is not only a list of terms and definitions but also visualized as a knowledge graph, illustrating the intricate relationships between various technical concepts (see \autoref{fig:GGEN1}). 


The project's codebase was developed as a Python package on GitHub using a template \cite{Copier} and DSPy \cite{Khattab2023} as an LLM framework. This modular approach facilitates seamless collaboration and easy incorporation of new features.

To overcome the limitations of LLMs in directly processing PDFs, a prevalent format for scientific publications, we implemented a pre-processing step that converts papers into manageable text sequences. This step involves extracting textual information using PyMuPDF \cite{PyMuPDFb}, automatically obtaining the title and DOI, and chunking the text into smaller sections. This chunking preserves context and makes it easier for the models to handle the input. 

\begin{figure}[h!]
    \centering
    \includegraphics[width=0.99\textwidth]{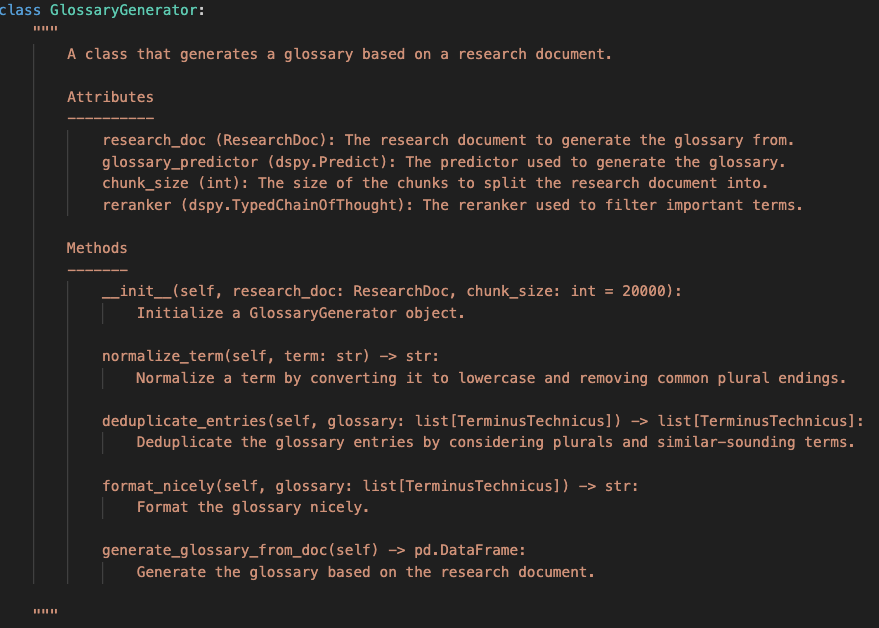} 
    \caption{Overview of the GlossaryGenerator class, responsible for processing text chunks and extracting relevant, correctly formatted terms and definitions.
    \label{fig:GGEN2}}
\end{figure}

We used GPT-3.5-Turbo \cite{GPT35Turbo} and GPT-4-Turbo \cite{GPT4Turbo} to extract scientific terms and their definitions from the text chunks. Employing Typed Predictors \cite{DSPy} and Chain-of-Thought prompting \cite{Wei2023} ensures the outputs are well-structured and contextually accurate, guiding the model to produce precise definitions through a simulated reasoning process. Post-processing involved identifying and removing duplicate entries, ensuring each term appears only once in the final glossary. \autoref{fig:GGEN2} shows details about the GlossaryGenerator class that was used to process documents into the corresponding glossaries. We selected a review article on zeolites \cite{Rhoda2022} (shown in \autoref{fig:GGEN1}) as a test publication to manually tune and evaluate the pipeline’s output.

The obtained glossary is transformed into an ontology that defines nodes and relationships for the knowledge graph. For instance, relationships like 'MATERIAL – exhibits $\rightarrow$ PROPERTY' illustrate how different terms are interconnected. The knowledge graph is constructed using the library Neo4J \cite{Neo4J} and Graph Maker \cite{GraphMaker} on the processed text chunks. We developed a user-friendly front-end interface with Gradio \cite{Gradio}, as shown in \autoref{fig:GGEN1}. This interface allows users to interact with the glossary, making it easier to navigate and customize the information.

The quick prototyping provided us with several ideas for future work. We can improve the glossary output by fine-tuning the LLM, incorporating retrieval-augmented generation, and parsing article images. Additionally, the user experience can be enhanced by allowing users to input specific terms for glossary explanations as a backup when the LLM omits certain terms. Integration with LaTeX would broaden usability, envisioning commands like \texttt{\textbackslash glossary} similar to \texttt{\textbackslash bibliography}. We also consider connecting the knowledge graph directly to the user interface and enhance its ontology creation feature. Overall, this rapidly developed prototype, with numerous future possibilities, demonstrates the potential of LLMs to assist researchers in their scientific outreach.

\end{document}